\documentclass[shortAfour,sageh,times]{sagej}

\usepackage{moreverb,url}

\usepackage{tabularx,ragged2e,array}
\usepackage[colorlinks,bookmarksopen,bookmarksnumbered,citecolor=red,urlcolor=red,draft]{hyperref}
\usepackage{arydshln}
\usepackage{caption}
\usepackage{subfigure}
\usepackage[ruled]{algorithm2e}

\usepackage{tikz}
\usepackage{pgfplots}
\usetikzlibrary{shapes,arrows,snakes,patterns,backgrounds,positioning,automata,3d,intersections}
\pgfdeclarelayer{background}
\pgfdeclarelayer{foreground}
\pgfsetlayers{background,main,foreground}
\tikzstyle{diagram} = [rounded corners=2.5pt]
\tikzstyle{blockdiagram} = [rounded corners=0.25pt,font={\scriptsize},semithick,>=latex]

\newcommand{\bgcoloratop}{gray!5}
\newcommand{\bgcolorabottom}{gray!5}

\newcommand{\bgcolorbtop}{gray!5}
\newcommand{\bgcolorbbottom}{gray!5}

\newcommand{\blockcolortop}{white}
\newcommand{\blockcolorbottom}{white}

\tikzstyle{block shadow} = [drop shadow={shadow xshift=1pt,shadow yshift=-1pt,opacity=0.33,top color=black!10,bottom color=black!25}]
\tikzstyle{background a} = [top color=\bgcoloratop,bottom color=\bgcolorabottom]
\tikzstyle{background b} = [top color=\bgcolorbtop,bottom color=\bgcolorbbottom]
\tikzstyle{block color} = [top color=\blockcolortop,bottom color=\blockcolorbottom,block shadow]
\tikzstyle{sum color} = [top color=white]

\definecolor{dlr_silver}{RGB}{138,141,143}
\definecolor{dlr_gray}{RGB}{70,70,70}
\definecolor{dlr_gray2}{RGB}{86,86,86}
\definecolor{dlr_yellow}{RGB}{255,255,220}

\definecolor{dlr_green}{RGB}{23,141,147}
\definecolor{dlr_green1}{RGB}{23,141,147}
\definecolor{dlr_green2}{RGB}{73,172,179}
\definecolor{dlr_green3}{RGB}{132,203,205}
\definecolor{dlr_green4}{RGB}{192,228,224}
\definecolor{dlr_green5}{RGB}{225,255,225}
\definecolor{soft_green}{RGB}{220,255,220}
\definecolor{soft_blue}{RGB}{240,240,255}

\definecolor{dlr_blue}{RGB}{27,120,170}
\definecolor{dlr_blue1}{RGB}{27,120,170}
\definecolor{dlr_blue2}{RGB}{97,154,191}
\definecolor{dlr_blue3}{RGB}{157,197,222}
\definecolor{dlr_blue4}{RGB}{204,225,238}
\definecolor{dlr_blue5}{RGB}{230,241,247}

\definecolor{dlr_red}{RGB}{190,54,81}
\definecolor{dlr_red1}{RGB}{190,54,81}
\definecolor{dlr_red2}{RGB}{213,90,107}
\definecolor{dlr_red3}{RGB}{235,129,136}
\definecolor{dlr_red4}{RGB}{245,206,206}
\definecolor{dlr_red5}{RGB}{250,230,230}

\definecolor{cb3_blue}{RGB}{128,177,211}
\definecolor{cb3_cyan}{RGB}{141,211,199}
\definecolor{cb3_yellow}{RGB}{255,255,179}
\definecolor{cb3_magenta}{RGB}{190,186,218}
\definecolor{cb3_red}{RGB}{251,128,114}
\definecolor{cb3_orange}{RGB}{253,180,98}
\definecolor{cb3_green}{RGB}{179,222,105}

\definecolor{cb_red}{RGB}{228,26,28}
\definecolor{cb_blue}{RGB}{55,126,184}
\definecolor{cb_green}{RGB}{77,175,74}
\definecolor{cb_magenta}{RGB}{152,78,163}
\definecolor{cb_orange}{RGB}{255, 127,0}
\definecolor{cb_yellow}{RGB}{255,255,51}
\definecolor{cb_brown}{RGB}{166,86,40}

\definecolor{mv_blue}{RGB}{134,164,194}
\definecolor{mv_cyan}{RGB}{136,209,203}
\definecolor{mv_light}{RGB}{191,211,224}
\definecolor{mv_brown}{RGB}{222,192,136}
\definecolor{mv_orange}{RGB}{247,177,0}
\definecolor{mv_pink}{RGB}{222,191,245}
\definecolor{mv_red}{RGB}{247,91,101}
\definecolor{mv_yellow}{RGB}{255,255,199}

\tikzstyle{block} = [draw=black, fill=gray!10, rectangle, minimum height=2.0em, minimum width=3em]
\tikzstyle{computer} = [draw=mv_light!50!black, fill=mv_light, rectangle, minimum height=2.0em, minimum width=10em,inner sep=5pt]
\tikzstyle{sensor}   = [draw=mv_orange!50!black, fill=mv_orange!50, rectangle, minimum height=2.0em, minimum width=8em,inner sep=5pt]
\tikzstyle{switch}   = [draw=mv_red!50!black, fill=mv_red!50, rectangle, minimum height=2.0em, minimum width=8em,inner sep=5pt]

\tikzstyle{sensor2}  = [draw=mv_yellow!50!black, fill=mv_yellow, rectangle, minimum height=3.0em, minimum width=9em,inner sep=5pt]
\tikzstyle{atom}  = [draw=mv_light!50!black, fill=mv_light, rectangle, minimum height=3.0em, minimum width=9em,inner sep=5pt]
\tikzstyle{gs1}   = [draw=mv_pink!50!black, fill=mv_pink, rectangle,   minimum height=3.0em, minimum width=9em,inner sep=5pt]
\tikzstyle{gs2}   = [draw=mv_orange!50!black, fill=mv_orange!50, rectangle, minimum height=3.0em, minimum width=9em,inner sep=5pt]
\tikzstyle{gsrt}  = [draw=mv_red!50!black, double, fill=mv_red!50, rectangle, minimum height=3.0em, minimum width=9em,inner sep=5pt]
\tikzstyle{state} = [draw=mv_orange!50!black, fill=mv_orange!50, rectangle, minimum height=3.0em, minimum width=8em]
\tikzstyle{block} = [draw=black, fill=gray!10, rectangle, minimum height=2.0em, minimum width=3em]
\tikzstyle{blocks} = [draw=blue!50!black, fill=blue!5, rectangle, minimum height=2em, minimum width=3em]
\tikzstyle{nlblock} = [draw=black, fill=gray!10, rectangle, minimum height=2em, minimum width=2em, double distance=1pt]
\tikzstyle{sum} = [draw, fill=gray!10, circle, node distance=1cm]
\tikzstyle{input} = [coordinate]
\tikzstyle{output} = [coordinate]
\tikzstyle{disturb} = [coordinate]
\tikzstyle{measur} = [coordinate]
\tikzstyle{pinstyle} = [pin edge={to-,thin,black}]
\tikzstyle{okvir} = [fill=green!15, rectangle,
    minimum height=8em, minimum width=15em]

\tikzstyle{branch} = [outer sep=0pt,draw=none,inner sep=0pt,fill=black,draw=none]

\tikzstyle{sum} = [draw,circle,inner sep=0mm,minimum size=0.5em,fill=white,sum color]
\tikzstyle{small sum} = [draw,circle,inner sep=0mm,minimum size=0.35em,fill=white,sum color]
\tikzstyle{branch} = [circle,inner sep=0pt,minimum size=1.5pt,fill=black,draw=black]
\tikzstyle{branch2} = [circle,inner sep=0pt,minimum size=2.5pt,fill=black,draw=black]

\newcommand{\inertia}{\boldsymbol{\mathcal{I}}}
\newcommand{\mass}{{\mathcal{M}}}

\newcommand{\norm}[1]{\ensuremath{\|{#1}\|}}

\newcommand{\vc}[1]{\boldsymbol{{#1}}}		

\newcommand{\Realv}[1]{\ensuremath{\mathbb{R}^{#1}}}	
\newcommand{\Realm}[2]{\ensuremath{\mathbb{R}^{{#1}\times{#2}}}}	
\newcommand{\K}{\mat{K\!}}			

\DeclareMathAlphabet{\mathsfsl}{OT1}{cmss}{m}{sl}	

\newcommand{\mat}[1]{\ensuremath{\boldsymbol{{#1}}}}		
\newcommand{\eye}[1]{\ensuremath{\boldsymbol{I}_{#1\times#1}}}		
\newcommand{\nil}[1]{\ensuremath{\boldsymbol{0}_{#1\times#1}}}		

\newcommand{\rand}{\textup{rand}}

\newcommand{\Ss}[1]{\ensuremath{\left({#1}\right)\!\times}}	

\newcommand{\vinf}{\vc{v}_{\infty}}
\newcommand{\vrel}{\vc{v}_{r}}

\newcommand{\fext}{\ensuremath{\vc{f}_{\!e}}}
\newcommand{\fint}{\ensuremath{\vc{f}_{\!i}}}

\newcommand{\fexthat}{\ensuremath{\hat{\vc{f}}_{\!e}}}
\newcommand{\finthat}{\ensuremath{\hat{\vc{f}}_{\!i}}}
\newcommand{\fdraghat}{\ensuremath{\hat{\vc{f}}_{\!d}}}

\newcommand{\tauext}{\ensuremath{\vc{\tau}_{\!e}}}
\newcommand{\tauexthat}{\ensuremath{\hat{\vc{\tau}}_{\!e}}}

\newcommand{\fig}[1]{Figure \ref{#1}}

\renewcommand{\journalname}{International Journal of Robotics Research (IJRR)}

\newcommand\BibTeX{{\rmfamily B\kern-.05em \textsc{i\kern-.025em b}\kern-.08em
T\kern-.1667em\lower.7ex\hbox{E}\kern-.125emX}}

\def\volumeyear{2018}

\begin{document}

\runninghead{Tomi\'{c} \emph{et al.}}
\title{Simultaneous Contact and Aerodynamic Force Estimation (s-CAFE) for Aerial Robots}

\author{Teodor Tomi\'{c}\affilnum{1,*}, %
        Philipp Lutz\affilnum{2}, %
        Korbinian Schmid\affilnum{3,*}, %
        Andrew Mathers\affilnum{4,$\dagger$} %
        and %
        Sami Haddadin\affilnum{5}
}

\affiliation{%
    \affilnum{1}
    Skydio, 114 Hazel Ave, Redwood City, CA 94061, USA%
    \\
    \affilnum{2}
    German Aerospace Center (DLR), Robotics and Mechatronics Center (RMC),
    M\"{u}nchner Stra{\ss}e 20, 82234 We{\ss}ling, Germany%
    \\
    \affilnum{3}
    Roboception GmbH,
    Kaflerstr. 2, 81241 Munich, Germany%
    \\
    \affilnum{4}
    Transportation Research Center,
    10820 State Route 347, East Liberty, OH 43319-0367, USA%
    \\
    \affilnum{5}
    Institut f\"{u}r Regelungstechnik,
    Gottfried Wilhelm Leibniz Universit\"at Hannover,
    Appelstr. 11, 30167 Hannover, Germany%
    \\
    \affilnum{*}
    The experimental part of the paper was done while these authors were at DLR
    \\
    \affilnum{$\dagger$}
    The experimental part of the paper was done while this author was at
    WindEEE Research Institute, 2535 Advanced Avenue, London, ON, Canada%
}

\corrauth{
    Teodor Tomi\'{c},
    Skydio, 114 Hazel Ave, Redwood City, CA 94061, USA
}

\email{teodor.tomic@gmail.com}

\begin{abstract}
In this paper, we consider the problem of multirotor flying robots physically interacting with the environment under wind influence.
The result are the first algorithms for simultaneous online estimation of contact and aerodynamic wrenches acting on the robot based on real-world data,
without the need for dedicated sensors.
For this purpose, we investigate two model-based techniques for discriminating between aerodynamic and interaction forces.
The first technique is based on aerodynamic and contact torque models, and uses the external force to estimate wind speed.
Contacts are then detected based on the residual between estimated external torque and expected (modeled) aerodynamic torque.
Upon detecting contact, wind speed is assumed to change very slowly. From the estimated interaction wrench, we are also able to determine the contact location.
This is embedded into a particle filter framework to further improve contact location estimation.
The second algorithm uses the propeller aerodynamic power and angular speed as measured by the speed controllers to obtain an estimate of the airspeed.
An aerodynamics model is then used to determine the aerodynamic wrench.
Both methods rely on accurate aerodynamics models.
Therefore, we evaluate data-driven and physics based models as well as offline system identification for flying robots.
For obtaining ground truth data we performed autonomous flights in a 3D wind tunnel.
Using this data, aerodynamic model selection, parameter identification, and discrimination between aerodynamic and contact forces could be done.
Finally, the developed methods could serve as useful estimators for interaction control schemes with simultaneous compensation of wind disturbances.
\end{abstract}


\maketitle

\newcommand\numberthis{\addtocounter{equation}{1}\tag{\theequation}}
\section{Introduction}
\subsection{Motivation and Contributions}
Autonomous multirotor flying robots with onboard visual navigation are being increasingly used in physical interaction scenarios.
Under real-world conditions such systems are also subject to local disturbances due to wind, and potentially also faults (e.g. collisions, loss of propellers).
The lumped external wrench (force and torque) $\vc{\tau}_{\!e}$ can therefore be interpreted as a sum of
the aerodynamic wrench $\vc{\tau}_{\!d}$, the physical interaction wrench $\vc{\tau}_{\!i}$, and a fault wrench $\vc{\tau}_{\!f}$,
such that ${\vc{\tau}_{\!e} = \vc{\tau}_{\!d} + \vc{\tau}_{\!i} + \vc{\tau}_{\!f}}$.
Therefore, robust operation under wind influence typically would require additional sensors to discriminate between these constituent wrenches.
In this paper, we develop novel model-based discrimination methods for solving this task that do not require application-specific sensors.

\begin{figure}
    \centering
    \includegraphics{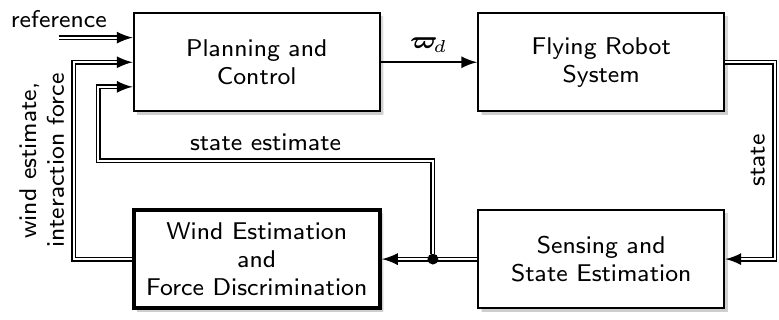}
    \caption{
        This paper focuses on wind estimation and discrimination between aerodynamic and interaction forces
        for autonomous flying robots.
        The figure depicts the overall relation in the high level overview.
    }
    \label{fig:introduction_overview}
\end{figure}

\begin{figure}
    \centering
    \includegraphics{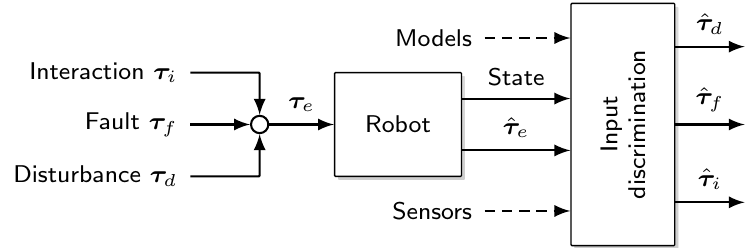}
    \caption{
        Input discrimination reconstructs the time-varying additive inputs summed in the external wrench $\vc{\tau}_{\!e}$,
        given an estimate thereof $\hat{\vc{\tau}}_{\!e}$, the robot's state, and additonal models and/or sensor inputs.
    }
    \label{fig:input_discrimination_intro}
\end{figure}

We therefore consider the problem of estimating the local wind speed $\vc{v}_w$ and discriminating between the components of the external wrench $\vc{\tau}_{\!e}$ acting on the flying robot.
We consider this ability as an additional component in the autonomy stack of flying robots, c.f. \fig{fig:introduction_overview}.
In particular, such estimators act as additional inputs into the planning and control stack.
For example, the estimated wind speed and could be used with an aerodynamics model to plan energy-efficient trajectories.
The estimated interaction force could be used for compliant interaction control, or for detecting contacts.
Additionally, the aerodynamic wrench may be compensated to provide zero steady-state tracking error even under wind influence.
The underlying input discrimination problem is depicted in \fig{fig:input_discrimination_intro}.
Given an estimate of the external wrench $\hat{\vc{\tau}}_{\!e}$, an input discrimination algorithm uses state information,
task-dependent models, and possibly additional sensors, to continuously reconstruct the constituent components of the external wrench.


\begin{figure}
    \centering
    \subfigure[Discrimination between aerodynamic and interaction force by using an onboard force sensor.]{
        \hspace{1.15cm}
        \includegraphics{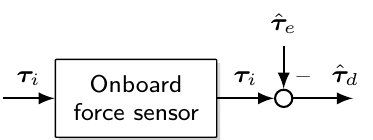}
        \hspace{1.15cm}
        \label{fig:force_discrimination_force_sensor}
        \label{fig:discrimination_methods_sensor}
    }
    \subfigure[Discrimination between aerodynamic and interaction force by onboard wind sensors.]{
        \includegraphics{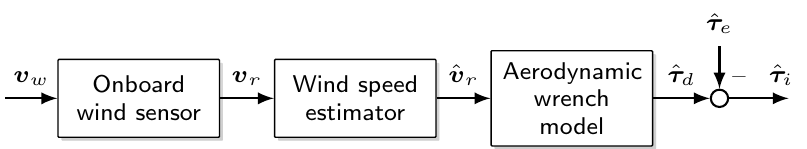}
        \label{fig:force_discrimination_wind_sensor}
        \label{fig:discrimination_methods_wind_estimation}
    }
    \subfigure[Discrimination between aerodynamic and interaction force by wind speed estimation .]{
        \includegraphics{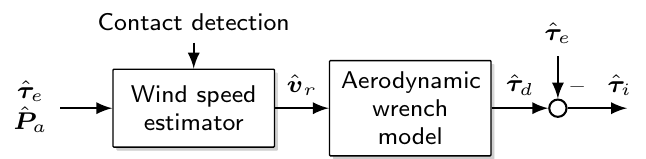}
        \label{fig:force_discrimination_wind_estimator}
        \label{fig:discrimination_methods_wind_estimator}
    }
    \caption{
        Overview of discrimination methods for slow aerodynamic and interaction wrenches.
        In Section \ref{sec:force_discrimination}, we develop model-based methods in the (c) family
        without using dedicated wind or force sensors.
        \label{fig:discrimination_methods}
    }
\end{figure}

\fig{fig:discrimination_methods} depicts three approaches to this problem.
In \fig{fig:force_discrimination_force_sensor}, an onboard force sensor at a known position is used.
The directly measured interaction force (or wrench in the case of a force-torque sensor), can simply be subtracted from $\hat{\vc{\tau}}_{\!e}$ to obtain the aerodynamic wrench.
The obtained interaction force is accurate, however it is localized to the sensor, which must also be carried onboard the vehicle.
This discrimination method has been used for interaction control scenarios in the literature.

An alternative approach is depicted in \fig{fig:force_discrimination_wind_sensor}.
Therein, an onboard wind sensor is used to obtain the relative airspeed $\vc{v}_r$.
This may be used in conjunction with an aerodynamic wrench model that maps the airspeed $\vc{v}_r$ to a wrench $\hat{\vc{\tau}}_{\!d}$.
The interaction wrench may then be obtained algebraically.
As of the writing of this paper, no small, lightweight, reliable and accurate wind sensors for multirotor vehicles exist yet,
however some significant advances have been made in this direction.
The interaction of the rotors with the surrounding airflow makes the calibration and placement of such sensors difficult.

Our first contribution is a family of novel model-based input discrimination methods depicted in \fig{fig:force_discrimination_wind_estimator}.
Similarly to \fig{fig:force_discrimination_wind_sensor}, we estimate wind speed and obtain the aerodynamic wrench from this estimate.
However, we also investigate several other methods to obtain an accurate wind speed estimate in Section \ref{sec:force_discrimination}.
The first one is established in literature, namely obtaining the airspeed $\vc{v}_r$ from the external wrench $\vc{\tau}_{\!e}$.
This scheme works only under the assumption that ${\vc{\tau}_{\!e} = \vc{\tau}_{\!d}}$.
To solve this issue, we develop a \emph{contact detection} algorithm based on an model that maps the aerodynamic forces to the aerodynamic torques.
Upon contact detection, wind estimation is essentially paused, changing the underlying assumption in the wind estimator.
This allows input discrimination when the wind speed is changing slowly compared to the interaction wrench.
The fundamental drawback of this approach is that both, interaction and aerodynamic wrenches, are estimated using the same sensor, namely the IMU.
Ideally, the wind speed would be provided by a measurement that is independent of $\vc{\tau}_{\!e}$, as would be the case with a dedicated wind sensor.
We therefore investigate the use of the propeller \emph{aerodynamic power} $P_a$ to estimate airspeed.
The approach is based on propeller aerodynamics, and measurement of the motor power (current and speed). Essentially, the propellers are then used as a wind sensor.
By finding the corresponding models, we are able to estimate the airspeed using the motor power.
Our second contribution is a method to estimate airspeed using aerodynamic power measurements only.
The underlying problem is analyzed in an optimization based approach, which is then enhanced by learned models.

\begin{figure}
    \centering
    \includegraphics[width=\columnwidth]{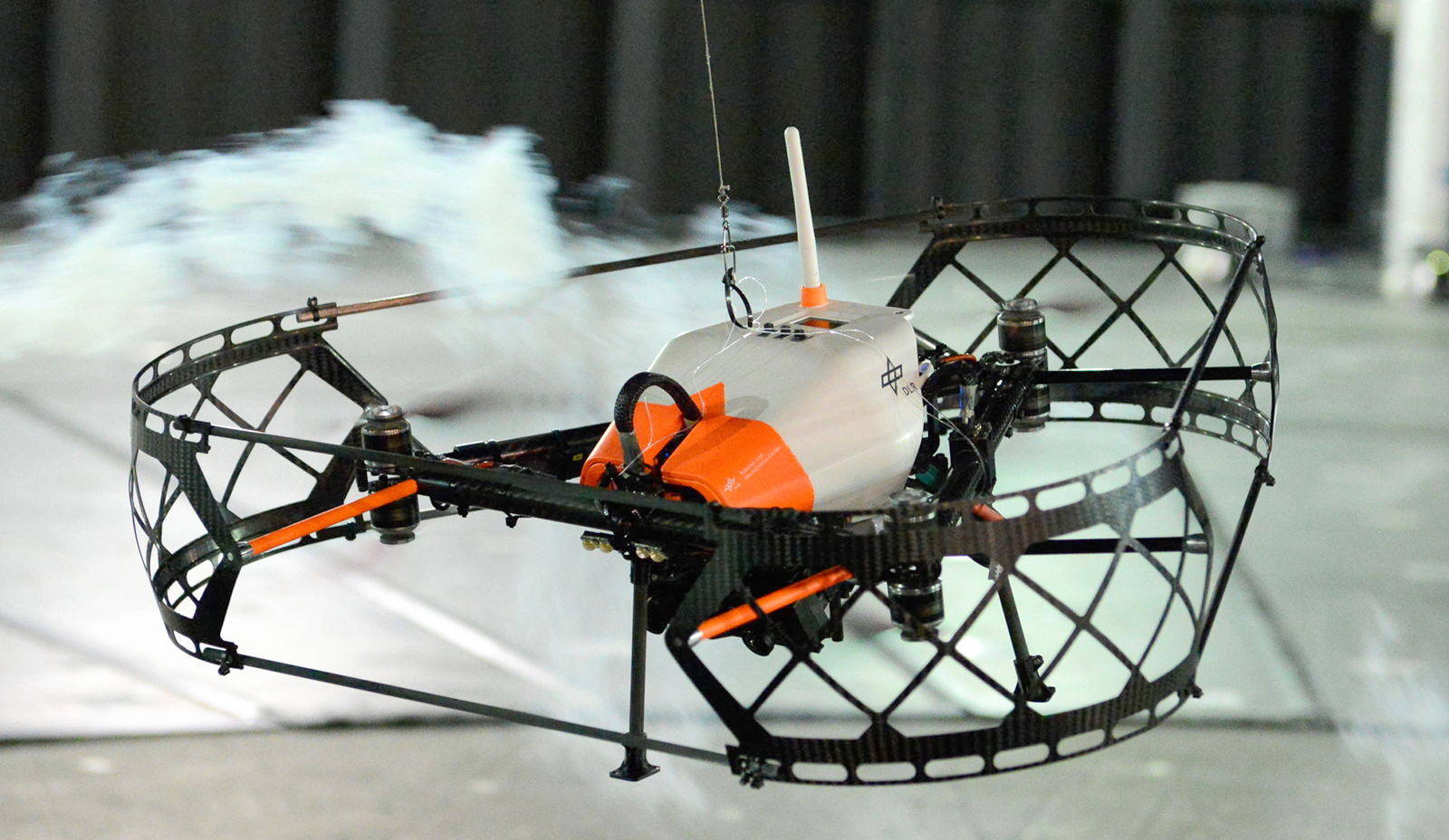}
    \caption{
        Experimental verification of the presented methods is done by flying a coaxial hexacopter in a 3D wind tunnel.
        The hexacopter hovers autonomously using a position controller, based on pose measurements from an external tracking system.
        For comparison, we also log data from a stereo vision based onboard pose estimator.
        The figure shows flow visualization at a wind velocity of around 6 m/s.
        We loosely suspended the robot on a filament for safety purposes.
    }
    \label{fig:ardea_in_flight}
\end{figure}

The fundamental challenge of model-based methods is obtaining relevant and accurate real-world applicable models.
For this purpose, we performed autonomous identification flight tests in a 3D wind tunnel, using a custom coaxial hexacopter, see Fig. \ref{fig:ardea_in_flight}.
Our third contribution is a detailed system identification procedure and a thorough analysis of the resulting aerodynamic models.
We take a data-driven approach to aerodynamics modeling, and analyze the effect of model complexity and inputs on airspeed reconstruction.
This may provide valuable guidance and insights to other researchers in the field.

The remainder of the paper is organized as follows.
Related work is presented in Section~\ref{sec:related_work}.
We give an overview of the complete system with dynamics, estimators, disturbance, and measured and estimated signals in Section~\ref{sec:preliminaries}.
The robot platform used in the experiments and simulations is described alongside our system identification approach in Section~\ref{sec:ardea},
followed by a description of the wind tunnel experiments in Section~\ref{sec:windeee}.
A description, performance analysis and generalization of aerodynamic models resulting from this data is presented in Section~\ref{sec:aero_model_evaluation}.
The optimization based approach to estimate airspeed from aerodynamic power is elaborated and compared to machine learning models in Section~\ref{sec:wind_from_power}.
Tha main contribution of this paper can be found int Section~\ref{sec:force_discrimination}, where we present
novel model-based methods for discrminating between aerodynamic and interaction wrenches based on the models identified in previous sections.
The section presents simulation results based on models obtained from experimental data.
The results also show how to incorporate the resulting wrenches into a physical interaction control scheme.
Finally, a conclusion is given in Section~\ref{sec:conclusion}.

\subsection{Related work}
\label{sec:related_work}

\begin{table*}
    \small \sf \centering
    \centering
    \caption{Overview of related work.
    Wind speed estimation for multirotor vehicles is a well established topic,
    and wind tunnel experiments are increasingly being carried out.
    Publications relating to applications of wind estimates show that there is a need for methods that provide such information.
    Force discrimination is still a rather unexplored topic.
    }
    \begin{tabularx}{\textwidth}{ l l }
        \toprule
        \textbf{Topic} &
        \textbf{References} \\
        \midrule
        \parbox[t]{10em}{\raggedright Model-based wind speed estimation} &
        \parbox[t]{\textwidth-13em}{\raggedright
        \cite{Waslander2009, Huang2009, Martin2010, Neumann2012, Omari2013, Abeywardena2014, Neumann2015, Tomic2015, Sikkel2016, Ware2016}
        }
        \vspace{0.75em}
        \\
        \parbox[t]{10em}{\raggedright Wind tunnel experiments} &
        \parbox[t]{\textwidth-13em}{\raggedright
        \cite{Marino2015, Schiano2014, Neumann2015, Planckaert2015, Jung2016, Prudden2016, Bruschi2016, Sikkel2016, Tomic2016}
        }
        \vspace{0.75em}
        \\
        \parbox[t]{10em}{Applications} &
        \parbox[t]{\textwidth-13em}{\raggedright
        \cite{Bangura2012, Sydney2013, Guerrero2013, Bangura2014, Ware2016, Bangura2017, Bennetts2017}
        }
        \vspace{0.75em}
        \\
        \parbox[t]{10em}{Force discrimination} &
        \parbox[t]{\textwidth-13em}{\raggedright
        \cite{Tomic2015, Manuelli2016, Rajappa2017, Tomic2017isrr}
        }
        \\
        \bottomrule
    \end{tabularx}
\end{table*}

\textbf{Model-based wind speed estimation}.
A large body of literature shows that it is possible to obtain the freestream velocity by using an aerodynamics model and onboard measurements in flight.
The methods fall broadly into two categories.
The first set uses a physical modeling based approach.
The exploited effects are usually blade flapping and propeller induced drag, which produce a horizontal force that can be measured by the onboard accelerometer.
The second set of methods uses a data-driven approach, where a regressor between a measured variable and the freestream velocity is found.

\emph{Physical modeling based estimation}.
\cite{Martin2010} showed that the blade flapping effect can be used to estimate the relative airspeed.
Metric wind velocity was also estimated by \cite{Waslander2009}, using a linear drag model and the propeller model from \cite{Huang2009}.
Estimation of the vertical wind component had limited accuracy, due to complexity of wind dependent thrust calculations and simplicity of the used model.
This paper shows the feasibility of model-based wind estimation.
More recently, \cite{Sikkel2016} used the blade flapping model, identified it by flying in a wind tunnel and used it for wind speed estimation.
Blade flapping and induced drag were incorporated as a feedforward term in a nonlinear control scheme by \cite{Omari2013}.
However, they assumed that the thrust is independent of the freestream velocity and that the wind velocity and yaw rates of the vehicle are negligible.
The dependence of thrust on freestream velocity was used by \cite{Huang2009} to add a feedforward term to a quadrotor position controller, in order to improve tracking performance during aggressive maneuvers.
\cite{Tomic2015} have shown how to incorporate this effect into external wrench estimation.
\cite{Abeywardena2014} estimate the horizontal wind velocity using accelerometer measurements, however the vertical wind component is not estimated.

\emph{Data-driven estimation}.
\cite{Neumann2012}, \cite{Neumann2015} and \cite{Ware2016} related the quadrotor pitch angle to the wind speed.
The assumption is that under static conditions the aerodynamic and control forces are in equilibrium,
hence the aerodynamic force is estimated indirectly from the position controller output.
As such, this method depends heavily on the controller and system parameters and is ill-suited to wind estimation during aggressive flight.

It can be concluded that model-based wind speed estimation is well established in literature, albeit with some limitations.

\textbf{Sensor-based wind speed estimation}.
Alternatively, airspeed probes have been used by \cite{Sydney2013}, \cite{Yeo2015}, and \cite{Yeo2016} to measure the freestream velocity of a quadcopter. They used this measurement to create a probabilistic map of the wind field and improve controller performance.
A successful evaluation of a small MEMS anemometers has also been performed by \cite{Bruschi2016}.
The anemometer was mounted 22 cm above the propellers of the quadrotor and tested in a wind tunnel. It performed well under different airflow conditions.
However, force sensors are discretely localized, and a reliable, lightweight wind sensor for multirotor UAV does not yet exist.

\textbf{Wind tunnel measurements}.
Wind tunnel measurements on multirotor vehicles have been increasingly carried out in recent years.
\cite{Schiano2014} and \cite{Planckaert2015} measured the forces and torques acting on a static quadrotor under varying conditions for model identification purposes.
Similarly, \cite{Jung2016} performed comprehensive wind tunnel tests on commercial multicopter vehicles, measuring the vehicle drag and thrust under varying conditions.
The purpose of the tests was to assess performance of multicopter systems.
\cite{Marino2015} measured the motor power in steady-state wind conditions, and tried to relate it to the wind velocity for estimation purposes.
They found that the mapping of power to wind velocity is not unique, and the solution quality varies with the flow conditions.
However, no online estimation scheme was proposed.
\cite{Ware2016} flew a quadcopter in a horizontal wind tunnel to identify the power used for flight under varying wind conditions and used it for path planning in an urban wind field.
\cite{Bruschi2016} evaluated the performance of a small anemometer mounted on a quadrotor.
\cite{Prudden2016} evaluated flow conditions around a quadcopter to find a feasible flow sensor mounting location where rotor influence is minimal.
They found that such a sensor would have to be mounted at least 2.5 rotor radii in front of the hub axis to compensate for induced flow effects.
\cite{Sikkel2016} flew a quadcopter in a wind tunnel to estimate an aerodynamic model based on blade flapping, and use it to estimate wind velocity.

\textbf{Applications.}
The estimated wind speed is commonly used in literature to improve control performance or plan time- or energy-optimal paths through a wind field.
\cite{Bangura2012} and \cite{Bangura2014} have shown that propeller power can be used to estimate and control its thrust.
They used momentum theory \citep{Leishman2006} to estimate and control the propeller aerodynamic power, which is directly related to thrust.
Therein, the estimated aerodynamic power has also been applied to estimate the propeller thrust when the freestream velocity was known.
Aerodynamic power control was applied to a quadrotor in \citep{Bangura2017} in order to improve flight control performance.
\cite{Sydney2013} used the estimated wind speed and an aerodynamics model to improve flight performance.
This was extended by an experimental validation in \citep{Sydney2015}.
\cite{Guerrero2013} used a kinematic model to plan time-optimal quadrotor trajectories in known wind fields.
\cite{Ware2016} planned energy-optimal trajectories in a planar urban wind field, which was estimated using a fast CFD solver and a known map.
\cite{Bennetts2017} use the onboard estimated wind speed for probabilistic air flow modeling.

\textbf{Discrimination between aerodynamic and contact forces}.
This topic has obtained the least attention so far.
In physical robot-environment interaction scenarios, it is common to assume the use of a dedicated force sensor \citep{Nguyen2013, Jung2012, Fumagalli2012, Alexis2016}.
If the external force is estimated, it is assumed that aerodynamic forces are non-existent or negligible during the interaction \citep{Fumagalli2013,Yuksel2014b,Tomic2014c}.
For handling collision scenarios, \cite{Briod2013} have used an Euler spring and force sensor to determine the collision force and direction.
Alternatively, \cite{Tomic2015} present two model-based methods to distinguish between aerodynamic and collision forces for collision detection and reaction.
They essentially use the fact that aerodynamic forces contain predominantly low frequencies, whereas collision forces contain also high frequencies.
In this way, proper collision reaction is possible even under wind influence.
However, neither of these methods can distinguish between \emph{slow} contact and aerodynamic forces from the \emph{estimated} external wrench.
More recently, \cite{Rajappa2017} proposed a discrimination scheme that employs a sensor ring around a flying robot to separate human interaction force from additional disturbances.
This scheme relies on adding localized sensors to the robot.
Related to this problem, \cite{Manuelli2016} implemented a Contact Particle Filter to obtain contact positions on an Atlas humanoid robot using external wrench information.
In \cite{Tomic2017isrr}, the authors apply the isolation pipeline developed in \cite{Haddadin2017} to force discrimination.
A simulation result is shown for force discrimination. It is based on minimizing the residual of the aerodynamic torque model to the external torque across
possible contact points on the convex hull.

In this paper, we further develop this principle in Section~\ref{sec:force_discrimination}.
Based on measurements from wind tunnel experiments, we provide the insight that the aerodynamic \emph{torque} may be modeled as a function of the aerodynamic force.
We use this model to design several novel schemes.
First, contact \emph{detection} may be done through a torque residual signal.
Second, it is possible to uniquely determine aerodynamic and contact forces at a point on the robot's convex hull.
By combining these two methods it is possible to compensate wind disturbances while also being compliant to contact forces
through the use of the compensated impedance controller proposed in \cite{Tomic2017}.
This is then extended into a particle filter framework that allows the fusion of additional information into the force discrimination problem.
Lastly, we use the aerodynamic power to estimate the wind speed, to directly obtain the aerodynamic wrench and discriminate aerodynamic and interaction wrenches.

\begin{figure*}
    \centering
    \includegraphics[width=\textwidth]{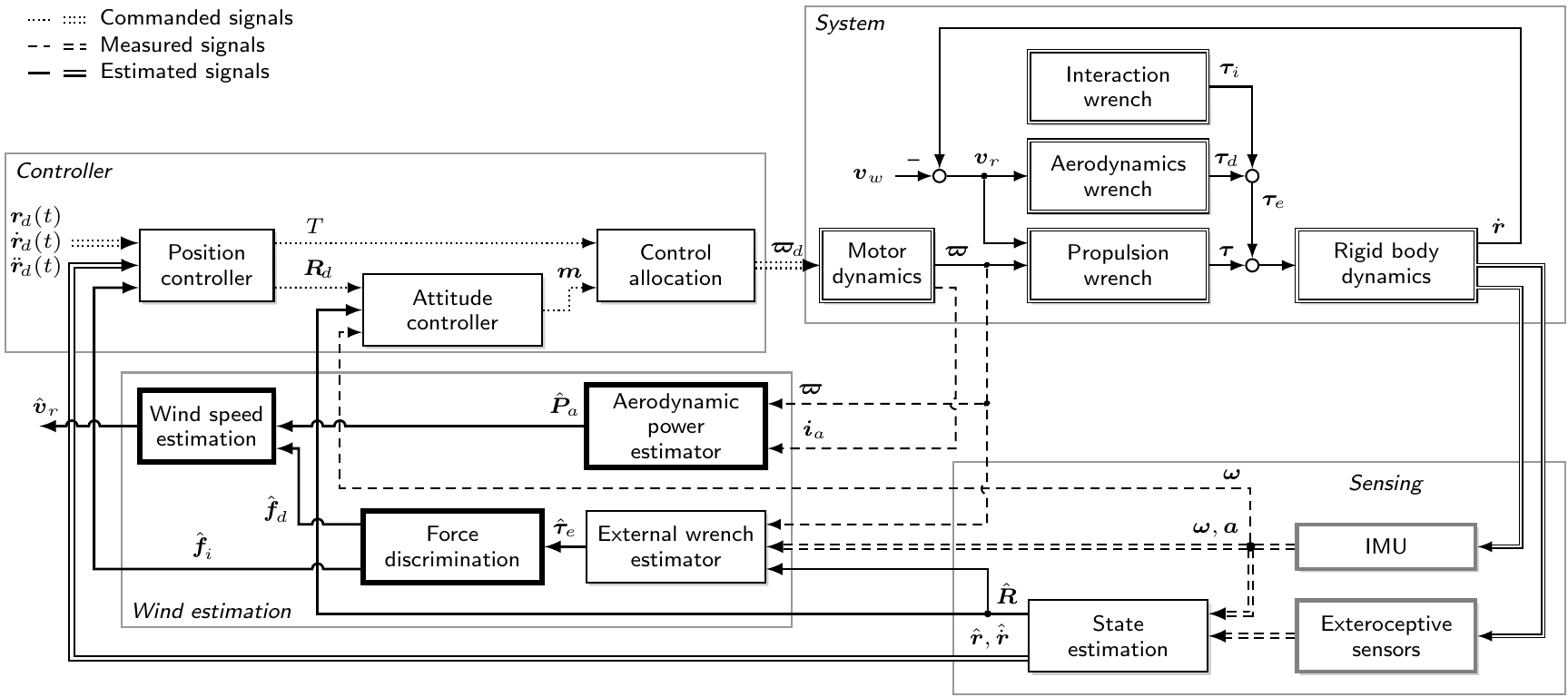}
    \caption{
        Overview of a mutirotor flying robot capable of estimating wind speed and discriminating between interaction and aerodynamic wrenches. The bold blocks are the focus of this paper.
        The \emph{aerodynamic power estimator} uses a propeller aerodynamics model and motor feedback to obtain the aerodynamic power $\hat{\vc{P}}_{\!a}$.
        \emph{Force discrimination} uses the external wrench, the robot's convex hull and an aerodynamics module to simultaneously
        estimate the interaction force $\vc{f}_{\!i}$ and the aerodynamic force $\vc{f}_{\!d}$.
        The aerodynamic force $\vc{f}_{\!d}$, along with $\hat{\vc{P}}_{\!a}$, may then used to perform \emph{wind speed estimation}.
        }
    \label{fig:system_overview_diagram}
\end{figure*}

\section{Preliminaries}
\label{sec:preliminaries}

\fig{fig:system_overview_diagram} depicts an overview of a mutirotor flying robot capable of estimating wind speed and discriminating between interaction and aerodynamic wrenches.
The \emph{System} consists of a rigid body, with the aerodynamics wrench $\vc{\tau}_{\!d}$, interaction wrench $\vc{\tau}_{\!i}$ and propulsion wrench $\vc{\tau}$ acting on it through the external wrench $\vc{\tau}_{\!e}$.
The rgid body dynamics are reviewed in Section~\ref{sec:rigid_body}, the propulsion wrench is described in Section~\ref{sec:propulsion_wrench}, along with relevant propeller aerodynamics in Section~\ref{sec:propeller_aerodynamics}.
The external wrench estimation is discussed in Section~\ref{sec:wrench_estimation}.
The propulsion wrench and motor dynamics are connected through the propeller speed $\vc{\varpi}$.
The desired rotor speed $\vc{\varpi}_{\!d}$ is commanded by the \emph{Controller}.
This is typically a cascade of position and attitude control.
Alternatively, geometric controllers or model predictive control may be used.
The \emph{Sensing} modules provide the controller with an estimate of the vehicle's state (position, orientation, velocity).
We use this, and the motor feedback, in the \emph{Wind estimation} blocks.
The \emph{aerodynamic power estimator} uses a propeller aerodynamics model and motor feedback to obtain the aerodynamic power $\hat{\vc{P}}_{\!a}$,
as described in Section~\ref{sec:motor_model}.
\emph{Force discrimination} uses the external wrench, the robot's convex hull and an aerodynamics module to simultaneously
estimate the interaction force $\vc{f}_{\!i}$ and the aerodynamic force $\vc{f}_{\!d}$, as described in Section~\ref{sec:force_discrimination}.
The interaction force $\vc{f}_{\!i}$ may then be used for interaction control, see Section~\ref{sec:compensated_impedance_control}.
The aerodynamic force $\vc{f}_{\!d}$, along with $\hat{\vc{P}}_{\!a}$, may allow \emph{wind velocity estimation}, see in Section~\ref{sec:aero_model_evaluation}.

\subsection{Rigid body dynamics}
\label{sec:rigid_body}
\begin{figure}
    \centering
        \includegraphics[scale=1.0]{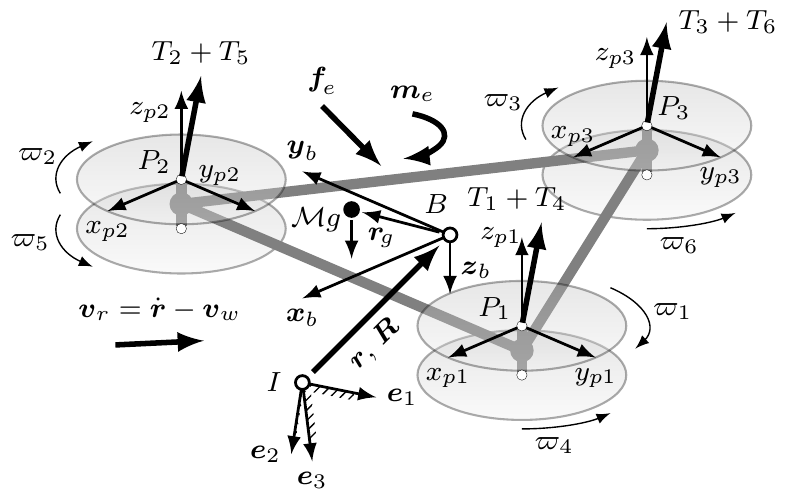}
    \caption{Free-body diagram of the coaxial hexacopter used in the experiments.
    The body frame $B$ is located at position $\vc{r}$
    and orientation $\mat{R}$ in the inertial frame $I$ and
    subject to wind velocity $\vc{v}_w$.
    This causes the external wrench $\vc{\tau}_e = [\vc{f}_{\!e}^T\; \vc{m}_e^T]^T$
    due to aerodynamic forces that in turn depend on the airspeed $\vc{v}_r$.
    The propellers rotate at angular velocities
    $\vc{\varpi} = [\varpi_1 \, \ldots \, \varpi_6]^T$, and
    generate the control wrench ${\vc{\tau} = [\vc{f}^T\; \vc{m}^T]^T}$ through
    thrusts $T_i$ and drag torques $Q_i$.
    The propeller frames $P_{1,2,3}$ are depicted in blue.
    }
    \label{fig:quadrotor}
\end{figure}

A free-body diagram of a coaxial hexacopter is depicted in \fig{fig:quadrotor}.
The equations of motion about the center of mass can be written as
\begin{align}
    \mass \ddot{\vc{r}} &=  \mass g \vc{e}_3 + \mat{R} \vc{f} + \mat{R} \vc{f}_e \\
    \mathcal{I} \dot{\vc{\omega}} &= \Ss{\mathcal{I} \vc{\omega}} \vc{\omega}
    + \vc{m} + \vc{m}_e \\
    \dot{\mat{R}} &= \mat{R} \Ss{\vc{\omega}}
    \label{eq:dynamics}
\end{align}
where $\mass$ is the robot mass,
${\vc{r} = [x,\, y,\, z]^T}$ is its position in the fixed North-East-Down (NED) inertial frame,
${\mat{R} \in SO(3)}$ is the rotation matrix from the body to the inertial frame,
${\mathcal{I} \in \Realm{3}{3}}$ is its moment of inertia,
$(\cdot)\times$ is the skew-symmetric matrix operator,
$g$ is the acceleration of gravity,
$\vc{\omega}$ is the body angular velocity,
$\vc{e}_3$ is the $z$-axis unit vector,
${\vc{f}}$ and ${\vc{f}_{\!e}}$ are the body-frame control and external forces,
and ${\vc{m}}$ and $\vc{m}_e$ are the control and external torques, respectively.
We denote the control wrench as ${\vc{\tau} = [\vc{f}^T\; \vc{m}^T]^T}$,
and the external wrench as ${\vc{\tau}_{\!e} = [\vc{f}_{\!e}^T\; \vc{m}_e^T]^T}$.
In our case, the body frame is located at the center of propellers.
To account for the offset of the center of mass, the control torque $\vc{m}$ contains the correction term
${\vc{m}_g = \mass g \Ss{\vc{r}_g} \vc{R}^T \vc{e}_3}$,
where $\vc{r}_g$ is the position of the center of gravity expressed in the body frame.

\subsection{Propulsion wrench}
\label{sec:propulsion_wrench}
Knowing the applied control wrench $\vc{\tau}$ is required for external wrench estimation.
It is usually obtained by a cascaded control structure, see \fig{fig:system_overview_diagram}.
For details on basic control of flying robots, we refer the reader to
\citep{Lee2010,Hoffmann2008,Hua2013,Tomic2014b}.
As $\vc{\tau}$ can not be directly measured during flight, an accurate model is required.
The control wrench generated by the propellers about the center of mass for $N$ propellers can be estimated by
\begin{equation}
    \vc{\tau} = \begin{bmatrix}
        \sum_{i=1}^N T_i \vc{n}_i \\
        \sum_{i=1}^N \left( T_i (\vc{r}_i + \vc{r}_g) \times \vc{n}_i + \delta_i Q_i \vc{n}_i \right)
    \end{bmatrix}
    =
    \mat{B} \vc{u}
    \label{eq:propulsion_wrench}
\end{equation}
where ${\vc{n}_i = \mat{R}_{bp,i} \vc{e}_3}$ is the axis of rotation of propeller $i$ located at $\vc{r}_i$
in the body frame,
$\mat{R}_{pb,i}$ the rotation matrix from the body to the propeller frame,
${\delta_i \in \{-1,1\}}$ is the propeller rotational sense,
and $\mat{B} \in \Realm{6}{N}$ the control allocation matrix.
In classical designs where propellers are coplanar, $\mat{R}_{bp,i}$ is an identity matrix.
In more recent multirotor designs, propellers are tilted to obtain more yaw control authority.
In those cases, $\mat{R}_{bp,i}$ will be the appropriate transformation.
The desired propeller velocities ${\vc{u} = [\varpi_1^2 \ldots \varpi_N^2]^T}$
can be obtained for control purposes by (pseudo-)inverting the matrix $\mat{B}$.
The rotor thrust and torque in hover may be obtained as
\begin{equation}
    T_{h,i} = \rho \, C_T D^4 \varpi_i^2,
    \label{eq:thrust}
\end{equation}
\begin{equation}
    Q_{h,i} = \rho \, C_Q D^5 \varpi_i^2 + I_r \dot{\varpi}_i,
    \label{eq:torque}
\end{equation}
where $C_T$ and $C_Q$ are nondimensional rotor thrust and torque coefficients, respectively.
They are typically obtained from static thrust measurements;
$D$ is the propeller diameter, and $\varpi$ is the propeller speed.
Additionally, $\rho$ is the air density,
$D$ the propeller diameter,
and $I_r$ the combined inertia of rotor and propeller.
Next, we show how the thrust and torque change under wind influence.
From aformationed values, $D$ is known from geometry and $\varpi$ is measured online.
Note that the thrust and torque change with airspeed, i.e. \eqref{eq:thrust} and \eqref{eq:torque} are only valid during hover conditions.
The air density ${\rho = p / (R_{\mathrm{air}} T)}$ depends on air pressure $p$, absolute temperature $T$ and specific gas constant of air $R_{\mathrm{air}} = 287.05$ J/kg\,K,
and may be estimated online or computed from onboard barometer measurements.
This leaves $C_T$, $C_Q$, and $\mathcal{I}_r$ to be identified.
The coefficients $C_T$ and $C_Q$ are commonly obtained from bench tests relating propeller angular speed to measured force and torque.
The rotor inertia $\mathcal{I}_r$ may be obtained by inertia estimation methods like swing tests, or dynamic identification on a separate motor.
In Section \ref{sec:parameter_identification}, we identify these parameters on a force-torque sensor in the full hexacopter configuration.

\subsection{Propeller aerodynamics}
\label{sec:propeller_aerodynamics}
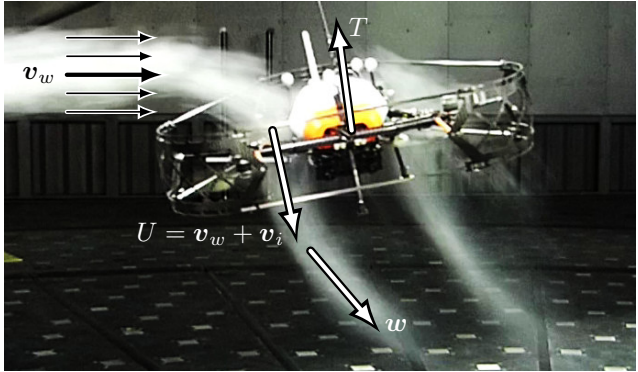
\begin{figure}
    \centering
    \begin{tikzpicture}[>=latex]
        \node[inner sep=0pt, anchor=south west,opacity=1.0] (image) at (0,0)
        {\includegraphics[width=\columnwidth]{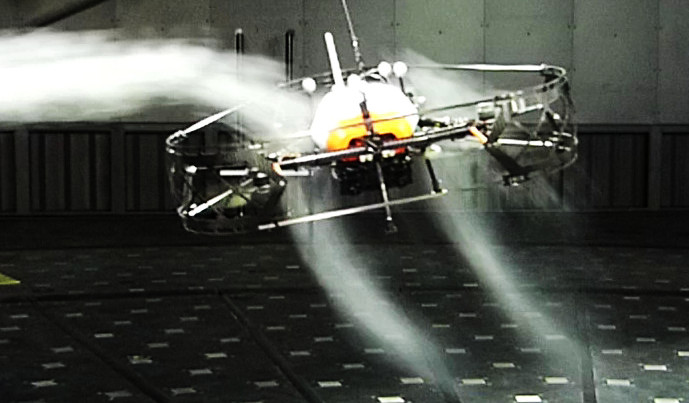}};

    \begin{scope}[x={(image.south east)}, y={(image.north west)}, >=latex, draw=black, ultra thick, line cap=round]
            \draw[->,white,line width=3.0pt] (0.1,0.8) -- +(0.163,0) ;
            \draw[->] (0.1,0.8) node[left]{$\vc{v}_{w}$} -- +(0.15,0) ;

            \draw[->,white,ultra thick] (0.1,0.70) -- +(0.144,0) ;
            \draw[->,semithick] (0.1,0.70) -- +(0.135,0) ;

            \draw[->,white,ultra thick] (0.1,0.75) -- +(0.144,0) ;
            \draw[->,semithick] (0.1,0.75) -- +(0.135,0) ;

            \draw[->,white,ultra thick] (0.1,0.85) -- +(0.144,0) ;
            \draw[->,semithick] (0.1,0.85) -- +(0.135,0) ;

            \draw[->,white,ultra thick] (0.1,0.90) -- +(0.144,0) ;
            \draw[->,semithick] (0.1,0.90) -- +(0.135,0) ;

            \draw[->,line width=3.3pt] (0.425,0.65) -- +(0.039,-0.305);
            \draw[->,white] (0.425,0.65) -- +(0.036,-0.28) node[left,white]{$U=\vc{v}_{w} + \vc{v}_i$} ;

            \draw[->,line width=3.3pt] (0.485,0.325) -- +(0.11,-0.22);
            \draw[->,white] (0.485,0.325) -- +(0.1,-0.2) node[right,white]{$\vc{w}$} ;

            \draw[->,line width=3.3pt] (0.55,0.65) -- +(-0.025, 0.308);
            \draw[->,white,ultra thick] (0.55,0.65) -- +(-0.022, 0.28) node[white,right]{$T$};
        \end{scope}
    \end{tikzpicture}
    \caption{Thrust $T$ is generated by increasing the wind velocity $\vc{v}_{w}$ by
            the propeller induced velocity $v_i$, which passes through the propeller in normal direction.
            The propeller slipstream finally merges into the wind flow to produce $\vc{w}$.
        }
    \label{fig:thrust_generation}
\end{figure}

\begin{figure}
\centering
    \begin{tikzpicture}[rounded corners=5pt]
        \node[inner sep=0pt, anchor=south west,opacity=1.0] (image) at (0,0)
        {\includegraphics{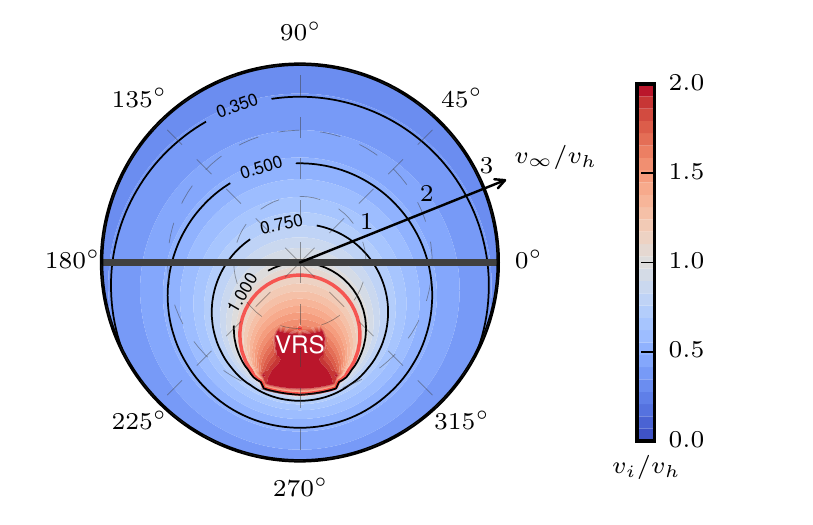}};

        \begin{scope}[x={(image.south east)}, y={(image.north west)}, >=latex', draw=red, ultra thick, red]
            \draw[->, thick, black] (0.1, 0.7)  -- (0.15, 0.65) node[at start,above,rotate=66]{{\sffamily \scriptsize Propeller plane}} -- (0.2, 0.5);
        \end{scope}
    \end{tikzpicture}
    \caption{Relative induced velocity $v_i/v_h$ in forward flight, depending on angle of attack
        and relative airspeed $v_{\infty}/v_h$, depicted on the radial axis.
        Unmodified momentum theory is invalid in area delineated in red,
        where the propeller is in the Vortex Ring State (VRS) \citep{Leishman2006}.
        The contour lines depict different values of ${v_i/v_h}$.
    }
    \label{fig:induced_velocity_validity}
\end{figure}

The forces exerted by a propeller depend on its freestream velocity (relative wind velocity).
The freestream velocity of the $k$-th propeller expressed in the propeller frame is
\begin{equation}
    \vc{v}_{\infty,k} = \mat{R}_{bp,k}^T \left(
        \mat{R}^T \vrel
        + \vc{\omega} \times \vc{r}_k
    \right)
    ,
\end{equation}
where
${\vrel=\dot{\vc{r}} - \vc{v}_w}$ is the true airspeed,
$\vc{v}_w$ is the wind velocity,
and
$\vc{r}_k$ is the location of the propeller relative to the center of gravity.
The thrust acts in positive $z$-direction of the propeller frame $P_k$, see \fig{fig:quadrotor}.
According to momentum theory \cite{Leishman2006} it can be written as
\begin{equation}
    T = 2 \rho A v_i U
    ,
    \label{eq:thrust_momentum_theory}
\end{equation}
where $A$ is the rotor disk surface area,
and ${U = \| v_i \vc{e}_3 + \vinf \|}$ is the velocity of the propeller slipstream.
The induced velocity $v_i$ can be obtained using 
\begin{equation}
    v_i = \frac{v_h^2}{\sqrt{v_{xy}^2 + (v_i - v_z)^2}},
    \label{eq:induced_velocity}
\end{equation}
which may be solved by several Newton-Raphson iterations with known $v_h$ and $v_{\infty}$ \cite{Leishman2006}.
A flow visualization of thrust generation and the relevant velocities are depicted in \fig{fig:thrust_generation}.
The horizontal and vertical components of the freestream velocity are
${\vc{v}_{xy} = \vc{v}_{\infty} - \vc{v}_z}$ and
${\vc{v}_{z} = \vc{e}_3^{T} \vc{v}_{\infty}}$, respectively.
Their norms are
${v_{xy} = \| \vc{v}_{xy} \|}$ and
${v_z = \| \vc{v}_z \|}$.
In hover conditions the induced velocity is ${v_h = \sqrt{{T_h}/{2 \rho A}}}$,
where the hover thrust is obtained from \eqref{eq:thrust}.
The ideal aerodynamic power of a propeller is
\begin{equation}
    P_a = 2 \rho A v_i U (v_i - v_z).
    \label{eq:aerodynamic_power}
\end{equation}
Furthermore, the aerodynamic power in forward flight is related to the hovering power following
\begin{equation}
    \frac{P_a}{P_h} = \frac{v_i - v_z}{v_h},
    \label{eq:aerodynamic_power_hover}
\end{equation}
with ${P_h = 2 \rho A v_h^3}$.
Nonidealities can be included through the figure of merit $F\!M \in [0\ldots 1]$,
such that ${P_a = P_m F\!M}$, where $P_m$ denotes the motor power.
The theory must be applied in the valid domain.
Unmodified momentum theory does not apply in the unsteady Vortex Ring State (VRS) \cite{Leishman2006},
as depicted in \fig{fig:induced_velocity_validity}.

Note that we may also model the aerodynamic power $P_a$ as a general function of $P_m$, ${P_a := g(P_m)}$ which can
be found in a parameter identification step. This will also include effects neglected by first-principles physics modeling,
such as motor losses or fluid-structure interaction effects.

\textbf{Coaxial rotors.} In the case of coaxial rotors, we may consider one propeller pair as a single propeller.
In that case, the hover induced velocity becomes
\begin{equation}
    v_{h} = \sqrt{\frac{T_1 + T_2}{2 \rho A}} = 
    D \sqrt{\frac{2 }{\pi}} \sqrt{C_{T,1} \varpi_1^2 + C_{T,2} \varpi_2^2},
    \label{eq:coaxial_vh}
\end{equation}
where $C_{T,1}$ and $C_{T,2}$ are thrust coefficients of the upper and lower propeller, respectively.
We obtain them using an identification procedure of the propulsion system, see Section \ref{sec:parameter_identification}.
Having obtained $v_h$, we use other aerodynamic quantities as described above.
Note that the torque coefficients (and thereby power) will also differ between the upper and lower propeller.

\subsection{External wrench estimation}
\label{sec:wrench_estimation}

For later use we briefly revisit the external wrench estimator from \cite{Tomic2014c}.
The estimator uses the control input, a system model and proprioceptive sensors only.
The external wrench estimate
${\hat{\vc{\tau}}_{\!e} = [\,\hat{\!\vc{f}}_{\!e}^{T}\; \hat{\vc{m}}_e^{T}]^T}$
is obtained from
\begin{equation}
        \hat{\vc{\tau}}_{\!e} \! =
    \!
    \begin{bmatrix}
        \int \mat{K}_I^f
        \left(
                        m\vc{a} - \vc{f} - \hat{\!\vc{f}}_{\!e}
        \right)
        \mathrm{d}t
        \\
    \mat{K}_I^m
        \left(
            \mathcal{I} \vc{\omega}
            -
            \int_0^t
            \left(
                \vc{m}
                + (\mathcal{I}\vc{\omega}) \times \vc{\omega}
                - \hat{\vc{m}}_e
            \right)
            \mathrm{d} t
        \right)
    \end{bmatrix}
    \label{eq:wrench_estimator}
\end{equation}
where $\mat{K}_I^f$ and $\mat{K}_I^m$ are filter gains, and
${\vc{a} = \mat{R}^T(\ddot{\vc{r}} - g\vc{e}_3)}$ is the acceleration
measured by an accelerometer in the center of mass and expressed in the body frame.
$\hat{\!\vc{f}}_{\!e}$ and $\hat{\vc{m}}_e$ are the estimated external force and torque,
also expressed in the body frame.
Note that $\vc{m}$ contains the term compensating the offset center of gravity.
The estimation dynamics are shown to be ${(s + K_I) \hat{\vc{\tau}}_{\!e} = K_i \vc{\tau}_{\!e}}$.
In contrast to e.\,g. \cite{Ruggiero2014} and \cite{Yuksel2014b}, this estimator does not require translational velocity measurements.
For the control input $\vc{\tau}$ we assume hover conditions, and obtain the control thrust and torque using \eqref{eq:thrust} and \eqref{eq:torque}.
Therefore, the estimator will also capture the modeling errors.
Any aerodynamic models identified using this estimate will implicitly also capture the change of propeller thrust and torque with airspeed.

\subsection{Reduced brushless DC motor model}
\label{sec:motor_model}
In order to estimate the propeller aerodynamic power, we employ the BLDC motor model
from \cite{Bangura2014}.
The mechanical part of motor dynamics can be written as
\begin{align}
    \tau_m &= (K_{q,0} - K_{q,1} i_a) i_a,
    \label{eq:motor_torque}
    \\
    I_r \dot{\varpi} &= \tau_m - D_r,
    \label{eq:rotor_torque}
\end{align}
where $i_a$ is the current through the motor
and $\varpi$ the rotor angular velocity.
The motor torque is $\tau_m$, with the torque constant modeled as ${K_q(i_a) = (K_{q,0} - K_{q,1}i_a)}$.
The parameter $I_r$ is the rotor inertia, and $D_r$ the aerodynamic drag torque acting on the rotor.
The total motor mechanical power is
${P_m = {P_a}/{F\!M} + P_r}$,
where the mechanical power $P_m$ and power consumed by rotor acceleration $P_r$ are
used to estimate the aerodynamic power using
\begin{align}
    P_m &= \tau_m \varpi = (K_{q,0} - K_{q,1}i_a) i_a \varpi,
    \label{eq:motor_power} \\
    P_r &= I_r \varpi \dot{\varpi},
    \label{eq:rotor_power} \\
    \hat{P}_a 
    &= F\!M \Bigl( (K_{q,0} - K_{q,1}i_a) i_a - I_r \dot{\varpi} \Bigr) \varpi
    .
\end{align}
Note that, in general, the figure of merit $FM$ can be a nonlinear function.
In summary, we need to estimate or measure the motor current $i_a$,
rotor speed $\varpi$ and rotor acceleration $\dot{\varpi}$.
The measurements $i_a$ and $\varpi$ can be obtained from modern electronic speed controllers (ESC),
and $\dot{\varpi}$ may be estimated, see \cite{Bangura2014}.

Current measurement on the speed controller allows to directly relate motor torque \eqref{eq:motor_torque} to
propeller aerodynamic torque \eqref{eq:torque} through \eqref{eq:rotor_torque}, leading to
\begin{equation}
    \begin{aligned}
        \tau_{m} &= Q, \\
        (K_{q,0} - K_{q,1} i_a) i_a &= I_r \dot{\varpi} + \rho \, C_Q D^5 \varpi_i^2,
    \end{aligned}
    \label{eq:current_torque}
\end{equation}
i.e. the motor current provides a direct measurement of the torque applied to the rotor without the need of numerical
differentiation to obtain $\dot{\varpi}$. By using this measurement to get the propulsion wrench \eqref{eq:propulsion_wrench},
and using it in the external wrench estimator \eqref{eq:wrench_estimator}, we obtain an accurate estimate of the external yaw
torque even under wind influence.
We further investigate this relation in Section \ref{sec:parameter_identification}.

\subsection{Aerodynamics compensated impedance control}
\label{sec:compensated_impedance_control}
Discrimination between aerodynamic and contact forces is a prerequisite for the
\emph{aerodynamics compensated impedance controller}, see \cite{Tomic2017}.
The main idea is to compensate the aerodynamic disturbance,
and control an impedance w.r.t. the estimated mechanical interaction force.
This allows for rendering the right impedance behavior even under complex disturbance
conditions.
The desired error dynamics
\begin{equation}
    \mass_v \ddot{\tilde{\vc{r}}}
    + \K_{d} \dot{\tilde{\vc{r}}}
    + \K_{p} \tilde{\vc{r}}
    = \fint
    \label{eq:position_error_dynamics}
\end{equation}
results in the position control law
\begin{align}
    \mat{R}_d \vc{f} &=
    \mass \ddot{\vc{r}}_d
        - \fdraghat
    + (\mu - 1)\, \finthat
    - \mu \! \left(
        \K_d \dot{\tilde{\vc{r}}}
        + \K_p \tilde{\vc{r}}
    \right)
    - \vc{g}
    ,
    \label{eq:position_compensated_impedance_controller}
\end{align}
where $\mat{R}_d$ is the desired orientation,
$\fdraghat$ is the estimated aerodynamic force
$\finthat$ is the estimated interaction force, and
$\ddot{\vc{r}}_d$, $\dot{\tilde{\vc{r}}}_d$, and $\tilde{\vc{r}}$ are
the desired acceleration, the velocity and position errors, respectively.
The derivative and proportional gains are $\mat{K}_{\!d}$ and $\mat{K}_{\!p}$, respectively.
The constant $\mu = \mass \mass_v^{-1}$ contains the desired virtual mass of the system $\mass_v$.
We do not assume any properties on $\fdraghat$ and $\finthat$ other than
${\fexthat = \fdraghat + \finthat}$.
The actual error dynamics of this controller will be influenced by the quality of discrimination
between $\fdraghat$ and $\finthat$.

\section{Flying robot platform}
\label{sec:ardea}
\subsection{System overview}
\begin{figure}
    \centering
    \includegraphics[width=.9\columnwidth]{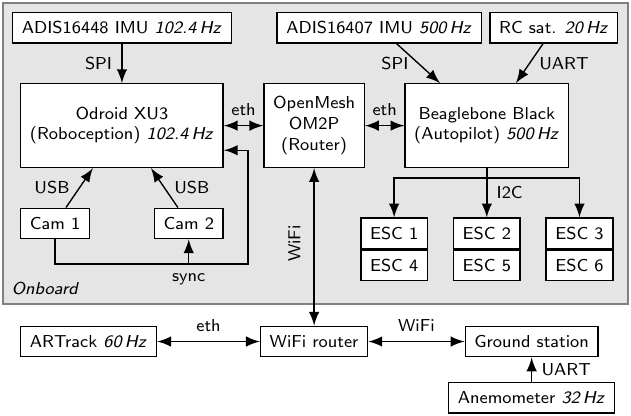}
    \caption{
        Experiment hardware setup.
        The autopilot is a Beaglebone Black running a Linux with realtime patch.
        The Odroid XU-3 is used as the onboard pose estimator, with a separate IMU and triggered stereo cameras.
        They are connected through an onboard wireless router, which also serves as a connection to the ground station.
    }
    \label{fig:hardware_setup}
\end{figure}

\textbf{Hardware.}
Flight experiments in the wind tunnel were carried out using a custom-built hexacopter in a coaxial configuration, depicted in \fig{fig:ardea_in_flight}.
For propulsion we used six identical T-Motor MN2212 920 Kv motors at a 4S voltage (14.8V), with T-Motor CF 10x3.3 propellers.
The computational hardware setup is outlined in \fig{fig:hardware_setup}.
A Beaglebone Black with realtime Linux is running the autopilot software, including the controller.
We use ESC32v2 speed controllers with firmware modified to support I2C communication.
The ESCs measure the motor current using a shunt resistor at the power supply pads.
An Odroid XU3 is used for onboard pose estimation by fusing stereo visual odometry and IMU data.
The two PointGrey Firefly cameras used for visual odometry are synchronized with a hardware trigger.
Communication between the two onboard computers is achieved over Ethernet.
The wireless router is also used as connection with the groundstation.

\textbf{Software}.
The orientation strapdown, attitude and position controllers run on the autopilot computer at 500\,Hz.
The motor feedback is obtained at 1/6th of the control rate.
Pose updates are sent to the autopilot either from the groundstation (motion capture system) or the Odroid (onboard pose estimation).
The received poses are used for position and attitude control.
Communication between components is done using a custom real-time capable middleware, over shared memory and UDP.
For onboard pose estimation
we employ a Roboception \cite{roboception}
Navigation Sensor which exposes the full 6D pose of the robot, its
velocities as well as unbiased IMU measurements.
The state estimator fuses keyframe delta
poses of a stereo vision odometry with IMU measurements by an indirect,
extended Kalman filter while latencies of the vision system are
compensated \cite{schmid2014local}. The whole pipeline runs on the
on-board Odroid XU3 computer.

\subsection{Parameter identification}
\label{sec:parameter_identification}

\begin{figure*}
    \centering
    \includegraphics{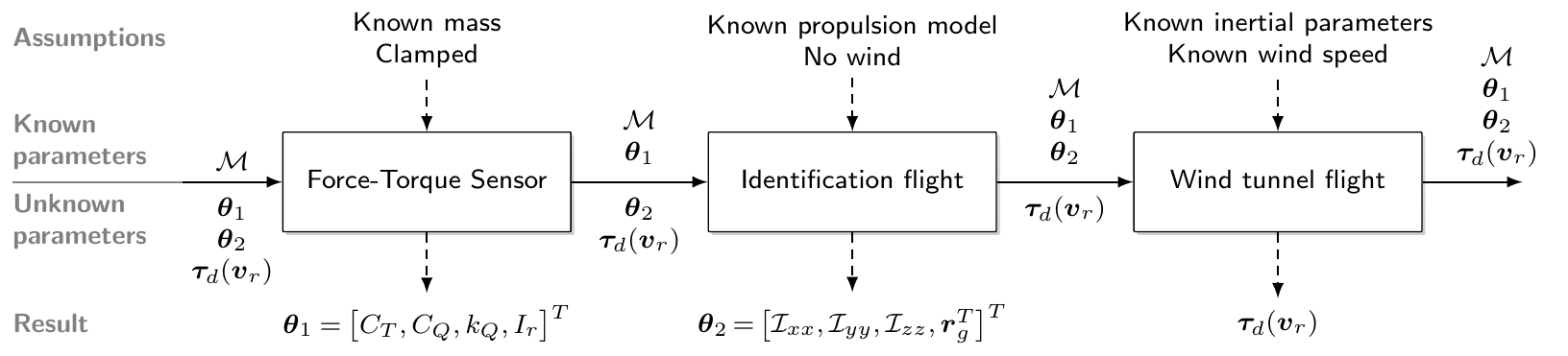}
    \caption{Parameter identification procedure.
        To minimize coupling effects in the high-dimensional parameter space, we perform
        the identification in three steps.
        Using the known vehicle mass $\mathcal{M}$, we estimate the propulsion parameters $\vc{\theta}_1$ on a force-torque sensor.
        Inertia and the center of gravity $\vc{\theta}_2$ are identified from an identification flight without wind.
        Lastly, aerodynamic models $\vc{\tau}_{\!d}(\vc{v}_r)$ are identified from wind tunnel experiments.}
    \label{fig:parameter_estimation_overview}
\end{figure*}

\textbf{Problem formulation.}
The methods presented in this paper are model-based, and as such require model parameters to be identified.
The rigid body and propulsion models can be represented as the linear regression model
\begin{equation}
    \mat{Y} \vc{\theta} = \vc{u},
\end{equation}
where ${\mat{Y} \in \Realm{N}{M}}$ is the regression matrix,
${\vc{\theta} \in \Realv{M}}$ is the vector of unknown parameters,
and ${\vc{u} \in \Realv{N}}$ is the known input.
Here, $N$ is the number of measurement samples, and $M$ is the number of parameters.
For the full hexacopter model, we have to identify
the mass $\mathcal{M}$ (1 parameter),
inertia $\inertia$ (6 parameters),
center of gravity $\vc{r}_{\!g}$ (3 parameters),
thrust and torque coefficients of the coaxial propellers (4 parameters),
propeller inertia (1 parameter)
and the motor torque constants (2 parameters), making in total 18 parameters to be identified.
By using only diagonal inertia terms and known mass, the 3 nondiagonal elements of the inertia matrix $\inertia$ are omitted and the number of unknown parameters is reduced to 15.
In order to further reduce the search space, we perform the parameter estimation in three stages,
as depicted in \fig{fig:parameter_estimation_overview}.

\textbf{Identification methods.}
We compare two methods to obtain the estimated parameters $\hat{\vc{\theta}}$.

\emph{Batch least squares.}
First, the \emph{batch least squares} \citep{Nocedal2006} solution is obtained by
\begin{equation}
    \hat{\vc{\theta}}_{\mathrm{LS}} =
    \bigl( \mat{Y}^T \mat{Y} \bigr)^{-1} \mat{Y}^T \vc{u},
\end{equation}
which minimizes the $\ell_2$ norm of the estimation error.

\emph{Iteratively reweighted least squares (IRLS)}.
Second, we minimize the $\ell_1$ norm of the model residuals by means of IRLS \citep{Chartrand2008}.
This provides robustness to outliers and creates a sparse model by driving some parameters to zero.
The estimate at step $k$ is obtained by solving the weighted least squares problem
\begin{equation}
    \hat{\vc{\theta}}_{\mathrm{IRLS}}^{(k)} =
    \bigl( \mat{Y}^T \mat{W}^{(k)} \mat{Y} \bigr)^{-1} \mat{Y}^T \mat{W}^{(k)} \vc{u}
\end{equation}
where ${\mat{W}^{(k)} = \mathrm{diag} \{ w_1^{(k)}, w_2^{(k)}, \ldots, w_N^{(k)}\}}$ is the weight matrix.
Minimization of an $\ell_p$-norm, ${0\leq p \leq 1}$, is obtained by setting the weights to
$w_i^{(k)} = \bigl( \vc{r}^{(k-1)2} + \epsilon \bigr)^{(p/2) - 1}$,
where ${\vc{r}^{(k)} = \mat{Y} \vc{\theta}^{(k)} - \vc{u} }$ is the estimation residual,
and $\epsilon \ll 1$ is a regularization parameter obtained as described in \cite{Chartrand2008}.

\textbf{Identification procedure.}
A difficulty with identifying all parameters from flight data is the lack of ground truth measurements
of the total torque acting on the robot.
We have also found that estimating the thrust and torque coefficients from flight data is sensitive to time delay
in the measurements (on the order of 20~ms), and can lead to physically meaningless parameters, such as negative thrust coefficients.
We therefore split identification of the system parameters into three parts as depicted in
\fig{fig:parameter_estimation_overview}:
\begin{enumerate}
    \item propeller and motor parameters are obtained using measurements on a force-torque sensor and the known mass,
    \item rigid body parameters are obtained from an identification flight,
    \item aerodynamic models are obtained by flying in a 3D wind tunnel.
\end{enumerate}
We treat the propulsion model as ground truth for the rigid body identification.
In the last step, we estimate the external wrench based on the previously identified models.
In our experiments only the aerodynamic wrench acts on the robot.
We therefore use the estimated external wrench to identify aerodynamic models.
The procedure and results are covered in depth in Section \ref{sec:aero_model_evaluation}.
\begin{figure}
    \centering
    \includegraphics{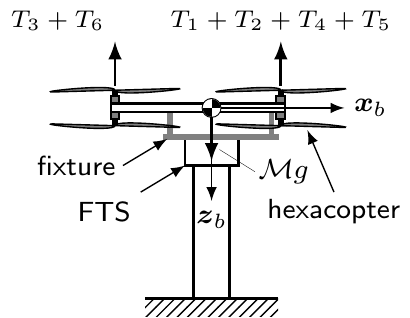}
    \includegraphics[scale=0.65]{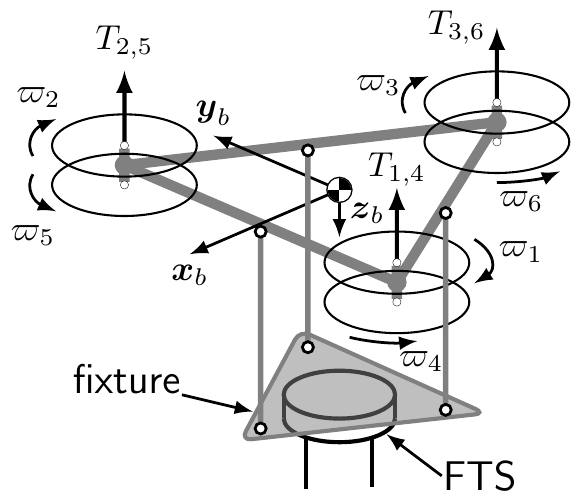}
    \caption{Setup of the force-torque sensor experiment.
        We fixed a hexacopter with coaxial propeller pairs to a force-torque-sensor (FTS).
        The lower and upper propellers have different thrust and torque coefficients
        due to interaction effects.
        }
    \label{fig:fts_setup}
\end{figure}

\textbf{Propulsion system parameters}.
The propulsion parameter vector for our coaxial hexacopter is
\begin{equation}
    {\vc{\theta}_1 := [C_{T,1}, \, C_{T,2}, \, C_{Q,1}, \, C_{Q,2}, \, \vc{K}_{q,1}^T, \, \vc{K}_{q,2}^T, \, I_{r} ]^T},
\end{equation}
and the regression matrix $\mat{Y}_{\!1}$ contains the rotor rates and motor current.
The motor torque coefficients ${\vc{K}_{q,i} = [K_{q,0i}, K_{q,1i}]^T}$ are split for upper and lower motors
($\vc{K}_{q,1}$ and $\vc{K}_{q,2}$ respectively)
because of the aerodynamic interaction between the propellers.
For this step in the identification procedure,
we fixed the hexacopter to an ATI 85 Mini force-torque sensor as depicted in \fig{fig:fts_setup}.
The wrench measured by the sensor is concatenated in ${\vc{u}_1 := \vc{\tau}_{\mathrm{FTS}}}$.
The pose of the hexacopter and the force-torque sensor were obtained by a motion capture system at 250~Hz,
while the onboard attitude controller ran at 500~Hz.
We logged the pose, IMU, motor speed and current as measured by the speed controllers,
the commanded control input, and the force and torque.
The relative orientation of the force-torque sensor to the IMU was calibrated beforehand.
The resulting parameter estimates are listed in Table \ref{tbl:ardea:parameters}.
Note the different motor constants between upper and lower propellers.
In comparison to the lower propellers, the upper propellers generate less thrust ($C_{T,1}< C_{T,2}$)
and require more power ($C_{Q,1} > C_{Q,2}$).

\begin{table}
    \small \sf \centering
    \caption{System parameters identified in the first identification step, using data from a force-torque sensor.
        Coaxial propeller pair coefficients are written as [upper,~lower].
    }
    \label{tbl:ardea:parameters}
    \renewcommand{\arraystretch}{1.2}
    \begin{tabular}{ r  l l  }
        \toprule
        \textbf{Parameter}  & \textbf{Value} \\
        \midrule
        $\mass$                & $2.445$ kg \\
        \hline
        $D$                & $0.254$ m \\
        $C_{T}$                & $[5.1137,~7.8176] \cdot 10^{-2}$ \\
        $C_{Q}$                & $[7.5183,~4.7597] \cdot 10^{-3}$ - \\
        $K_{q,0}$            & $[2.9404, 1.4545] \cdot 10^{-2}$ N\,m/A \\
        $K_{q,1}$            & $[-1.4099,~-3.3360] \cdot 10^{-3}$ N\,m/A${}^2$ \\
        $I_{r}$                & $2.1748 \cdot 10^{-4}$ kg\,m${}^2$ \\
        \bottomrule
    \end{tabular}
\end{table}

\begin{figure*}
    \centering
    \includegraphics{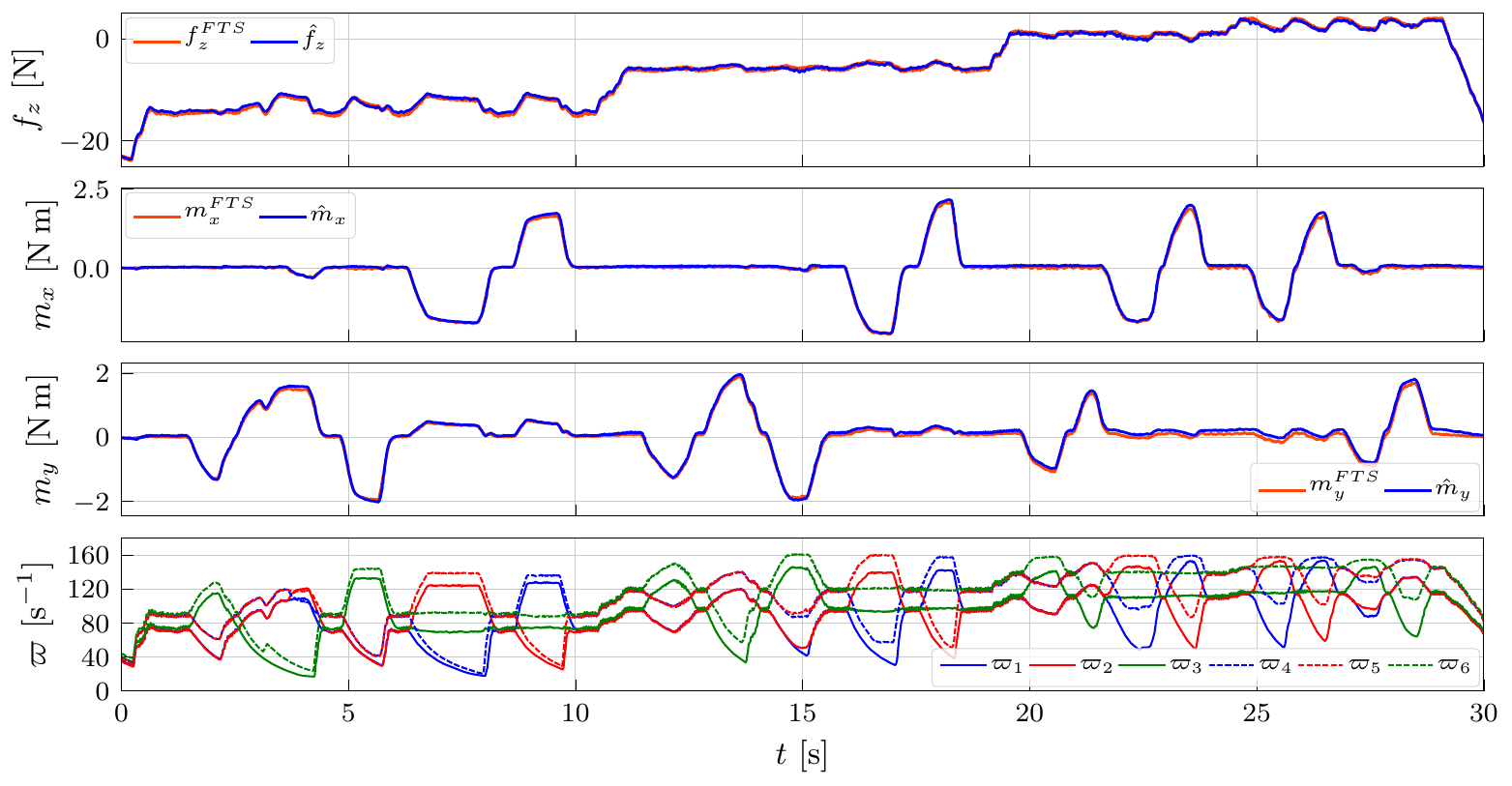}
    \caption[]{Validation of the identified propulsion model
        (\protect\tikz[baseline=-0.5ex]{\draw[thick,blue ] (0,0) -- (1.5ex,0);})
        on the setup depicted in \fig{fig:fts_setup},
        compared to force-torque sensor measurements
        (\protect\tikz[baseline=-0.5ex]{\draw[thick,red  ] (0,0) -- (1.5ex,0);}).
        For clarity we only show the thrust $f_z$ and torques $m_x$ and $m_y$.
        The propulsion forces and torques are obtained using the measured motor speeds, shown in the bottom plot.
        The measured motor speeds $\vc{\varpi}$ of the upper propellers are shown as solid lines
        (\protect\tikz[baseline=-0.5ex]{\draw[thick,blue ] (0,0) -- (1.5ex,0);},
        \protect\tikz[baseline=-0.5ex]{\draw[thick,red ] (0,0) -- (1.5ex,0);},
        \protect\tikz[baseline=-0.5ex]{\draw[thick,green!50!black] (0,0) -- (1.5ex,0);}),
        and lower propellers are shown as dashed lines
        (\protect\tikz[baseline=-0.5ex]{\draw[dash pattern=on 2pt off 2pt, thick,blue ] (0,0) -- (1.5ex,0);},
         \protect\tikz[baseline=-0.5ex]{\draw[dash pattern=on 2pt off 2pt, thick,red  ] (0,0) -- (1.5ex,0);},
         \protect\tikz[baseline=-0.5ex]{\draw[dash pattern=on 2pt off 2pt, thick,green!50!black] (0,0) -- (1.5ex,0);}).
        }
    \label{fig:fts_results}
\end{figure*}

Comparison of the identified model to the force-torque sensor measurements is shown in \fig{fig:fts_results}.
We see that the identified propulsion model closely matches the force-torque sensor measurements.
In this case, using the measured motor speeds to obtain the control wrench shows only a minor improvement over
using the commanded speeds.

\begin{figure}
    \centering
    \includegraphics{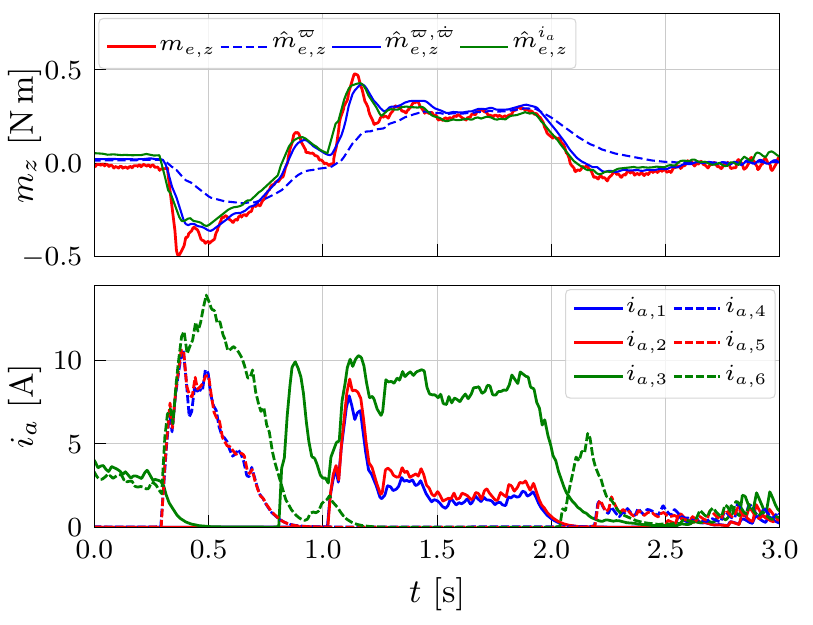}
    \caption{Estimation of the external yaw torque, using
        the measured motor speed only ($m_{e,z}^{\varpi}$),
        using the motor speed and rotor acceleration ($m_{e,z}^{\varpi, \dot{\varpi}}$)
        and using motor current ($m_{e,z}^{i_a}$).
        The estimator gain is ${\mat{K}_{i,m}=36}$ to make the signals more discrenible.
        }
    \label{fig:fts_results_yaw}
\end{figure}

\fig{fig:fts_results_yaw} shows the yaw torque estimation using different measurements.
The model most widely used in literature uses only the motor speed, and is shown as $\hat{m}_{e,z}^{\varpi}$.
This simple model does not capture fast transitions well because the rotor acceleration torque is not modeled.
Adding also the rotor acceleration ($\hat{m}_{e,z}^{\varpi,\dot{\varpi}}$) as in \eqref{eq:torque} improves
accuracy during fast changes of the desired torque, but requires estimation of the rotor acceleration.
Lastly, the motor torque may be obtained directly from the measured motor current by \eqref{eq:current_torque},
shown in \fig{fig:fts_results_yaw} as $\hat{m}_{e,z}^{i_a}$.
Note that the motor torque is used to obtain the yaw component of the propulsion wrench \eqref{eq:propulsion_wrench}.
The measured current is also depicted for illustrative purposes.
In this case, we do not need a propeller model, while the accuracy is similar to the model using rotor acceleration.
Note that in the case of actuator failure (e.g. partially losing a propeller), using the motor current will
provide a better estimate of the yaw torque, as the method does not explicitly model the propeller drag torque.

\begin{figure}
    \centering
    \includegraphics{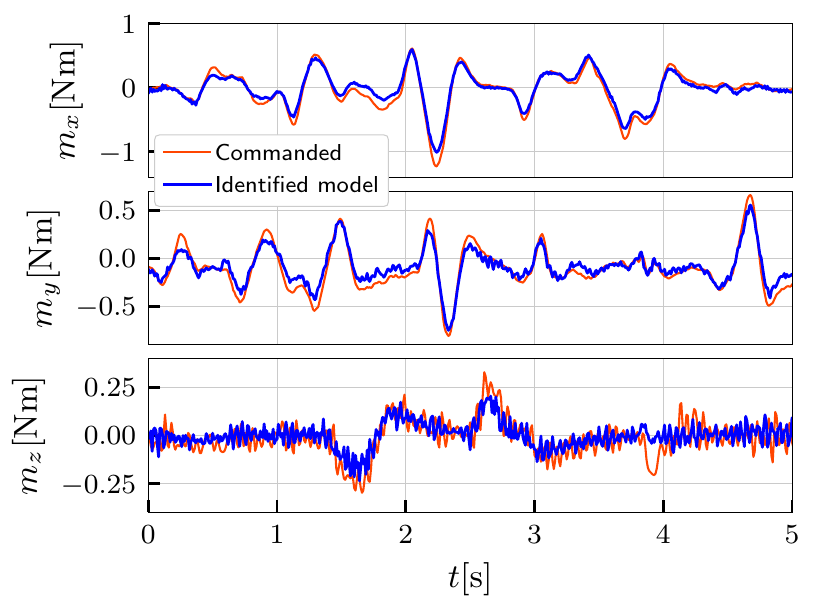}
    \caption{Rigid body torque prediction using parameters obtained by the IRLS method, minimizing the $\ell_1$-norm.
        Overall the model shows a good match to the ground truth.
    }
    \label{fig:parameter_estimation_results}
\end{figure}

\textbf{Rigid body parameters}.
For a diagonal inertia tensor, the rigid body parameter vector is
\begin{equation}
    {\vc{\theta}_2 := [\mathcal{I}_{xx}, \, \mathcal{I}_{yy}, \, \mathcal{I}_{zz}, \, \vc{r}_g^T]^T}.
\end{equation}
On the right-hand side, $\vc{u}_2$ is obtained from the identified propulsion model and
the known mass $\mass$ as
\begin{equation}
    \vc{u}_2 = \mat{Y}_{\!1} \vc{\theta}_1 -  \mass \vc{y}_{\mass},
\end{equation}
where $\vc{y}_{\mass}$ is the regression matrix column associated with the mass.
Furthermore, the off-diagonal inertia terms are more than an order of magnitude lower than the diagonal terms,
which allows us to simplify the model to diagonal inertia.
The identified parameters are listed in Table \ref{tbl:ardea:parameters_comparison}.
\fig{fig:parameter_estimation_results} compares the propulsion model torque to the torque predicted by
the identified rigid body model.
We show the $\ell_1$-identified parameters, as the predicted torque is almost indistinguishable from $\ell_2$.
The result confirms correctness of the identified dynamics model.

\begin{table}
    \small \sf \centering
    \caption{Results of the rigid body parameter identification step, using data from an identification flight,
        and the identified propulsion model.
        Results obtained by batch least squares ($\ell_2$) and IRLS ($\ell_1$) do not differ significantly.
    }
    \label{tbl:ardea:parameters_comparison}
    \renewcommand{\arraystretch}{1.2}
    \begin{tabular}{ l l l }
        \toprule
        $\ell_2$ &
        $\inertia$  & $\mathrm{diag}\{2.58, 2.46, 4.32\} \cdot 10^{-2}$  kg\,m${}^2$ \\
        & $\vc{r}_g$            & $[4.28, -1.11, -11.1]^T \cdot 10^{-3}$  m \\
        \midrule
        $\ell_1$ &
        $\inertia$  & $\mathrm{diag}\{2.54, 2.58, 5.46\} \cdot 10^{-2}$  kg\,m${}^2$ \\
        & $\vc{r}_g$            & $[4.01, -1.05, 8.64]^T \cdot 10^{-3}$  m \\
        \bottomrule
    \end{tabular}
\end{table}

Identification of aerodynamic models was done through wind tunnel experiments, which are described next.

\section{Wind tunnel experiments}
\label{sec:windeee}

\begin{figure*}
    \centering
    \begin{tikzpicture}
        \node[inner sep=0pt, anchor=north west,opacity=1.0] (image) at (0,0)
        {
            \includegraphics[width=1.0\columnwidth]{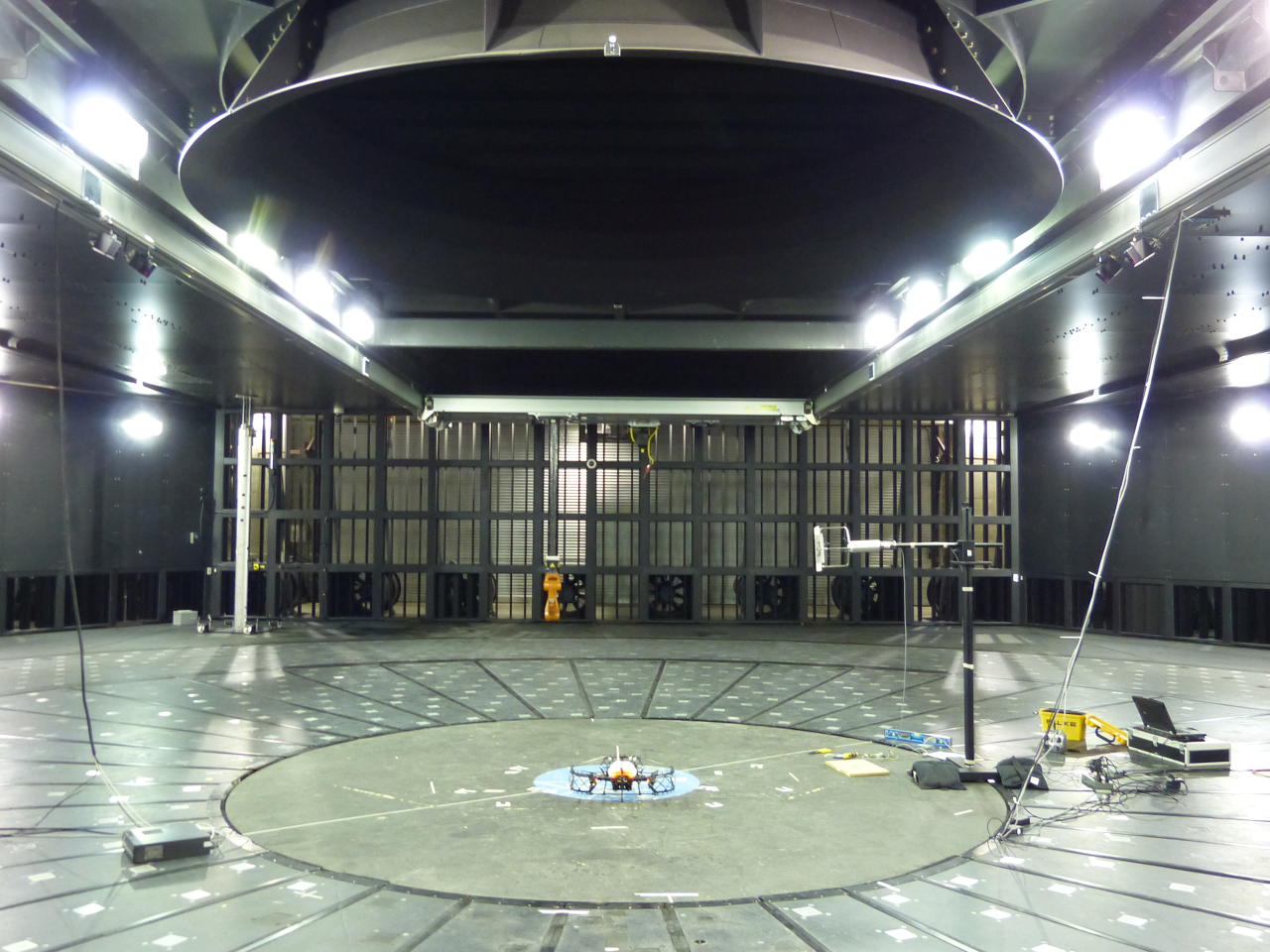}
        };

        \begin{scope}[x={(image.north east)}, y={(image.south west)}, >=latex, draw=white, ultra thick, black, font=\footnotesize]

            \begin{scope}[coordinate,circle,inner sep=10pt]
                \node (art1)          at    (0.1000,0.2667) {};
                \node (art2)          at    (0.88281,0.26667) {};
                \node (ardea)         at    (0.49063,0.81250) {};
                \node (anemometer)    at    (0.65625,0.57292) {};
                \node (groundstation) at    (0.91406,0.76042) {};
                \node (artcontroller) at    (0.13281,0.88021) {};
                \node (vanes)         at    (0.50000,0.16667) {};
            \end{scope}

            \begin{scope}[draw=white, ultra thick, opacity=1.0]
                \draw (art1)            circle(15pt);
                \draw (art2)            circle(15pt);
                \draw (anemometer)      circle(15pt);
                \draw (ardea)           circle(15pt);
                \draw (groundstation)   circle(15pt);
                \draw (artcontroller)   circle(15pt);
            \end{scope}

            \begin{scope}[circle, ultra thick, black, opacity=1.0, inner sep=2pt]
                \node[semithick, fill=white,draw=white] at (vanes) {5};
                \draw[draw=white] (art1)            -- +( 25pt,-20pt) node[semithick,draw=white,fill=white]{1};
                \draw[draw=white] (art2)            -- +(-25pt,-20pt) node[semithick,draw=white,fill=white]{1};
                \draw[draw=white] (anemometer)      -- +(-15pt, 20pt) node[semithick,draw=white,fill=white]{3};
                \draw[draw=white] (artcontroller)   -- +( 15pt, 25pt) node[semithick,draw=white,fill=white]{2};
                \draw[draw=white] (groundstation)   -- +(  6pt, 25pt) node[semithick,draw=white,fill=white]{4};
                \draw[draw=white] (ardea)           -- +( 20pt,-20pt) node[semithick,draw=white,fill=white]{6};
            \end{scope}
        \end{scope}
    \end{tikzpicture}%
    \hspace{1cm}%
    \includegraphics[width=.5\columnwidth]{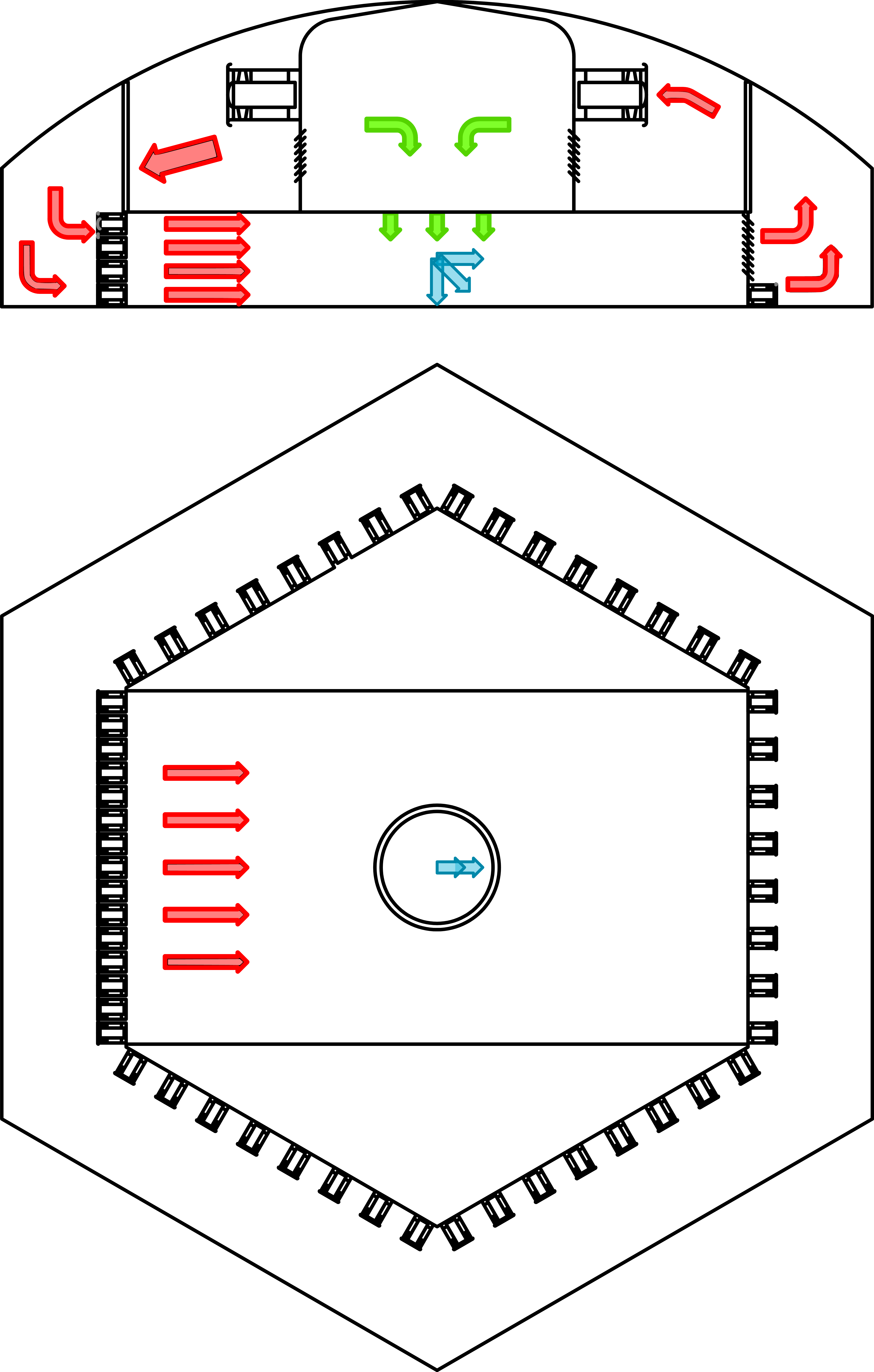}%
    \caption{
        Left: experimental setup inside the wind tunnel:
        four ART Tracking cameras (1) and ART controller (2), RM Young Model 81000 Sonic Anemometer (3),
        groundstation laptop (4),
        vertical wind component inlet (5).
        The flying robot (6) is located in the center of the flying area.
        Right:
        schematic layout of the wind tunnel test. Red arrows show horizontal flow component,
        green arrows show vertical flow component,
        blue arrows show net wind vector.
    }
    \label{fig:experimental_setup}
    \label{fig:windeee_schematic}
\end{figure*}

Experiments to identify aerodynamic models were carried out at
the Wind Engineering, Energy and Environment (WindEEE) Dome, see \cite{Hangan2014}, in London, ON, Canada.
It is the world's first 3D wind chamber, consisting of a hexagonal test area 25\,m in diameter and an outer return dome 40\,m in diameter. Mounted on the peripheral walls and on top of the test chamber are a total of 106 individually controlled fans and 202 louver systems. Additional subsystems, including an active boundary layer floor and "guillotine" allow for further manipulation of the flow. These are integrated via a sophisticated control system which allows dynamic manipulation with thousands of degrees of freedom to produce various time and spatially dependent flows including straight uniform, atmospheric boundary layer, shear gusts, downbursts and tornados at multiple scales. A pair of 5\,m diameter turntables allow for a wide variety of objects to be tested inside and outside the facility.

For this project WindEEE was configured to produce straight flow closed-loop and downburst flows concurrently. In this configuration the test area was restricted to a 4.5\,m diameter, 3.8\,m tall region at the centre of the facility. See Fig. \ref{fig:windeee_schematic} for a schematic drawing of the layout. A rectangular array of 36 fans (9 wide by 4 high) located on the south chamber wall were used to produce horizontal flow and 6 large fans above the test chamber were used to generate the downward flow. The respective flow rates from the horizontal and vertical component fans were manipulated individually to generate net wind vectors ranging in velocity from 1--5\,m/s and vertical plane angularity from 0--90$^\circ$. In some cases both the velocity and vertical plane angularity were manipulated dynamically to produce time-dependent wind vectors that either varied in speed or angularity over a given test run.

\textbf{Motion capture noise}.
Before the experiments we compared the effect of the wind tunnel on motion capture noise by
increasing the horizontal speed to 6 m/s, and keeping the hexacopter stationary. We found that
the wind tunnel did not have a noticable effect on noise in the position and orientation measurements
of the marker attached to the hexacopter.

\begin{figure}
    \centering
    \subfigure[3 m/s]{\includegraphics[width=.32\columnwidth]{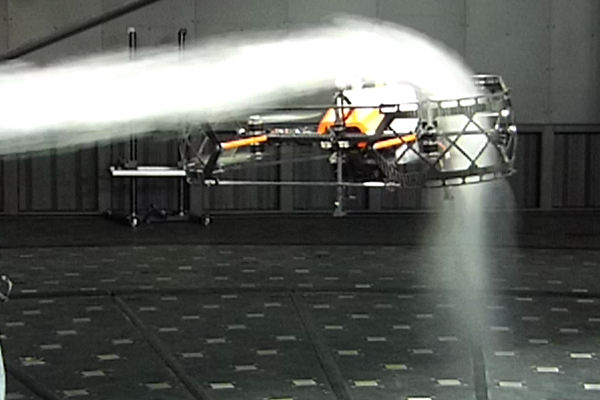}}\hfill
    \subfigure[6 m/s]{\includegraphics[width=.32\columnwidth]{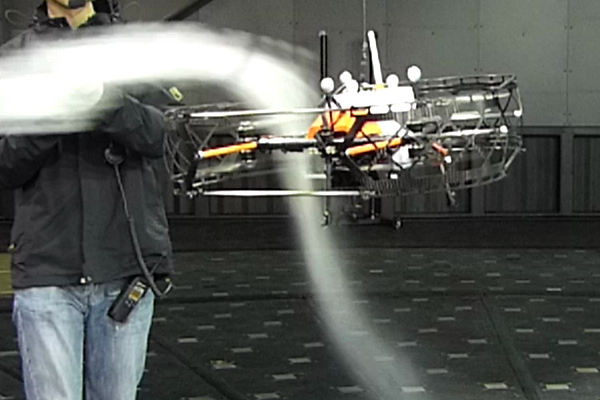}}\hfill
    \subfigure[8 m/s]{\includegraphics[width=.32\columnwidth]{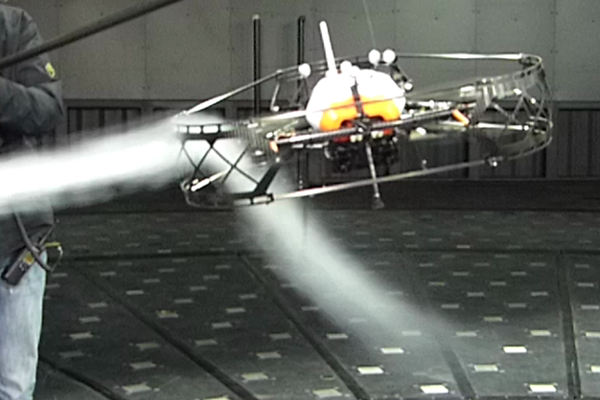}}
    \caption{Flow visualization in the wind tunnel at various horizontal wind speeds.
    Effects due to wind become visibly noticeable around 8 m/s.
    At lower speeds, propellers dominate the airflow.}
    \label{fig:flowvis}
\end{figure}

\textbf{Flow visualization}. For illustrative purposes, we show flow visualization at various wind speeds in \fig{fig:flowvis}.
In steady-state hover, each coaxial rotor pair has to provide a third of the total robot weight.
Therefore, the induced velocity of a coaxial rotor pair at hover is
\begin{equation*}
    v_h = \sqrt{\frac{\tfrac{1}{3} \mass g}{\tfrac{1}{2} \rho D^2 \pi}}
    = \sqrt{\frac{2 \cdot 2.445 \cdot 9.81}{3 \cdot 1.182 \cdot 0.254^2 \pi}}
    = 8.17~\mathrm{m/s},
\end{equation*}
where the air density in the wind tunnel was measured to be 1.182~kg/m${}^3$, based on ambient temperature and pressure.
From the flow visualization, deformation of the airflow becomes noticable at 8~m/s, or about one
induced velocity at hover ($v_{\infty} / v_h \approx 1$).
Note also that due to high $v_h$ and limited range of tested airspeeds, we expect the changes in the
propeller aerodynamic power to be small.

\begin{figure*}
    \centering
    \includegraphics{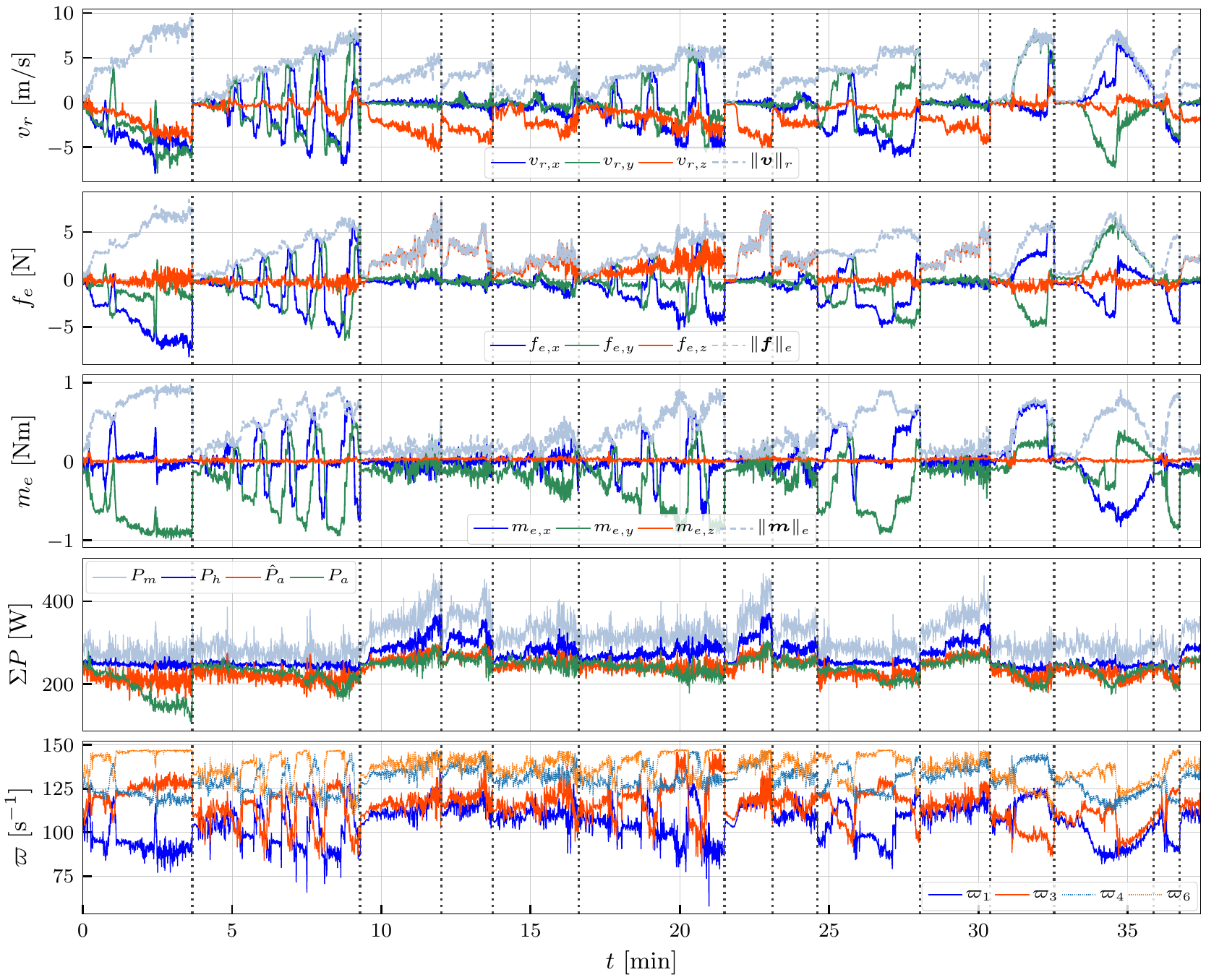}
    \caption{
        The training and validation set for aerodynamics models contains over 37 minutes of flight data
        under horizontal wind speed up to 8\,m/s, and vertical wind speeds up to 5\,m/s.
        Individual flights are delimited by a dotted vertical line.
        From top to bottom:
        relative airspeed in the body frame;
        external force in the body frame;
        external torque;
        sum of coaxial motor and rotor power,
        motor speeds of two coaxial pairs.
    }
    \label{fig:training_set}
    \label{fig:windeee_dataset}
\end{figure*}
\textbf{Dataset.}
For training aerodynamic models, we flew in horizontal, vertical and combined airflows with varying wind speeds.
The robot was hovering in position-controlled mode and was yawing during the experiments.
The concatenated dataset, shown in \fig{fig:training_set}, contains over 37 minutes of flight data
under horizontal wind speed up to 8\,m/s, and vertical wind speeds up to 5\,m/s.
Individual flights are delimited by a dotted vertical line.
Larger vertical wind speeds were not possible due to actuator saturation resulting in loss of yaw control authority.
The relative airspeed is depicted in the body frame. We varied the yaw angle throughout the flights.
We logged the pose from the external tracking system and the onboard visual-inertial navigation system;
IMU data (accelerometer, gyroscope);
control input; motor speed, current, and voltage; anemometer data (wind velocity, direction, temperature, speed of sound).
This is then used to compute the external wrench, relative airspeed, and aerodynamic power.
The plot shows the sum of powers of the three coaxial motor pairs.
The hover power was obtained from ${P_h = 2\rho A v_h^3}$, where $v_h$ is obtained by \eqref{eq:coaxial_vh}.
The expected aerodynamic power is $P_a$, with induced velocity calculated from
\eqref{eq:induced_velocity} and \eqref{eq:aerodynamic_power}, using the relative airspeed
obtained from the anemometer data and external tracking system.
Aerodynamic power is fitted as described in Section~\ref{sec:propeller_aerodynamic_power}.
\fig{fig:training_set} also shows the individual motor speeds of two coaxial pairs.
It can be seen that the motors were saturating for come flow conditions, limiting the maximum achievable wind speed.

\begin{figure}
    \centering
    \includegraphics[width=\columnwidth]{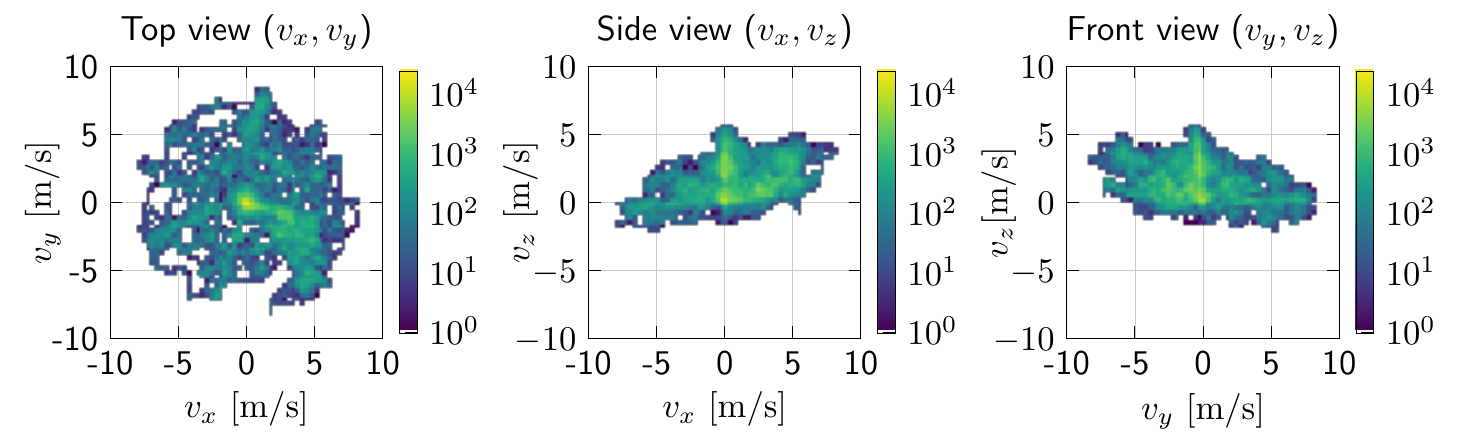}
    \caption{
        Flight data distribution. The intensity shows the number of data points collected at a certain
        relative airspeed. The data is represented here in 80 horizontal and vertical bins.
        Ideally, the training data would be a sphere. However, due to the limited space in the wind tunnel and
        no possibility to get wind from below, we were only able to cover a limited set of $z$-velocities.
        The data is shown here with the $z$-axis pointing upwards.
    }
    \label{fig:training_density}
\end{figure}

\fig{fig:training_density} depicts the distribution of relative airspeeds $\vc{v}_r$ achieved in the dataset.
The shape is an inclined oblate spheroid.
The implication of this shape is that our models will be valid only within this shape,
i.e. the models will have to extrapolate (generalize) beyond the measured airspeeds.
Most notably, the data does not include downward motion,
which would lead to the vortex ring state, but is dominated by horizontal and upward relative airspeeds.
Ideally, the dataset would have been a sphere, however only downward vertical wind speeds
were possible in the wind tunnel.
We used this dataset to identify and evaluate aerodynamic models, which were later used for
simulation studies on force discrimination.

\section{Data-driven wind estimation}
\label{sec:aero_model_evaluation}

In this section, we use the wind tunnel dataset to evaluate aerodynamic models for wind speed estimation.
The evaluation is carried out for combinations of model types and input sets.
We further investigate quality of fit depending on regularization, as well as model generalization.
These findings are then applied to models for simulation and force discrimination.
Here, we try to answer the following questions:
\begin{itemize}
    \item What is the minimal set of inputs required?
    \item Is is possible to estimate the airspeed using motor power measurements only?
    \item Which model/method works/generalizes best?
\end{itemize}
We first investigate the mapping $\vc{v}_r(\vc{u})$, and the effect of different inputs $\vc{u}$ on the model quality,
i.e. the quality of the fit to existing data, and qualitative extrapolation properties of the models.
To estimate the relative airspeed $\vc{v}_r \in \Realv{3}$, we chose to use a combination of the following inputs:
\begin{itemize}
    \item external force $\vc{f}_{\!e} \in \Realv{3}$,
    \item external wrench $\vc{\tau}_{\!e} \in \Realv{6}$,
    \item individual propeller aerodynamic power ${\vc{P}}_{\!a} \in \Realv{6}$, ${\vc{P}}_{\!h} \in \Realv{6}$,
    \item coaxial propeller aerodynamic power $\bar{\vc{P}}_{\!a} \in \Realv{3}$, $\bar{\vc{P}}_{\!h} \in \Realv{3}$,
    \item propeller rotational speed $\vc{\varpi} \in \Realv{6}$.
\end{itemize}
Note that e.g. $\vc{P}_{\!h}$ is a nonlinear map of $\vc{\varpi}$ and is therefore redundant to that input.
We use \eqref{eq:coaxial_vh} to obtain the aerodynamic power of the coaxial pairs
$\bar{\vc{P}}_{\!a} \in \Realv{3}$ and $\bar{\vc{P}}_{\!h} \in \Realv{3}$ by treating each pair as a single propeller.
Because of the complex shape of our robot, we hypothesize that simple models (e.g. blade flapping based)
are not be expressive enough to predict the airspeed.
We therefore investigate models with increasing complexity to provide insights for other researchers.
Lastly, we treat all forces and airspeeds in the body frame.
This implicitly captures aerodynamic angles in the vector components, as they are projected to the body frame.

\subsection{Propeller aerodynamic power}
\label{sec:propeller_aerodynamic_power}
\begin{figure}
    \centering
    \includegraphics[width=.9\columnwidth]{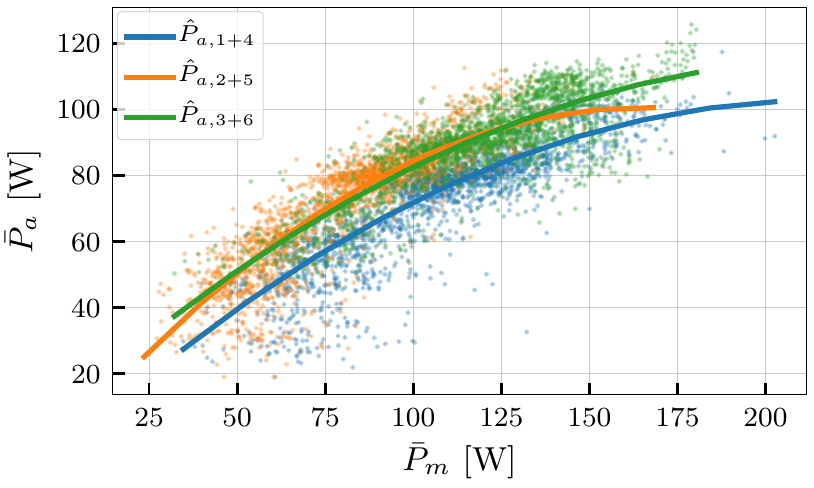}
    \caption{Fit of aerodynamic power for individual propeller pairs. The points represent the individual
    data points for each propeller pair, while the line shows a quadratic fit.}
    \label{fig:aerodynamic_power_fit}
\end{figure}

Using individual motor powers for coaxial propellers leads physically meaningless results,
where the figure of merit of the lower motors can be above 1, whereas it should be around 0.5--0.6 (see \cite{Leishman2006}).
In order to use the propeller aerodynamics model presented in Section~\ref{sec:propeller_aerodynamics},
we must fit the motor power measurements \eqref{eq:motor_power} to the aerodynamic power \eqref{eq:aerodynamic_power}.
\fig{fig:aerodynamic_power_fit} shows the fit of each propeller pair data to the quadratic model
\begin{equation}
    \bar{P}_{a,j+k} = P_0 + \beta_1 \bar{P}_{m,j+k} + \beta_2 \bar{P}_{m,j+k}^2,
\end{equation}
where the mechanical power $\bar{P}_m$ of each coaxial pair ${j+k}$ is the sum of individual motors' mechanical power, i.e.
\begin{equation}
    \bar{P}_{m,j+k}:=\hat{P}_{m,j} + \hat{P}_{m,k}
\end{equation}
The mechanical power of motor $i$ without the rotor acceleration term is obtained by
\begin{equation}
    \hat{P}_{m,i} = \Bigl(\bigl(K_{q,0i} - K_{q,1i}i_{a,i} \bigr) i_{a,i} - I_r \dot{\varpi}_i \Bigr) \varpi_i
    .
    \label{eq:estimated_mechanical_power}
\end{equation}
The model fit uses IRLS and shows good agreement with the data.
Helicopter aerodynamics literature (e.g. \cite{Leishman2006}) proposes a linear fit from shaft power to aerodynamic power, in the form of a figure of merit $F\!M$.
However, we only measure the current and cannot directly measure the shaft torque.
The quadratic model therefore also captures losses in the conversion of electrical to mechanical power.
Now that we can relate motor power to momentum theory, we may use it in further analyses.

\begin{figure}
    \centering
    \includegraphics[width=\columnwidth]{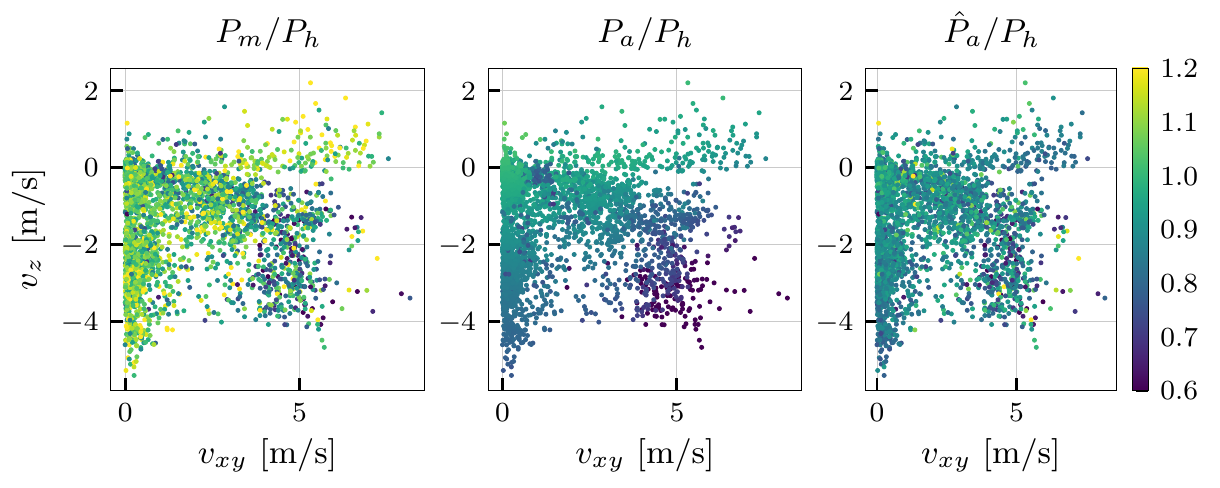}
    \caption{Power ratio at different airspeeds, relative to aerodynamic power in hover.
    From left to right: motor power, aerodynamic power, fitted aerodynamic power.}
    \label{fig:power_ratio}
\end{figure}

Momentum theory may be used to predict the ratio of aerodynamic power in forward flight to the aerodynamic power in flight.
This is visualized for our dataset in \fig{fig:power_ratio}, for the motor power, and the fitted aerodynamic power.
Note that momentum theory predicts a decline in power at oblique angles of attack (high horizontal and vertical airspeed components).
However, we could not observe this effect during the combined airflow experiment (67${}^{\circ}$ inflow angle), which is the first
flight in \fig{fig:training_set}. Notably, the motor power saturates on the low side.
We postulate that this may be due to motor speed saturations from compensating the simultaneously large external force and torque during the flight.
This produces more losses in the motor, and we cannot deduce information about the wind speed.
The fit is better in the remainder of the flights, where motors do not saturate for large parts of the flight.
Based on this, we conclude that any method that uses motor power as input will not be valid in periods of motor input saturation.

\subsection{Aerodynamic models}
Next we present various models used in our evaluation.
The problem considered here is regressing from a set of inputs $\vc{u}$
to the vector of body-frame relative airspeed $\vc{v}_r$ (three velocity components).
Data fitting was done using a 67\%--33\% train--test split of the complete wind tunnel dataset.
For efficient training the inputs and outputs were normalized to a range of $[0, 1]$.
We used the Python package {\sf \small scikit-learn} \citep{scikit-learn} for its machine learning functions.

\subsubsection{Physics-based models}

Physics-based models are obtained from first-principles modeling, such as conservation of momentum.
As such, they provide insight into the structure of the problem.
However, some effects are usually neglected due to assumptions contained therein.

\textbf{Blade flapping.}
It is well established in multirotor literature that the dominant horizontal force in multicopters is due to
propeller induced drag \citep{Waslander2009, Omari2013}.
The widely used induced-drag and blade flapping model \citep{Omari2013} can be written as
\begin{equation}
    \vc{f}_{\!d}(\vc{v}_r) = \mat{D}_l \;\vc{v}_r \sum_i \varpi_i
    ,
\end{equation}
where $\mat{D}_l$ is the matrix of coefficients, and $\varpi_i$ is the speed of the $i$--th propeller.
We refer the reader to the cited literature for the derivation of the model.

\textbf{Parasitic drag.}
The blade flapping model may be extended with a quadratic parasitic term due to form drag such that
\begin{equation}
    \vc{f}_{\!d}(\vc{v}_r) = \mat{D}_l \;\vc{v}_r \sum \varpi_i + \mat{D}_q \; \vc{v}_r | \vc{v}_r |
    ,
\end{equation}
where $\mat{D}_q$ is the parasitic drag matrix.
We use these models as a starting point to design more general regression models.

\subsubsection{Linear regression models}
We consider physics-based models in the general framework of linear regression.
The linear regression model can be written as
\begin{equation}
    \vc{y}(\vc{u}) = \mat{X}\!(\vc{u}) \, \vc{w},
\end{equation}
where $\vc{y} \in \Realv{N_y}$ is the model output, $\vc{u} \in \Realv{N_u}$ is the input,
$\mat{X}(\vc{u}) \in \Realm{N_y}{N_w}$ is the regression matrix,
and $\vc{w} \in \Realv{N_w}$ are the model weights.
The weights are obtained by solving the minimization problem
\begin{equation*}
    \min_{\vc{w}} \tfrac{1}{2 N} \| \mat{X} \vc{w} - \vc{y} \|_2^2 + \alpha_1 \| \vc{w} \|_1,
\end{equation*}
where $N$ is the number of samples, $\alpha_1$ is the regularization factor, and $\|\vc{w}\|_1$ is the
$\ell_1$ norm of the model weights.
This will drive some of the model weights to zero, leading to sparse models depending on $\alpha_1$.

To apply this model to physics-based models, $\mat{X}$ contains the linear and quadratic terms of the input,
while $\vc{w}$ contains the matrix elements.
In our evaluation, we consider the linear model and quadratic model.
The \emph{linear model} can be written as
\begin{equation}
    \vc{y}(\vc{u}) = \mat{W} \vc{u},
\end{equation}
where $\mat{W}$ is the matrix of weights, and $\vc{u}$ is the input vector.
This gives a maximum of $N_u \times N_y$ nonzero parameters.
Similarly, the \emph{quadratic model} can be written as
\begin{equation}
    \vc{y}(\vc{u}) = \mat{W}_{\!1} \vc{u} + \mat{W}_{\!2} \vc{u} | \vc{u} |,
\end{equation}
where $\mat{W}_{\!1}$ is the matrix of linear weights,
$\mat{W}_{\!2}$ is the matrix of quadratic weights,
This gives a maximum of $2 \times N_u \times N_y$ nonzero parameters.

\subsubsection{Multilayer perceptron model}
To compare a purely data-driven approach to physically based modeling, we apply nonlinear regression to the data.
For this purporse we use a multilayer perceptron with one hidden layer using the $\tanh$ activation function,
and a linear output layer. We have found that using more layers did not improve the fit, and using other activation functions
did not change the results significantly.
The model can be summarized as
\begin{equation}
    \vc{y}(\vc{u}) = \mat{W}_{\!2} \Bigl( \sigma \bigl( \vc{W}_{\!1} \vc{u} + \vc{b}_1 \bigr) + \vc{b}_2 \Bigr)
\end{equation}
where $\mat{W}_{\!1} \in \Realm{N_h}{N_u}$,
$\mat{W}_{\!2} \in \Realm{N_y}{N_h}$ are the neuron weight matrices,
$\vc{b}_1 \in \Realv{N_h}$ and $\vc{b}_2 \in \Realv{N_y}$ are the bias vectors,
and ${\sigma(\cdot) = \tanh(\cdot)}$ is the activation function.
This gives a maximum of $(N_u + N_y + 1) \times N_h + N_y$ nonzero parameters.
For the perceptron we use $\ell_2$ regularization of the weights using $\alpha_2$ as regularization parameter.
Training is performed by stochastic gradient descent \cite{scikit-learn}.

\subsection{Model performance}
\label{sec:airspeed_models}
\begin{figure*}
    \centering
    \begin{tikzpicture}[>=latex]
        \node[inner sep=0pt, anchor=south west,opacity=1.0] (image) at (0,0)
        {\includegraphics[width=0.95\textwidth]{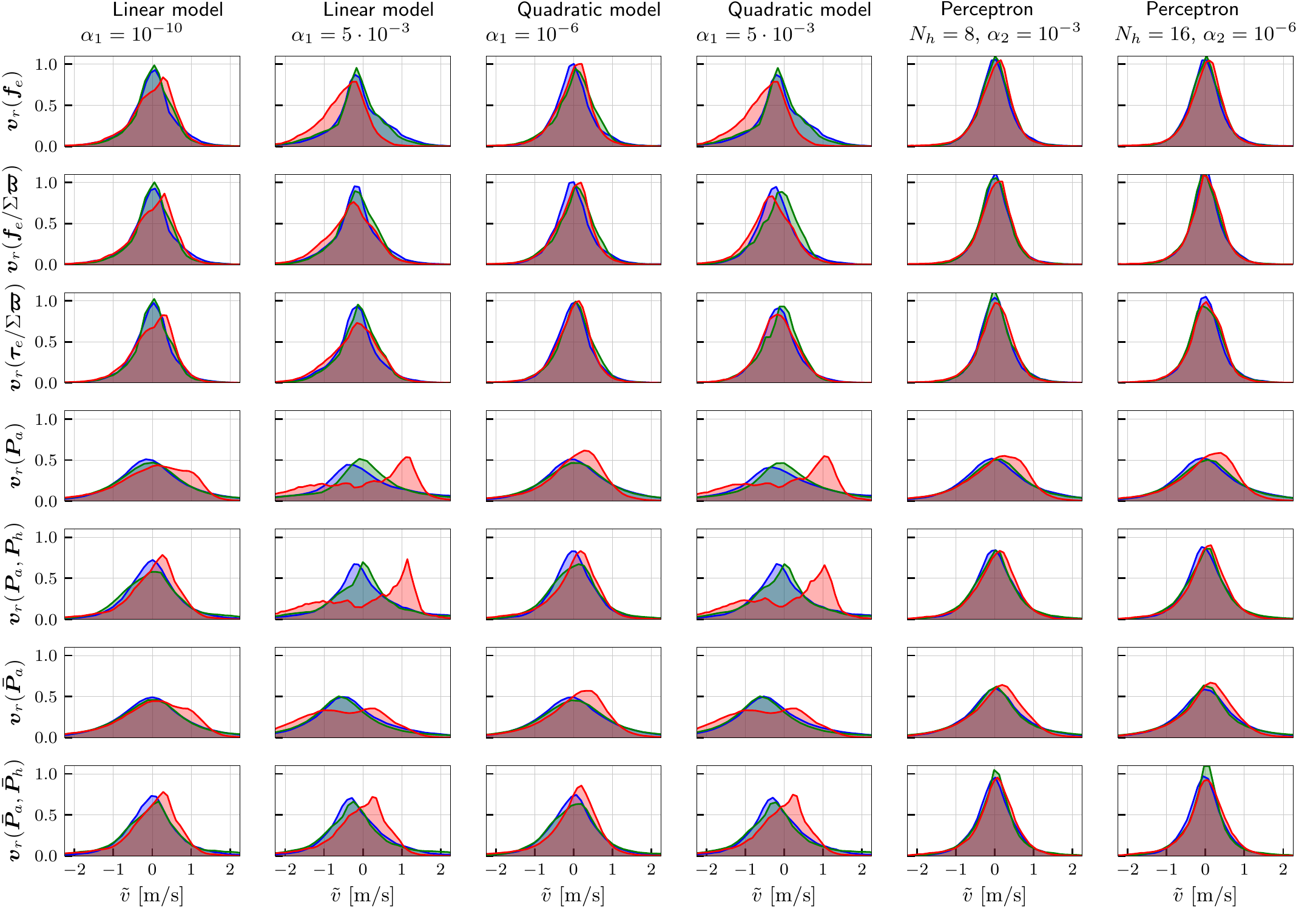}};

        \begin{scope}[x={(image.south east)}, y={(image.north west)}, >=latex, draw=gray, densely dashed, very thick, line cap=round]
            \draw (-0.03, 0.305) -- +(1.02,0);
            \draw (-0.03, 0.5625) -- +(1.02,0);
            \draw (-0.03, 0.8225) -- +(1.02,0);
            \node[rotate=90] at (-0.02, 0.16250) {\sffamily \scriptsize Coaxial aero. power};
            \node[rotate=90] at (-0.02, 0.43375) {\sffamily \scriptsize Aerodynamic power};
            \node[rotate=90] at (-0.02, 0.69250) {\sffamily \scriptsize Rotor speed normalized};
            \node[rotate=90] at (-0.02, 0.91125) {\sffamily \scriptsize External force};

            \node at (1.0, 0.885) {\sffamily \scriptsize A};
            \node at (1.0, 0.755) {\sffamily \scriptsize B};
            \node at (1.0, 0.625) {\sffamily \scriptsize C};
            \node at (1.0, 0.495) {\sffamily \scriptsize D};
            \node at (1.0, 0.365) {\sffamily \scriptsize E};
            \node at (1.0, 0.235) {\sffamily \scriptsize F};
            \node at (1.0, 0.105) {\sffamily \scriptsize G};

            \node at (0.115, -0.015) {\sffamily \scriptsize 1};
            \node at (0.275, -0.015) {\sffamily \scriptsize 2};
            \node at (0.435, -0.015) {\sffamily \scriptsize 3};
            \node at (0.595, -0.015) {\sffamily \scriptsize 4};
            \node at (0.755, -0.015) {\sffamily \scriptsize 5};
            \node at (0.915, -0.015) {\sffamily \scriptsize 6};
        \end{scope}
    \end{tikzpicture}
    \caption{
    Normed error histograms of predicted airspeed as a function of different input sets, for different models.
    The histograms for $x$ ({\color{blue}blue}), $y$ ({\color{green}green}) and $z$ ({\color{red}red}) airspeed are overlaid to
    better depict per-axis model performance.
    The letters on the right (rows) and numbers on the bottom (columns) define a grid for easier reference in the text.
    }
    \label{fig:aero_model_histogram}
\end{figure*}
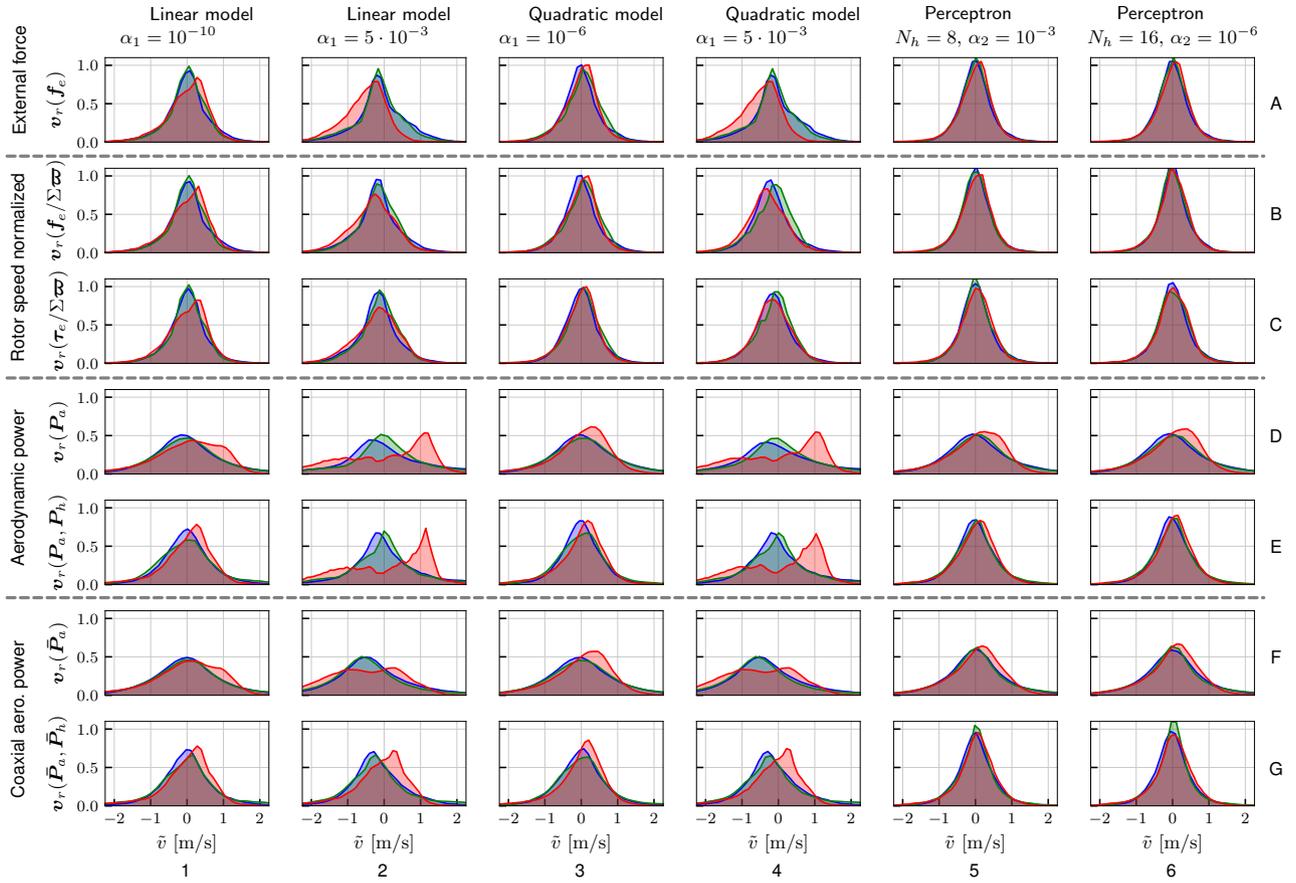

\fig{fig:aero_model_histogram} shows error histograms of the test data set for combinations of models and inputs
as listed in Tables \ref{tbl:aero_models} and \ref{tbl:aero_inputs}, respectively.
Simple blade flapping based models have a problem predicting the body-vertical relative airspeed.
The plot therefore contains histograms of individual relative airspeed components.
\begin{table}
    \small \sf
    \centering
    \caption{Summary of tested regression models.}
    \begin{tabular}{ll}
        \toprule
        \textbf{Model} & \textbf{Formulation} 
        \\
        \midrule
        Linear model &
        $\vc{y} = \mat{W} \vc{u}$
        \\
        Quadratic model &
        $\vc{y} = \mat{W}_{\!1} \vc{u} + \mat{W}_{\!2} \vc{u} | \vc{u} |$
        \\
        Perceptron &
        $\vc{y} = \mat{W}_{\!2} \Bigl( \tanh \bigl( \vc{W}_{\!1} \vc{u} + \vc{b}_1 \bigr) + \vc{b}_2 \Bigr)$
        \\
        \bottomrule
    \end{tabular}
    \label{tbl:aero_models}
\end{table}
\begin{table}
    \centering
    \caption{Summary of tested inputs for predicting relative airspeed.}
    \footnotesize \sf
    \begin{tabular}{lllc}
        \toprule
        \textbf{ID} & \textbf{Label} & \textbf{Input data} & $N_u$ \\
        \midrule
        A &
        External force &
        $\vc{u}_A = \vc{f}_{\!e} $ &
        3
        \\
        B &
        Rotor speed normalized $\vc{f}_{\!e}$ &
        $\vc{u}_B = \vc{f}_{\!e} / \sum_i \varpi_i$ &
        3 \\
        C &
        Rotor speed normalized $\vc{\tau}_{\!e}$ &
        $\vc{u}_C = \vc{\tau}_{\!e} / \sum_i \varpi_i$ &
        6 \\
        D &
        Aerodynamic power &
        $\vc{u}_D = \hat{\vc{P}}_a$ &
        6 \\
        E &
        Aerodynamic and hover power &
        $\vc{u}_E = \bigl[\hat{\vc{P}}_a^T \vc{P}_h^T \bigr]^T$ &
        12 \\
        F &
        Coaxial aero. power &
        $\vc{u}_F = \bar{\vc{P}}_a$ &
        3 \\
        G &
        Coaxial aero. and hover power &
        $\vc{u}_G = \bigl[\bar{\vc{P}}_a^T \bar{\vc{P}}_h^T \bigr]^T$ &
        6 \\
        \bottomrule
    \end{tabular}
    \label{tbl:aero_inputs}
\end{table}
From the results in \fig{fig:aero_model_histogram}, we can draw the following conclusions:

(1) Even the linear model (column 1) obtains good results for all input sets, for low regularization.
As the model becomes sparser (i.e. contains fewer nonzero parameters) with increased regularization (column 2), it cannot predict the airspeed accurately anymore.

(2) Comparing the wrench input (row C) to the force input (row B), we see that the external torque does not add significant information to the external force.
In fact, the histograms are almost identical.

(3) Multiplying the wrench by the inverse sum of rotor speeds $\sum_i \varpi_i$ improves the results, see linear and quadratic models
with higher regularization (B2, B4, C2, C4). This indicates that this is a more accurate representation of the underlying physics (blade flapping and induced drag).
However, it does not significantly improve the nonlinear perceptron models, nor models with low regularization.

(4) The best regression results were achived by perceptron with $N_h=16$ hidden neurons for the $\fext / \sum_i \varpi_i$ input (B6).
The difference is most notable in the $z$-axis, where the offset present in most histograms is absent.
We use this as a benchmark of the best possible regression for this dataset, and may compare simpler models to this one.
In comparison, the preceptron with $N_h=8$ hidden neurons (column 5) is almost as good with a lot fewer parameters.

(5) Increasing the perceptron model complexity does not significantly improve the airspeed regression after just several
neurons in the hidden layer. This indicates that the underlying model has low complexity.
We have similarly found that multiple hidden layers did not improve the regression (not shown in the histogram).

\begin{figure*}
    \centering
    \subfigure[Dataset distribution]{
        \includegraphics{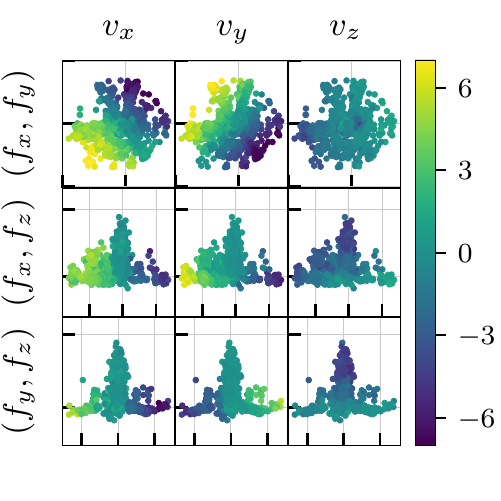}
        \label{fig:model_shapes_dataset}
    }%
    \hfill
    \subfigure[Linear model]{
        \includegraphics{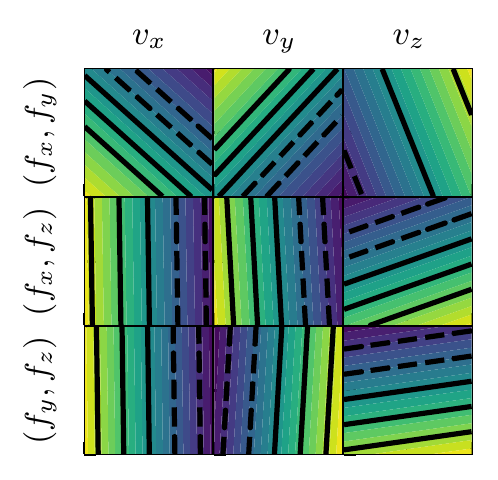}
        \label{fig:model_shapes_linear}
    }
    \hfill
    \subfigure[Perceptron model with 8 hidden $\tanh$ neurons, $\alpha=10^{-6}$]{
        \includegraphics{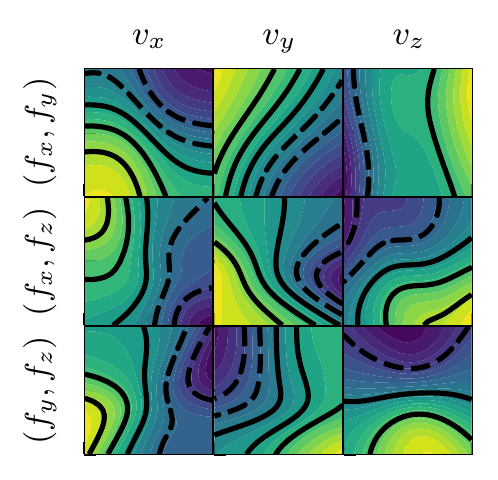}
        \label{fig:model_shapes_perceptron}
    }
    \caption{
    Distribution of external force samples used for training and validation of the mapping \emph{force (input) to relative airspeed (output)}.
    The color indicates the airspeed component $v_x$, $v_y$ or $v_z$ per column, brighter shades depict larger values.
    The distribution shows where learned models will be valid.
    Each row shows a different plane of the forces.
    The first row shows ($f_x$, $f_y$) for $|f_z| < 0.1$ N,
    the second ($f_x$, $f_z$) for $|f_y| < 0.1$ N,
    and the third one ($f_y$, $f_z$) for $f_x < 0.1$ N.
    }
    \label{fig:model_shapes}
\end{figure*}
(6) The relative airspeed may be predicted by using only the aerodynamic power as input.
Adding the aerodynamic power in hover (which is directly related to the control input, i.e. motor speeds) improves the fit.
This is due to the fact that in steady state, the control input is equal to the external wrench, ${\vc{\tau} = \vc{\tau}_{\!e}}$.
The aerodynamic power in hover $P_h$ is computed from the motor speeds $\vc{\varpi}$, which relate to the control input
through the control allocation matrix $\mat{B}$ as ${\vc{\varpi} = \mat{B}^{+} \vc{\tau}}$.
Therefore, the aerodynamic power in hover is a function of the external wrench, i.e. ${\vc{P}_h := \vc{P}_h(\vc{\tau}_{\!e}})$.
This result must therefore be taken with caution, as the model essentially learns this nonlinear transformation.
Our dataset does \emph{not} include external forces other than wind, and the robot was hovering in place.
The desired result of using this regression as an independent measurement of the airspeed is therefore not attained.

(7) The offset in the $z$-velocity for power inputs indicates difficulty in modeling the aerodynamic power from
motor power measurements (without the hover power).

(8) Aerodynamic power of coaxial propeller pairs is a more accurate representation than considering each propeller individually,
as it is a better representation of underlying physics.

In summary, we may answer two of the questions posed at the beginning of this section:
\begin{enumerate}
    \item The minimum set of inputs is the external force $\vc{f}_{\!e}$.
    \item It is possible to estimate the airspeed using motor power measurements. However, this will be noisier than when using the external force.
\end{enumerate}
The last question about model quality (i.e. model selection) requires more investigation.
To answer, we next investigate the influence of regularization on the prediction quality of the models.

\subsection{Model generalization}
In order to reason about model fit and generalization, the dataset is visualized in \fig{fig:model_shapes_dataset}.
Therein, we consider the mapping from external force (input) to relative airspeed (output).
Rows show three planar sections of the three-dimensional input force samples in the dataset.
The upper-left plot $(f_x,f_y)$ shows the $x,y-$plane, where $f_z$ is close to zero.
From top to bottom the data points are $(f_x, f_y, 0)$, $(f_x, 0, f_z)$, and $(0, f_y, f_z)$.
In each plot, the color visualizes the output intensity -- lighter colors indicate higher airspeed values.
The columns represent different views of the model outputs, i.e. the relative airspeeds $v_x$, $v_y$, and $v_z$, respectively.
\fig{fig:model_shapes_linear} and \fig{fig:model_shapes_perceptron} depict how the dataset was fitted by different models.
Every contour line in the plots show a 3 m/s increase in airspeed (dashed lines for negative values).
\fig{fig:model_shapes_dataset} depicts the distribution of the training points, simplifying visually reasoning about the range of validity of the fitted models.
As expected from the shape of training airspeeds in \fig{fig:training_set}, horizontally the input is a circle.
However, the total shape is conic, because the number of combined airflow datapoints with substantial horizontal airspeed is limited.
Clearly, fitted models will extrapolate in the regions where airspeed data has not been sampled.

\begin{figure}
    \centering
    \subfigure[Quadratic model prediction of $v_{r,z}$ for $(0, f_y, f_z)$.]{
        \includegraphics[width=.9\columnwidth]{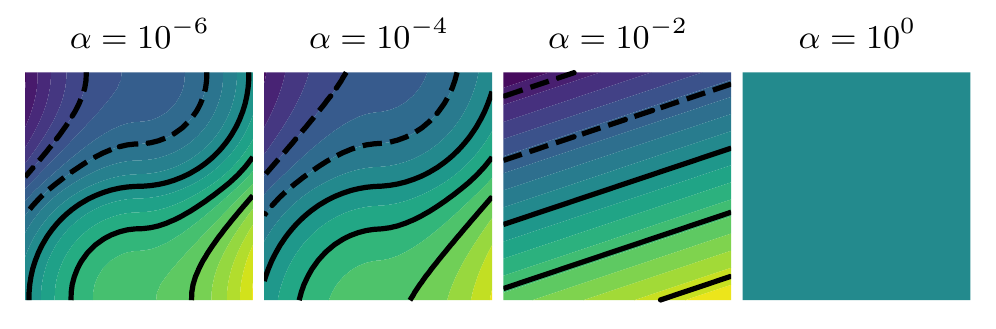}
    }
    \subfigure[Perceptron with $N_h=8$ prediction of $v_{r,z}$ for $(0, f_y, f_z)$.]{
        \includegraphics[width=.9\columnwidth]{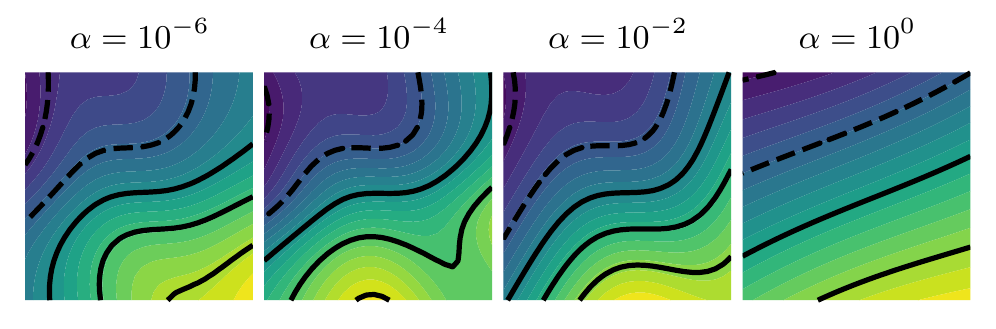}
    }
    \caption{Change of model shape with increasing regularization.}
    \label{fig:model_regularization_shape}
\end{figure}

The difference in expressivity between a linear and a perceptron model can also be seen from this visualization.
Obviously, the perceptron model is more expressive. However, given the distribution of the training dataset,
it is prone to overfitting to the training data. Outside of the training data distribution, the output may be physically meaningless.
Therefore, a regularization parameter must be found for each model that both fits the training distribution, and generalizes outside of it.
To furthter visualize the change in model outputs,
\fig{fig:model_regularization_shape} shows the shape of $v_{r,z}$ for a quadratic model and a perceptron with 8 hidden neurons,
with varying regularization values. Obviously, both models converge to a linear fit as regularization increases.
However, the visualization alone does not tell us which model performs best.
\begin{figure}
    \centering
    \includegraphics[width=.8\columnwidth]{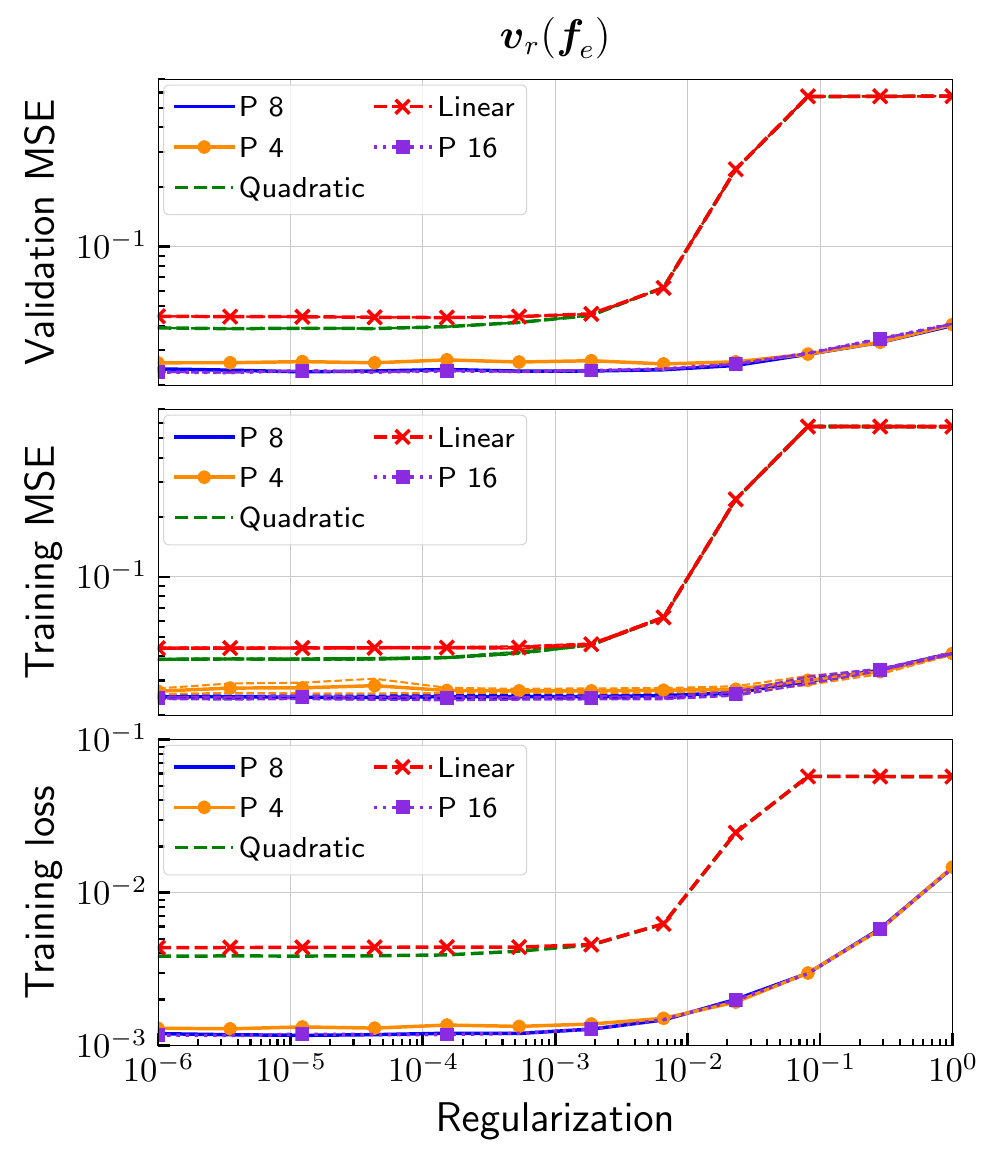}
    \caption{Training and validation mean square errors for 10-fold cross-validation of different models fitting $\vc{v}_r (\vc{f}_e)$,
               with varying regularization factors. Perceptron models with 4, 8, and 16 neurons in the hidden layer are labeled as P 4, P 8, and P 16, respectively.}
    \label{fig:model_regularization_models}
\end{figure}
We adopt the following procedure to choose a model regularization parameter for good generalization.
First, we split the complete dataset into training (70\%) and validation (30\%) datasets.
The data points are therefore drawn from the same distribution. However, training never sees data from the validation set (holdout).
We then perform $K$-fold cross-validation.
The model is trained on a fraction of the training data, and a validation score is obtained on the remaining data.
The procedure is repeated $K$ times to get a mean and standard deviation of the training error and training loss (which includes regularization).
Finally, in order to obtain the validation error, we train a model on the complete training dataset.
The validation error is computed between the model-predicted airspeed and the ground truth airspeed in the holdout set.
To reason about generalization, we are interested in the training loss to see when the regularization term in the cost function starts to dominate the loss,
i.e. the model starts becoming linear.

\begin{figure}
    \centering
    \subfigure[Quadratic model]{
        \includegraphics[width=.8\columnwidth]{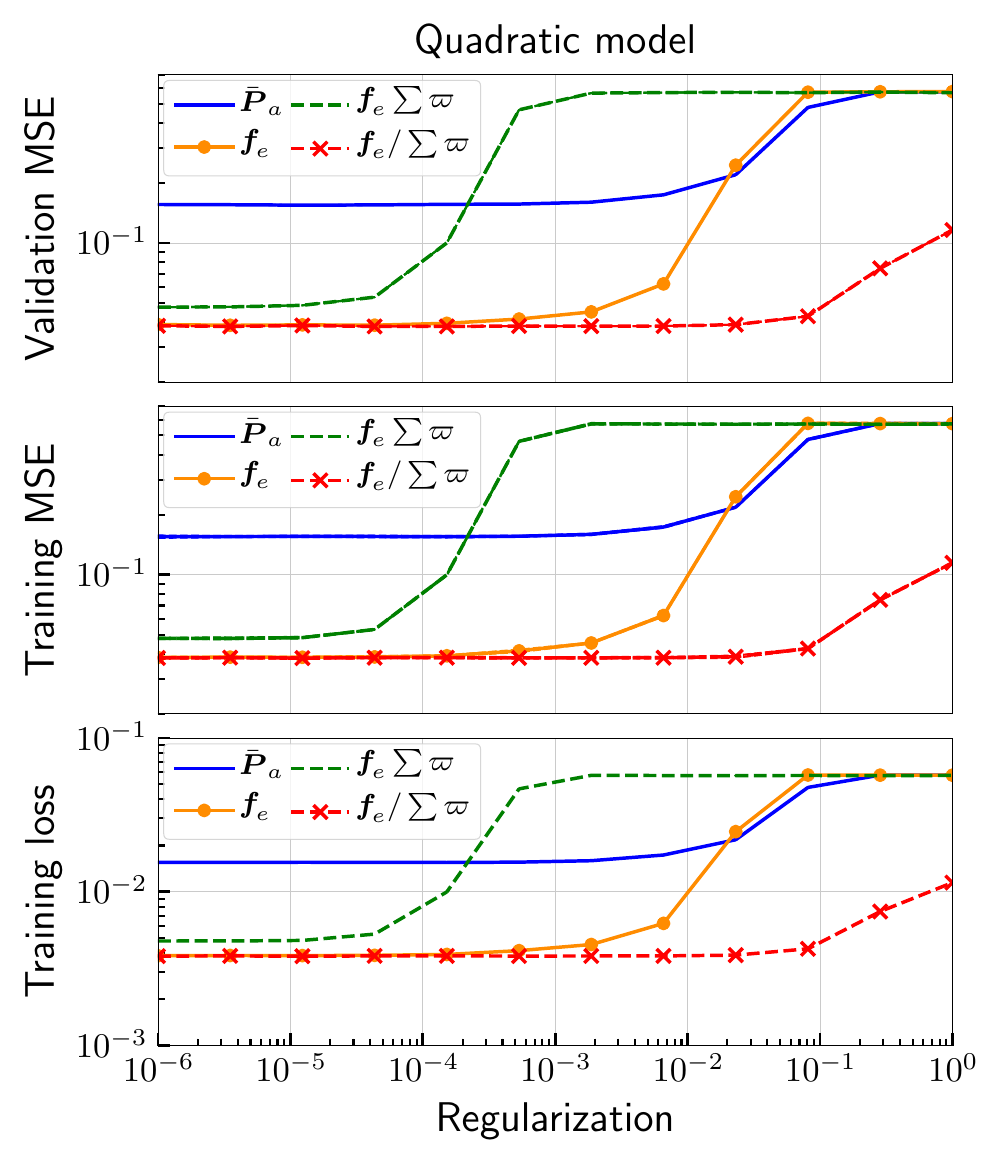}
        \label{fig:model_regularization_quadratic}
    }
    \subfigure[Perceptron model with $N_h=8$]{
        \includegraphics[width=.8\columnwidth]{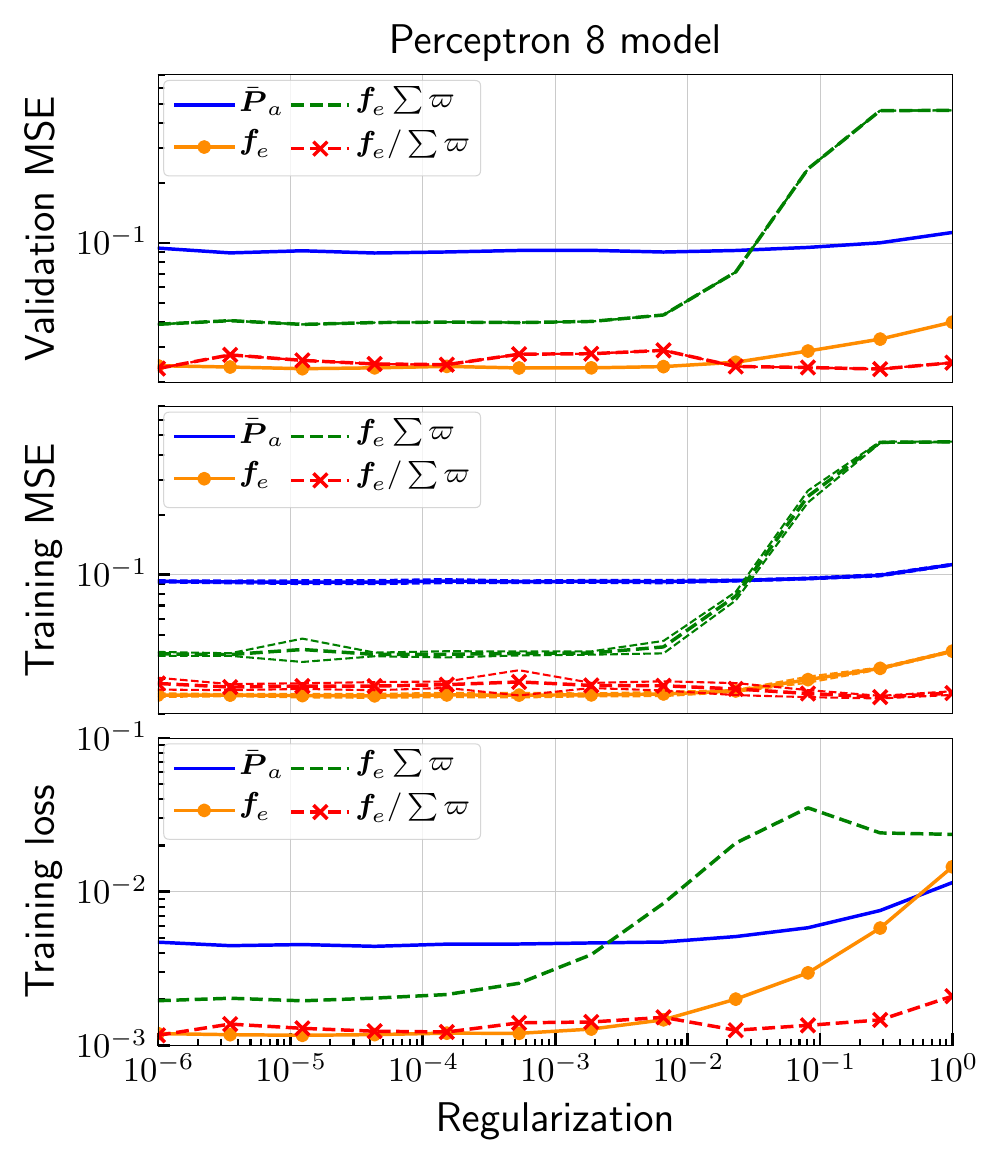}
        \label{fig:model_regularization_perceptron}
    }
    \caption{Training and validation mean square errors for 10-fold cross-validation of a Perceptron model with $N_h=8$ and 
            a quadratic model fitting $\vc{v}_r (\vc{u})$, with varying $\vc{u}$ and regularization factors.}
\end{figure}
\fig{fig:model_regularization_models} shows the training loss, cross-validation training mean squared error (MSE) with standard deviation, and validation MSE
for different models mapping external force to airspeed.
We can conclude the following.
First, the underlying structure of the data is simple. This is indicated by the fact that increasing the number of neurons in the perceptron's hidden layer
does not significantly improve the fit.
Second, both training and validation errors increase with regularization, indicating that overfitting is unlikely.
This trend could also be caused by insufficient model complexity (expressiveness). However, given the simple structure of the data, this is unlikely to be the case.
Lastly, for this particular model input, the quadratic model is only slightly better than the linear model, as they start converging around $\alpha_1=10^{-3}$.

Next, we compare different inputs to predict airspeed.
\fig{fig:model_regularization_perceptron} shows a comparison for a perceptron with 8 hidden neurons, and \fig{fig:model_regularization_quadratic} depicts a comparison for a quadratic model.
The results show that for the external force input, i.e. model $\vc{v}_r(\vc{f}_{\!e})$, the difference between the models is very small.
Therefore, we recommend using the simpler quadratic model over a perceptron.
Second, we have compared the effect of incorporating the sum of rotor rates into the input, motivated by the blade flapping model \citep{Omari2013}.
For modeling airspeed as a function of force, we expect that a division by the rotor rates is necessary, due to the structure of this physical model.
The results in \fig{fig:model_regularization_perceptron} and \fig{fig:model_regularization_quadratic} confirm this.
Namely, multiplying the external force $\vc{f}_{\!e}$ by the sum of rotor rates $\sum\vc{\varpi}$ increases the MSE over just using the external force.
Converesely, using ${\vc{f}_{\!e} / \sum\vc{\varpi}}$ as input does not have this effect.
The MSE for this model does not increase as fast with regularization, however this may be attributed to scaling in the inputs.

Lastly, we investigated the use of coaxial aerodynamic power $\bar{\vc{P}}_a$ to estimate the airspeed.
The MSE is higher than when using the external force. This can be attributed to the measurement noise of the motor current,
i.e. motor power, which will limit the model performance.
The training and validation MSE show the same behavior as with the external force models,
however somewhat flatter for this particular range of the regularization parameter.
This also indicates a slight scaling issue in the input.
Here, the perceptron model (validation MSE~$\approx0.009$ m/s) performs almost twice as good as the quadratic model (validation MSE~$\approx0.016$ m/s),
since it can fit an underlying nonlinear mapping.
When using aerodynamic power to estimate airspeed, we therefore recommend using a perceptron model.

To summarize, in this subsection we have used $K$-fold cross-validation to analyze the effect of regularization on model generalization.
We have concluded that our model training was set up properly, and that the models are not overfitting to the data.
When using the external force as input, the underlying structure is simple, so a simple quadratic model is sufficient to represent it.
When using aerodynamic power as input, the increased expressivity of perceptron models tends itself better to model the underlying nonlinear physics.
We postulate that current sensing with smaller measurement noise would improve the performance of the latter model, however we cannot confirm this using our data.

\subsection{Aerodynamic torque models}
\label{sec:aero_torque_models}
\begin{figure}
    \centering
    \includegraphics[width=0.9\columnwidth]{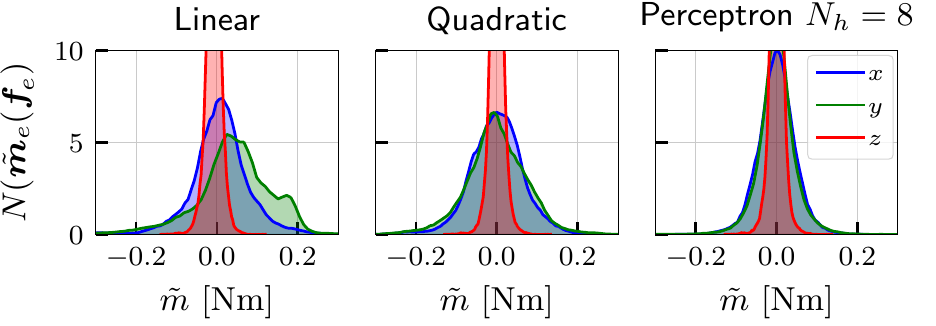}
    \caption{Error histograms for models predicting the aerodynamic torque as a function
        of the aerodynamic force $\vc{m}_d\bigl(\vc{f}_{\!d}\bigr)$, for the validation dataset.
        }
    \label{fig:aerodynamic_torque_histogram}
\end{figure}
\label{sec:aerodynamic_torque_model}
In Section \ref{sec:airspeed_models}, using the complete external wrench as input
did not significantly improve the fit when compared to only using the force.
This indicated that there is a functional relationship between the aerodynamic torque and force.
Physically, this can be understood as the aerodynamic force acting at the center of pressure.
In other words, the torque and force are related by the point of action of the aerodynamic force.

We therefore investigated whether the aerodynamic torque can be modeled as a function of the force.
\fig{fig:aerodynamic_torque_histogram} shows the error histogram for a linear model ${\vc{m}_d:=\mat{D} \vc{f}_{\!d}}$,
a quadratic model ${\vc{m}_d:=\mat{D}_l \vc{f}_{\!d} + \mat{D}_q |\vc{f}_{\!d}| \vc{f}_{\!d}}$,
and a perceptron with 8 hidden neurons.
Because the geometric shape of our robot is asymmetrical, we expect the perceptron model to perform best.
Our results indicate that even the linear model can describe the relationship for our dataset, albeit with a bimodal error distribution for the $y$-axis.
The quadratic model gets rid of this offset and performs significantly better.
Finally, he perceptron shows the best performance, as it can capture the orientation-dependent relationship.
To find the best model, cross-validation should be performed. In the interest of brevity, we skip this step for torque models.
Note that for other flying robots with simpler geometry, even a linear model could describe the relationship with good accuracy.

These models are the basis of force discrimination methods in Section \ref{sec:force_discrimination}.

\section{Physics model based wind estimation}
\label{sec:wind_from_power}

In this section, we present a novel method to obtain wind velocity from
aerodynamic power measurements, based on momentum theory.
The aim is is to provide a measurement that is independent of the external force, in order
to discriminate between aerodynamic and physical interaction forces.
We start by rewriting the aerodynamics of one propeller \eqref{eq:induced_velocity},
\eqref{eq:aerodynamic_power} and \eqref{eq:aerodynamic_power_hover} as a system of
nonlinear equations
${\vc{F}(v_i, v_z, v_{xy}, v_h, P_a) = \vc{0}}$, with ${\vc{F} = [F_1,\, F_2,\, F_3 ]^T}$, where
\begin{equation}
    \begin{aligned}
    F_1 &= v_i^4 - 2 v_i^3 v_z + v_i^2 (v_z^2 + v_{xy}^2) - v_h^4 = 0
    ,
    \\
    F_2 &= v_i U (v_i - v_z) - {P_a}/{(2 \rho A)} = 0
    ,
    \\
    F_3 &=  v_h^2(v_i - v_z) - {P_a}/{(2 \rho A)} = 0
    .
    \end{aligned}
    \label{eq:f_single_prop}
\end{equation}
We consider $P_a / (2\rho A)$ and $v_h$ to be known inputs,
and want to determine ${\vc{x} = [v_x, \, v_y, \, v_z, \, v_i]^T}$.
This system of nonlinear equations is underdetermined, as we have two knowns and three unknowns,
since $v_x$ and $v_y$ are coupled in $v_{xy}$.
Due to this mapping, the solution of \eqref{eq:f_single_prop} will be a manifold, and depends on the initial guess.
Hence, we cannot use \eqref{eq:f_single_prop} to uniquely determine the unknowns.
To solve this problem, we expand the system of equations to include multiple measurements.
We then introduce a transformation of \eqref{eq:f_single_prop} into a common frame.
This allows us to estimate all three wind velocity components and the propeller induced velocities
by solving a nonlinear least squares (NLS) problem.

\textbf{Multiple measurements.}
Let us assume a constant wind velocity ${\vc{v}_w = [v_{x},\, v_y, \,v_z]^T}$ through $N$ measurements.
This assumption holds in several cases.
First, we can combine instantaneous measurements from multiple propellers that are rigidly attached
(e.g. quadcopter). These may also be rotated w.\,r.\,t. the body frame.
Second, we can combine measurements from multiple poses at different time instants in a small time window.
Third, if the flight is not aggressive, i.\,e. the orientation does not change significantly, we can estimate the body-frame freestream velocity.
In effect, we use information gained from $N$ measurements to obtain the wind velocity components.

We may extend the state to $N$ measurements
\begin{equation}
    \vc{x}\rvert_N = [v_{x}, \, v_{y}, \, v_{z}, \, v_{i,2}, \, v_{i,2}, \, \ldots \,v_{i,N}]^T
    ,
\end{equation}
and solve the extended system of equations
\begin{align*}
        &\vc{F}\rvert_N(v_{x}, v_y, v_z, v_{i,1}, v_{h,1}, P_{a,1}, \ldots, v_{i,N}, v_{h,N}, P_{a,N})
    = \vc{0},
    \\
    &\vc{F}\rvert_N = [F_{1,1},\, F_{2,1},\, F_{3,1},\,
    \,\ldots, \,F_{1,N}, \,F_{2,N}, \,F_{3,N}]^T
    ,
    \numberthis
    \label{eq:f_multiple_measurements}
\end{align*}
where $F_{1,k}$, $F_{2,k}$ and $F_{3,k}$ are evaluations of
\eqref{eq:f_single_prop} for the $k$-th measurement.
A Jacobian is needed to solve \eqref{eq:f_multiple_measurements}.
The Jacobian for the $k$-th measurement is defined as
\begin{equation}
    \mat{J}_k = \begin{bmatrix}
        J_{11,k} & J_{12,k} & J_{13,k} & J_{14,k} \\
        J_{21,k} & J_{22,k} & J_{23,k} & J_{24,k} \\
        J_{31,k} & J_{32,k} & J_{33,k} & J_{34,k}
    \end{bmatrix}
    ,
    \label{eq:f_jacobian}
\end{equation}
where ${J_{ij,k} = \partial F_{i,k} / \partial x_{j,k}}$.
We can now construct the extended Jacobian ${\mat{J}\rvert_N \in \Realm{3N}{N+3}}$.
For three measurements we have
${\vc{x}\rvert_{3} = [v_{x},\, v_{y}, \, v_{z}, \, v_{i,1},\, v_{i,2},\, v_{i,3}]^T}$ and
\begin{equation*}
    \mat{J}\rvert_{3} = \left[
        \begin{array}{cccc:c:c}
            J_{11,1} & J_{12,1} & J_{13,1} & J_{14,1} & 0 & 0 \\
            J_{21,1} & J_{22,1} & J_{23,1} & J_{24,1} & 0 & 0 \\
            J_{31,1} & J_{32,1} & J_{33,1} & J_{34,1} & 0 & 0 \\
            \hdashline
            J_{11,2} & J_{12,2} & J_{13,2} & 0 & J_{14,2} & 0 \\
            J_{21,2} & J_{22,2} & J_{23,2} & 0 & J_{24,2} & 0 \\
            J_{31,2} & J_{32,2} & J_{33,2} & 0 & J_{34,2} & 0 \\
            \hdashline
            J_{11,3} & J_{12,3} & J_{13,3} & 0 & 0 & J_{14,3} \\
            J_{21,3} & J_{22,3} & J_{23,3} & 0 & 0 & J_{24,3} \\
            J_{31,3} & J_{32,3} & J_{33,3} & 0 & 0 & J_{34,3}
        \end{array}
    \right]
    ,
\end{equation*}
which is straightforward to extend to $N$ measurements.
Notice that the first three columns are due to the three airspeed components,
which are assumed equal across measurements.
The other columns are due to the induced velocity, which is different between measurements.

\textbf{Transformed formulation.}
When combining measurements from different poses, the wind velocity has to be transformed
into a common coordinate frame. Otherwise, the constant wind velocity assumption will not hold.
Define the freestream velocity of propeller $k$ as
\begin{equation}
    \vc{v}_{k} = 
    \begin{bmatrix}
        v_{x,k} \\
        v_{y,k} \\
        v_{z,k}
    \end{bmatrix}
    =
    \mat{R}_k
    \begin{bmatrix}
        v_{x} \\
        v_{y} \\
        v_{z}
    \end{bmatrix}
    +
    \vc{v}_{0,k}
    =
    \mat{R}_k \vc{v} + \vc{v}_{0,k}
    ,
    \label{eq:freestream_velocity_transformed}
\end{equation}
and use the transformed velocities when calculating \eqref{eq:f_single_prop} and \eqref{eq:f_jacobian}.
We assume that the robot is moving between measurements.
Therefore, the offset velocity $\vc{v}_{0,k}$ can be obtained from a pose estimation system as the relative velocity of the robot between two measurements.
We may also use the propeller offset velocity due to the
body angular velocity, i.\,e. ${\vc{v}_{0,k} = \mat{R}_{pb,k} \vc{\omega} \times \vc{r}_k}$,
where $\mat{R}_{pb,k}$ is the rotation from the body to the $k$-th propeller frame.
The Jacobian of this formulation can be found as \eqref{eq:jacobian_elements} in Section \ref{sec:jacobian}.

This formulation allows us to determine all three components of the freestream velocity independently.
It also can be used to obtain the instantaneous wind velocity components when the propellers are not mounted to the multicopter frame in a coplanar configuration.

\begin{figure}
    \centering
    \subfigure[No noise in $P_a$, contours at real $v_i$]
        {\includegraphics[width=.45\columnwidth]{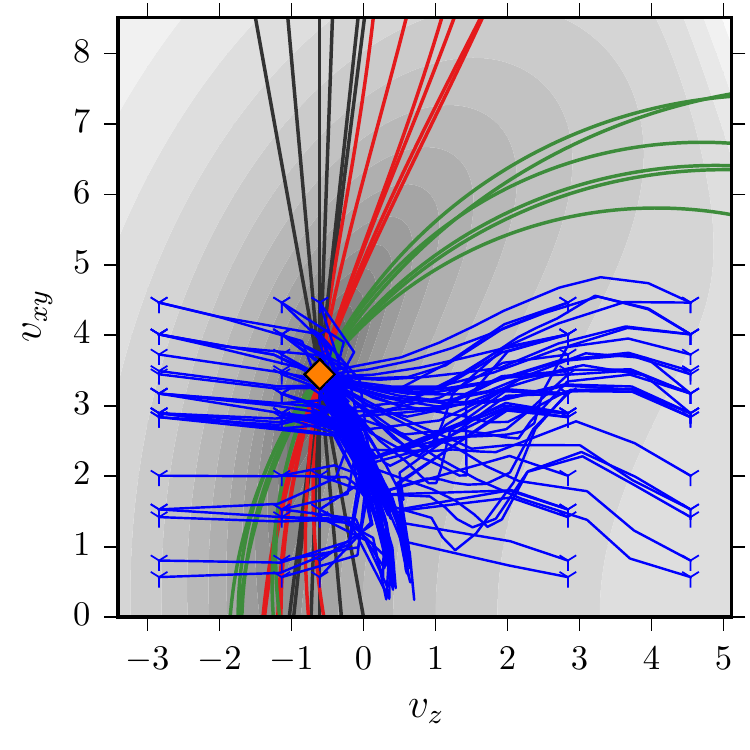}
        \label{fig:mesh_clean}}
    \hfill
    \subfigure[Noise in $P_a$, contours at converged $v_i$]
        {\includegraphics[width=.45\columnwidth]{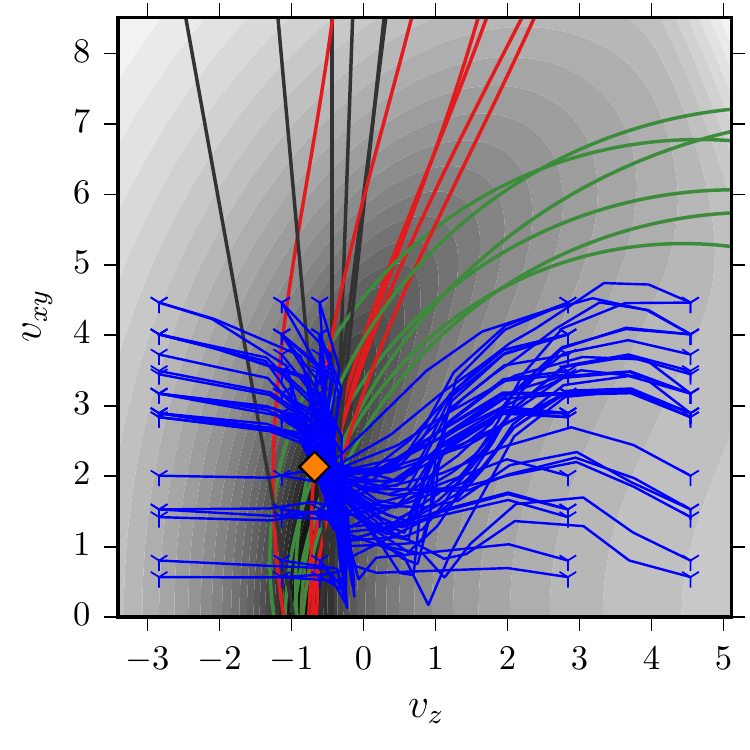}
        \label{fig:mesh_deltaPa}}
    \caption{Zero contours of \eqref{eq:minim_goal_function}, with color indicating the function value.
        Green lines are contours of $F_1^{(k)}$, magenta lines are contours of $F_2^{(k)}$,
        and black lines are contours of $F_3^{(k)}$ at ${v_i = \mathrm{const}}$.
        The blue lines show convergence of a Levenberg-Marquardt solver for
        different initial guesses. The converged solution is depicted as a red cross.
        The velocity components $v_x$ and $v_y$ lumped into $v_{xy}$.
        We used ${N=6}$ measurements, velocity ${v_{\infty}=3.5}$\,m/s, angle of attack ${\alpha=10^{\circ}}$,
        ${v_h \in [4.1 \ldots\,6.7]}$\,m/s, and measurement angles up to $10^{\circ}$.
        Without noise on the power measurement (\fig{fig:mesh_clean}), the solution converges to the
        exact wind velocity ${\vc{v}_w = [-3.45,\,0\,-0.61]^T}$\,m/s.
        With noise in the power measurements, the least-squares solution moves depending on measurement conditioning.
    }
    \label{fig:mesh_goal_function}
    \vspace{-10pt}
\end{figure}

\textbf{Solving the system of equations.}
In perfect conditions, the solution to \eqref{eq:f_multiple_measurements} will be at the intersection of
all nonlinear functions, where $\vc{F} = \vc{0}$.
This corresponds to a multidimensional root-finding problem.
However, when the measured aerodynamic power does not match momentum theory (i.\!e. under model mismatch), the functions will not necessarily intersect.
We therefore want to find the point that is closest to all functions.
In this case we have to solve a nonlinear least squares problem with the objective function
\begin{equation}
    f = \tfrac{1}{2} \vc{F}^T \vc{F}
    ,
    \label{eq:minim_goal_function}
\end{equation}
for example using a Levenberg-Marquardt solver \cite{Nocedal2006,Boyd2004}.

When an exact solution exists, it will be at ${f=0}$, i.\,e. the intersection of ${\vc{F}=\vc{0}}$.
Otherwise, if there is a model mismatch or noise in $P_a$, we get a least squares solution.
\fig{fig:mesh_goal_function} shows convergence of the solver for different initial guesses and noise on $P_a$.

\textbf{Limiting the search space.}
The space of \eqref{eq:minim_goal_function} can contain local optima.
From the underlying physics, the same measured power can be obtained by various wind and induced velocities.
The optimized variables are velocities. We may therefore use physical considerations to determine the set of feasible solutions.
A flying robot must expend power to generate thrust, which implies ${T>0}$ and ${P_a>0}$,
for which we use \eqref{eq:thrust_momentum_theory} and \eqref{eq:aerodynamic_power}, respectively.
The induced velocity is ${v_i<v_h}$ in the normal working state, and ${v_i > v_h}$ in the VRS.
We exclude VRS from the search space because momentum theory is invalid in that state.
Therefore, we limit the induced velocity to ${0 < v_i < v_h}$.
Likewise, we can limit $\vc{v}_w$ in case its maxima are known.
In order to limit the search space using the Levenberg-Marquardt method, we add a quadratic barrier function $F_4$ to
the optimization problem formulation \cite{Nocedal2006}.
which increases the size of the problem, as the function becomes
${\vc{F} \in \Realv{4N}}$, and the Jacobian becomes
${\mat{J}\rvert_N \in \Realm{4 N}{N+3}}$.

\textbf{Normalization.}
In order to improve stability of the numerical solution,
we normalize the goal function to its initial value $f_0$, i.\,e. we minimize ${{f}' = f_0^{-1} f}$.
Furthermore, the functions $F_{1\ldots 3}^{(k)}$ are normalized to $v_h^{(k)}$, such that
${{F}_1' = {F_1}/v_h^{4}}$,
${{F}_2' = {F_2}/{v_h^{3}}}$,
${{F}_3' = {F_3}/{v_h^{3}}}$,
${{F}_4' = {F_4}/{v_h^{2}}}$.
In this way, the function values are dimensionless and have the same order of magnitude.

\begin{figure}
    \centering
    \includegraphics[width=\columnwidth]{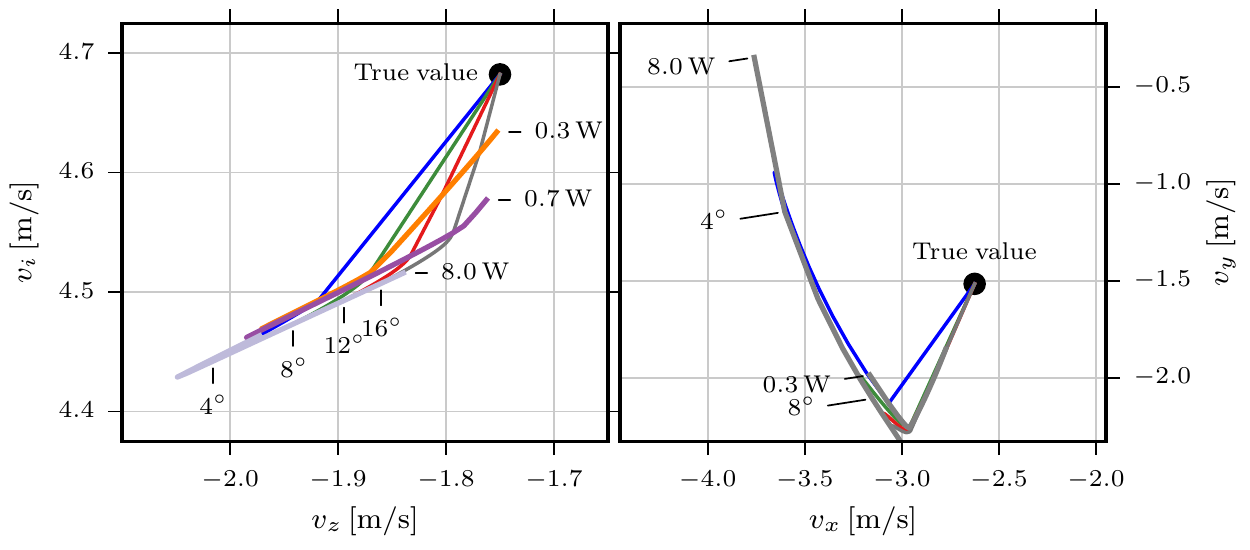}
    \caption{Sensitivity of the NLS solution to error in aerodynamic power
        $\tilde{P}_a$ up to 8\,W, using ${N=12}$ measurements, and maximum measurement angles
        $\sigma_{\alpha}$ up to ${20^{\circ}}$.
        For ${\sigma_{\alpha}=0^{\circ}}$, the solution diverges out of the depicted range and is not shown.
        Wind speed is chosen to be ${v_{\infty}=3.5}$\,m/s.
        Larger measurement angles lead to a more robust solution, as the estimated wind velocity is
        closer to the real value even for high errors in the aerodynamic power.
        The vertical wind component $v_z$ and propeller induced velocity $v_i$ are estimated
        with good accuracy for a wide range of $\tilde{P}_a$.
        However, the horizontal wind components $v_x$, $v_y$ diverge from their real values
        even for low $\tilde{P}_a$.
    }
    \label{fig:sensitivity_convergence_alpha}
    \vspace{-10pt}
\end{figure}

\begin{figure}
    \centering
    \includegraphics[width=.75\columnwidth]{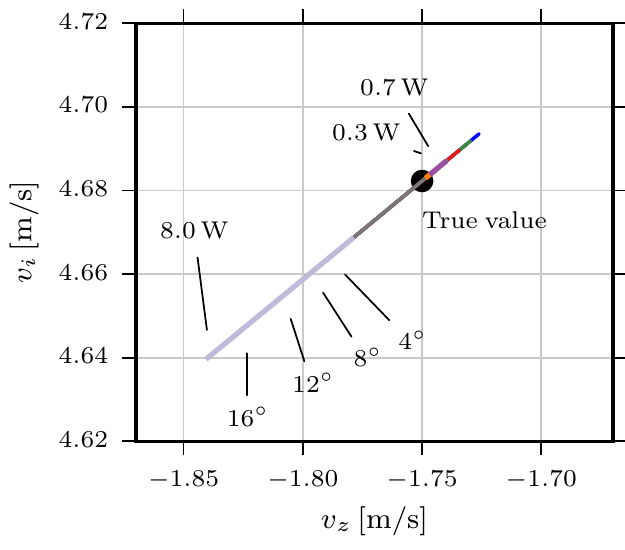}
    \caption{Sensitivity analzsis of the NLS solution for the same case as in \fig{fig:sensitivity_convergence_alpha},
        however for perfect knowledge of horizontal velocity components ($v_x$, $v_y$).
        These may be obtained from the induced drag model, i.\,e. the external force.
        When the horizontal wind velocity is known, the vertical component maz be determined
        robustly from aerodynamic power measurements.
    }
    \label{fig:sensitivity_convergence_alpha_vxyfixed}
    \vspace{-12pt}
\end{figure}

\textbf{Sensitivity analysis.}
Measurement noise will shift the estimated wind velocity in a nonlinear manner, see \fig{fig:mesh_goal_function}.
A sensitivity analysis helps estimating this effect.
\fig{fig:sensitivity_convergence_alpha} depicts the converged solutions for increasing
noise amplitude in the measured power.
Since the quality of the solution will depend on the distribution of measurement poses,
we uniformly distribute these under different maximum angles, from 5$^{\circ}$ to 20$^{\circ}$.
Higher relative angles between measurement poses increase robustness of the solution.
However, estimation of the horizontal wind velocity components is very sensitive to power measurements.

We therefore propose to estimate the horizontal velocity components using the induced drag model,
i.\,e. from the external force.
As shown in \fig{fig:sensitivity_convergence_alpha_vxyfixed}, this allows a
robust estimation of the vertical wind velocity component and the propeller induced velocity
even for a high error in power measurements.
A minimum angular distance between measurements should also be considered when choosing suitable measurements for the NLS problem.
Having an offset velocity $\vc{v}_0$ in \eqref{eq:freestream_velocity_transformed} additionally reduces sensitivity to noise in the power measurements.

\begin{figure}
    \centering
    \includegraphics[width=\columnwidth]{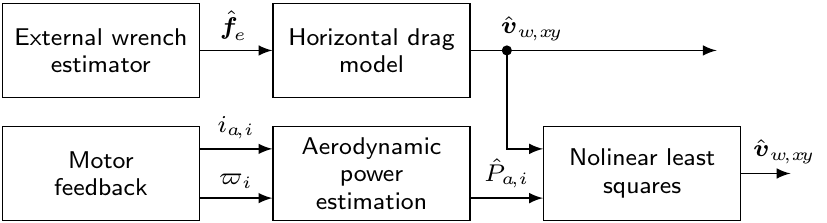}
    \caption{
        The combined estimation scheme calculates the horizontal wind velocity
        from the external wrench. This is then used as an input to the
        wind velocity estimator using the motor power, which leads to improved accuracy.
    }
    \label{fig:hybrid_wind_estimation}
\end{figure}

\textbf{Combined wind estimator.}
In order to overcome limitations of the two presented methods, see Section \ref{sec:aero_model_evaluation} and Section \ref{sec:wind_from_power}, we finally propose a combined wind estimator, see \fig{fig:hybrid_wind_estimation}.
The horizontal velocity components ($v_x$, $v_y$) are obtained from the external wrench or aerodynamic power, using models identified in Section~\ref{sec:aero_model_evaluation}.
We then use the estimated aerodynamic power and known ($v_x$, $v_y$) to calculate
$v_z$ using the nonlinear least squares formulation, by minimizing \eqref{eq:minim_goal_function}.

\begin{figure*}
    \centering
    \includegraphics{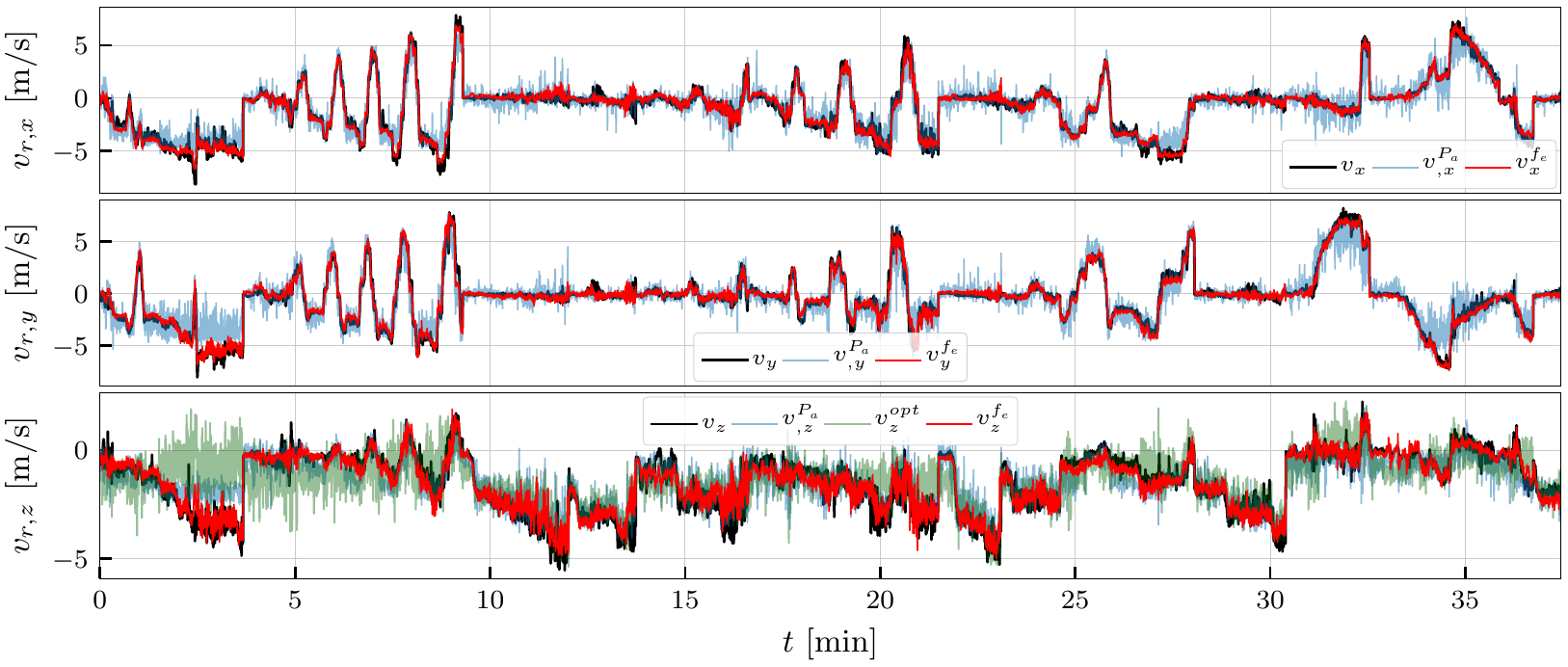}
    \caption{Estimation of relative airspeed for the complete dataset from \fig{fig:training_set},
        using machine learning models ${\vc{v}_r(\vc{f}_e \sum \varpi)}$ (shown as $v^{f_e}$),
        ${\vc{v}_r(\vc{P}_{\!a})}$ (shown as $v^{P_a}$), and the optimization based combined wind estimator described in this section (shown as $v^{opt}$).
        For the latter, the horizontal velocity was obtained from motor power.}
    \label{fig:power_based_estimation_timeseries}
\end{figure*}
\begin{figure}
    \centering
    \includegraphics{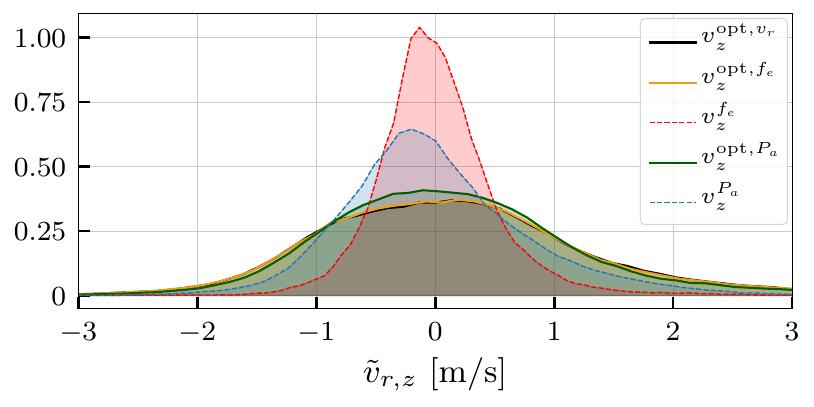}
    \caption{Normalized error histograms of the vertical airspeed component $v_z$ for the dataset from \fig{fig:training_set}.
        For the combined wind estimator, we compare different sources of horizontal airspeed components:
        The model ${\vc{v}_r(\vc{f}_e \sum \varpi)}$ is used in $v_z^{opt,f_e}$.
        The model ${\vc{v}_r(\vc{P}_{\!a})}$ is used in $v_z^{opt,P_a}$.
        Lastly, we use the exact airspeed in $v_z^{opt,v_r}$.}
    \label{fig:power_based_estimation_histogram}
\end{figure}

\subsection{Comparison to data-driven estimation}
Next, we compare the optimization-based combined wind estimator to machine learning based models.
We also compare different sources of horizontal velocity in the optimization.
\fig{fig:power_based_estimation_timeseries} shows components of the airspeed for the complete wind tunnel dataset.
Ground truth is shown in black; estimation using the external force and rotor speed $\vc{v}_r(\vc{f}_e/\sum{\vc{\varpi}})$ is shown in red;
estimation using aerodynamic power of the coaxial pairs $\vc{v}_r(\bar{\vc{P}}_a)$ is shown in light blue.
Finally, for the $z-$component we also show the velocity obtained by the combined estimator in light green as $v_z^{opt}$.
In this case we used the $\vc{v}_r(\bar{\vc{P}}_a)$ model to obtain the horizontal airspeed components.
The result therefore uses only motor power to estimate airspeed.
The result mirrors conclusions from Section~\ref{sec:aero_model_evaluation}.
The mean of all models follows the ground truth, however when using motor power, the result is noisier.
This is due to the noisy motor current measurements in our experiments.
Second, large errors occur where the aerodynamic power cannot be estimated correctly due to motor saturation, such
as at the beginning of the dataset.

For the optimization based combined estimator, we used a window of 3 measurements, utilizing all 3 motor pairs, for a total of 9 measurements per optimization.
As these are close in time and orientation, the results become inaccurate and noisy with increasing horizontal airspeed components.
The result is otherwise comparable to the machine learning based method, however the latter outperforms the optimization slightly.
This shows that the optimization method can be used to obtain airspeed from real in-flight measurements.

\fig{fig:power_based_estimation_histogram} compares how utilizing different sources of horizontal airspeed for the combined estimator affects prediction accuracy.
A histogram is shown only for the $z-$component of the relative airspeed.
As a benchmark, we show the airspeed obtained by using the external force as $v_z^{f_e}$. It has the smallest variance in comparison.
The machine learning model $\vc{v}_r(\bar{\vc{P}}_a)$ is shown as $v_z^{P_a}$, and has the second smallest variance.
Using the ground truth horizontal velocity in the combined wind estimator is shown as $v_z^{\mathrm{opt},v_r}$.
Using the external force model for this purpose is depicted as $v_z^{\mathrm{opt},f_e}$.
Finally, the histogram for the result shown in \fig{fig:power_based_estimation_timeseries} is depicted as $v_z^{\mathrm{opt},P_a}$.
Interestingly, the combined estimator is not sensitive to the source of horizontal airspeed.
However, the variance of the estimated velocity is higher than when using machine learning based models.

In conclusion, we have shown that airspeed may be obtained using motor power measurements only.
We have described the underlying physics of the problem in the optimization based approach.
Accuracy of the estimation will clearly be limited by the accuracy of the estimated aerodynamic power.
Therefore, having a good measurement of the motor current and a good estimate of the aerodynamic power are
crucial to the applicability of this approach.

\section{Discrimination between aerodynamic and interaction forces}
\label{sec:force_discrimination}

Now that the dynamics and aerodynamic models have been identified, we investigate the problem of discriminating between the different terms of the external wrench acting on the robot.
In this section, we present several novel methods for input discrimination in the context of the awareness pipeline presented in \cite{Tomic2017isrr}.
The awareness pipeline is a generalization of Fault Detection, Identification and Isolation \cite{Haddadin2008, Haddadin2014}.
\emph{Detection} provides a binary signal (true/false) whether a signal is present.
We briefly review collision detection, and provide a novel method for \emph{contact detection under wind influence}.
\emph{Isolation} deals with determination of the kind and location of the signal.
An example is obtaining the contact or collision position on the robot's convex hull upon detecting either.
Lastly, \emph{identification} deals with monitoring the time-variant behavior of the signal.
In this context, this is the reconstruction of the signal of interest.

\textbf{Problem statement.}
The goal of input discrimination is outlined in \fig{fig:input_discrimination_intro}:
given an external wrench $\vc{\tau}_{\!e}$ that is a sum of the fault wrench $\vc{\tau}_{\!f}$ (e.g. caused by collisions),
the physical interaction wrench $\vc{\tau}_{\!i}$ (e.g. caused by a person pushing the vehicle),
and the disturbance wrench $\vc{\tau}_{\!d}$ (caused by wind), obtain a reconstruction of the constituent terms.
In other words, given
\begin{equation}
    \vc{\tau}_{\!e} = \vc{\tau}_{\!f} + \vc{\tau}_{\!i} + \vc{\tau}_{\!d},
\end{equation}
force discrimination deals with detecting the presence of a signal, extracting context-dependent information (isolation),
and obtaining the time-varying reconstruction of the signal (isolation).
In the following, we assume that the external wrench $\tauext$ is perfectly known.
When using the external wrench estimator this means that $\tauexthat \approx \tauext$.

\subsection{Collision detection under wind influence}
\label{sec:collision_detection}
We briefly review collision detection, which has been discussed in more detail in \cite{Haddadin2014,DeLuca2006}, and in \cite{Tomic2015} and \cite{Tomic2017} for aerial robots in particular.
Discrimination between aerodynamic and \emph{collision} forces may be achieved by considering the respective signals' frequency characteristics.
The frequency ranges of the constituent terms are used to design appropriate filters.
Collisions may be detected by applying a highpass filter $H(f)$ on the external force, with break frequency $\omega_f$ to obtain the \emph{collision detection signal}
\begin{equation}
    C\!D = \begin{cases}
        1   \quad \mathrm{if} \;\; \exists i : \; H(| \hat{f}_{e,i} |, \omega_f) > f_{c,i} \\
        0   \quad \mathrm{otherwise},
    \end{cases}
\end{equation}
where $f_{c,i}$ is the collision detection threshold.
The threshold is be state-dependent, as increasing airspeed will also increase noise in the external wrench.
Note that quasistatic contact forces cannot be distinguished from aerodynamic forces from the frequency content alone.
A method is needed to distinguish consituent terms of the external wrench in the same frequency range.

\subsection{Obtaining the contact location}
\label{sec:contact_position}
First we briefly revisit how to obtain the contact location $\vc{r}_c$ from a contact force $\vc{f}_c$ and torque $\vc{m}_c$,
as presented in \cite{Tomic2014c} and \cite{Tomic2017}.
This may be used after discriminating the interaction force and torque by any presented method to obtain the contact position.
By using the torque model ${\vc{m}_c = \vc{r}_c \times \vc{f}_{\!c}}$, the contact location
is obtained by intersecting the ray
\begin{equation}
    \vc{r}_c = \vc{o} + k \vc{d},
    \quad
    \vc{o} = \frac{\vc{f}_{\!c} \times \vc{m}_c} {\vc{f}_{\!c} ^T \vc{f}_{\!c}},
    \quad
    \vc{d} = \vc{f}_{\!c},
    \label{eq:contact_position}
\end{equation}
with the vehicle's convex hull to obtain the unknown parameter $k$.
An intersection with the hull is required in order to obtain a unique solution.
Note that this problem has two solutions when the ray passes through the hull.
\begin{figure}[h]
    \centering
    \includegraphics{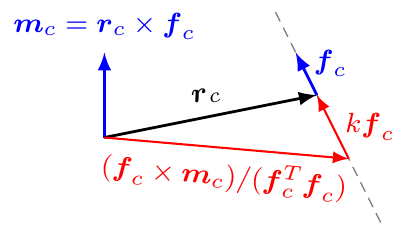}
    \caption{The collision position $\vc{r}_c$ can be obtained from the external force $\vc{f}_{\!c}$ and torque $\vc{m}_c$
        through the line of action of $\vc{f}_{\!c}$.
        All forces lying on the dashed ray produce the same torque.
        The free parameter $k$ is then found by intersecting the ray with the robot's convex hull.
    }
    \label{fig:inverse_cross_product}
\end{figure}

\subsection{Contact detection under wind influence}
\label{sec:contact_detection}
Contact detection is based on the aerodynamic torque model as a function of the aerodynamic force $\hat{\vc{m}}_{d}(\vc{f}_{\!d})$, see Section~\ref{sec:aero_torque_models}.
When only the aerodynamic force is acting on the robot, the external torque $\vc{m}_e$ will match the aerodynamic torque model,
i.e. $\hat{\vc{m}}_{d}(\vc{f}_{\!e}) = \vc{m}_e$, up to modeling errors.
If there is another torque-generating force acting on the robot, the external torque will not match the model.
Based on this insight we define the \emph{aerodynamic torque residual} $\tilde{\vc{m}}_d$ as
\begin{equation}
    \tilde{\vc{m}}_d = \hat{\vc{m}}_{d} \bigl( \vc{f}_{\!e} \bigr) - \vc{m}_e.
    \label{eq:torque_residual}
\end{equation}
The assumption of the nominal state is that only the aerodynamic wrench is acting on the robot.
The identified aerodynamic torque model is evaluated at the current external force $\vc{f}_{\!e}$.
The residual $\tilde{\vc{m}}_d$ will be nonzero if there is an additional wrench acting on the robot that does not correspond to the aerodynamics model.
The \emph{contact detection signal} $C\!D_1$ may now be defined as
\begin{equation}
    C\!D_1 = \begin{cases}
        1   \quad \mathrm{if} \; \| \tilde{\vc{m}}_d \| > \delta_{d} \\
        0   \quad \mathrm{otherwise},
    \end{cases}
\end{equation}
where $\delta_d$ is the threshold on the residual norm.

\textbf{Failure cases}.
The aerodynamic torque model may be as simple as a linear combination ${\hat{\vc{m}}_{d}(\vc{f}_{\!d}) := \mat{D} \vc{f}_{\!d}}$,
i.e. the aerodynamic force acting at a center of pressure.
Assume
\begin{equation}
    \begin{aligned}
        \vc{f}_{\!e} &= \vc{f}_{\!d} + \vc{f}_{\!i}, \\
        \vc{m}_e &= \mat{D} \vc{f}_{\!d} + \vc{m}_i,
    \end{aligned}
    \label{eq:torque_residual_failure_assumptions}
\end{equation}
the aerodynamic model is perfectly known,
and $\vc{f}_{\!i} \neq \vc{0}$, $\vc{m}_i \neq \vc{0}$, and $\tilde{\vc{m}}_d =\vc{0}$.
Then, by applying \eqref{eq:torque_residual_failure_assumptions} to \eqref{eq:torque_residual},
\begin{equation*}
    \mat{D} \vc{f}_{\!d} + \mat{D} \vc{f}_{\!i} - \mat{D} \vc{f}_{\!d} - \vc{m}_i = \vc{0}.
\end{equation*}
This means that the contact detection scheme fails if
\begin{equation*}
    \mat{D} \vc{f}_{\!i} = \vc{m}_i.
\end{equation*}
In other words, the scheme does not work if the exerted wrench exactly matches the aerodynamics model,
i.e. is indistinguishable from aerodynamic effects.
In the nonlinear case, the equivalent failure is
\begin{equation*}
    \vc{0} = \hat{\vc{m}}_{d} \bigl( \vc{f}_{\!d} + \vc{f}_{\!i} \bigr) - \vc{m}_d\bigl( \vc{f}_{\!i} \bigr) - \vc{m}_i.
\end{equation*}

\textbf{Limitations.}
\begin{figure*}
    \centering
    \subfigure[${\vc{f}_{\!i} = [1,\ 0,\ 0]^T}$ N]{%
        \begin{tikzpicture}[>=latex, draw=black, very thick, line cap=round]
            \node[inner sep=0pt, anchor=south west,opacity=1.0] (image) at (0,0)
            {\includegraphics[width=0.5\columnwidth]{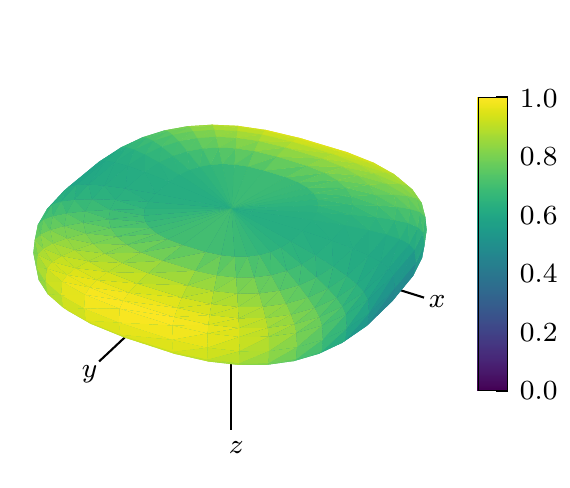}};
            \begin{scope}[x={(image.south east)}, y={(image.north west)}]
                \draw[->] (0.41, 0.575) ++(-20:-1.1cm) -- ++(-20:1.1cm);
                \draw[->] (0.41, 0.575) ++(40:.25cm) ++(-20:-1.1cm) -- ++(-20:1.1cm);
                \draw[->] (0.41, 0.575) ++(40:-.25cm) ++(-20:-1.1cm) -- ++(-20:1.1cm);
            \end{scope}
        \end{tikzpicture}%
        \label{fig:residual_observability_1_0_0}}
    \subfigure[${\vc{f}_{\!i} = [0,\ 1,\ 0]^T}$ N]{%
        \begin{tikzpicture}[>=latex, draw=black, very thick, line cap=round]
            \node[inner sep=0pt, anchor=south west,opacity=1.0] (image) at (0,0)
            {\includegraphics[width=0.5\columnwidth]{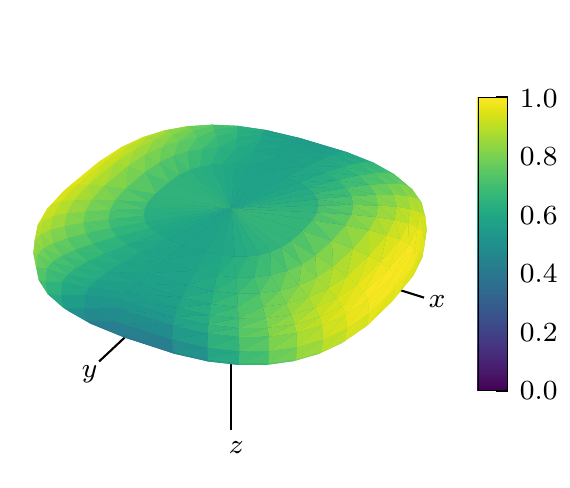}};
            \begin{scope}[x={(image.south east)}, y={(image.north west)}]
                \draw[->] (0.41, 0.575) ++(220:-1cm) -- ++(220:1cm);
                \draw[->] (0.41, 0.575) ++(170:.25cm) ++(220:-1cm) -- ++(220:1cm);
                \draw[->] (0.41, 0.575) ++(170:-.25cm) ++(220:-1cm) -- ++(220:1cm);
            \end{scope}
        \end{tikzpicture}%
        \label{fig:residual_observability_0_1_0}}
    \subfigure[${\vc{f}_{\!i} = [0,\ 0,\ 1]^T}$ N]{%
        \begin{tikzpicture}[>=latex, draw=black, very thick, line cap=round]
            \node[inner sep=0pt, anchor=south west,opacity=1.0] (image) at (0,0)
            {\includegraphics[width=0.5\columnwidth]{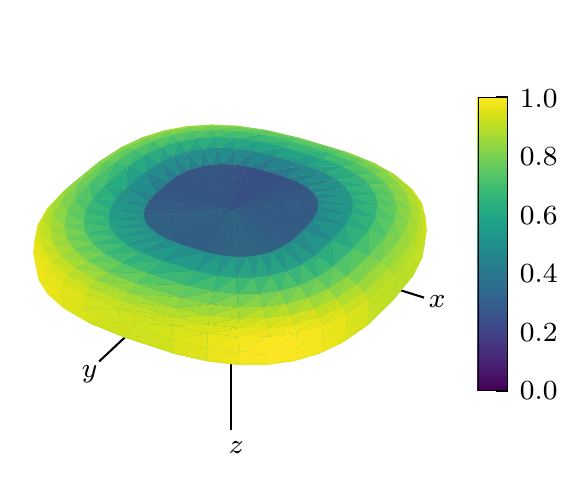}};
            \begin{scope}[x={(image.south east)}, y={(image.north west)}]
                \draw[->] (0.41, 0.575) ++(0, 1cm) -- ++(270:1cm);
                \draw[->] (0.41, 0.575) ++(20:.25cm) ++(0, 1cm) -- ++(270:1cm);
                \draw[->] (0.41, 0.575) ++(20:-.25cm) ++(0, 1cm) -- ++(270:1cm);
            \end{scope}
        \end{tikzpicture}%
        \label{fig:residual_observability_0_0_1}}
    \subfigure[${\vc{f}_{\!i} = [1,\ 0,\ 1]^T}$ N]{%
        \begin{tikzpicture}[>=latex, draw=black, very thick, line cap=round]
            \node[inner sep=0pt, anchor=south west,opacity=1.0] (image) at (0,0)
            {\includegraphics[width=0.5\columnwidth]{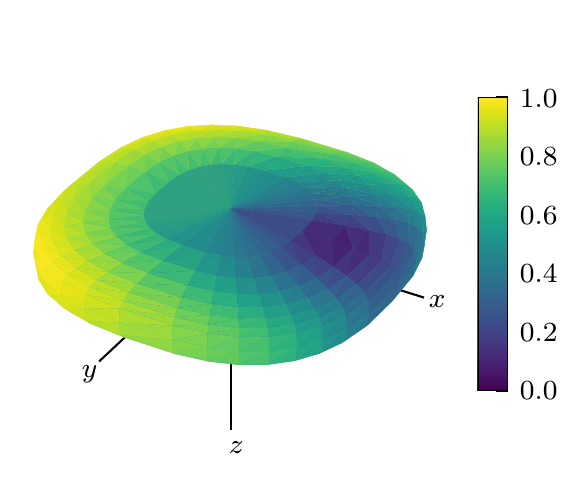}};
            \begin{scope}[x={(image.south east)}, y={(image.north west)}]
                \draw[->] (0.41, 0.575) ++(120:1cm) -- ++(-60:1cm);
                \draw[->] (0.41, 0.575) ++(-20:.35cm) ++(120:1cm) -- ++(-60:1cm);
                \draw[->] (0.41, 0.575) ++(-20:-.35cm) ++(120:1cm) -- ++(-60:1cm);
            \end{scope}
        \end{tikzpicture}%
        \label{fig:residual_observability_1_0_1}}
    \caption{Residual for constant forces at positions around a superellipsoidal convex hull.
        A contact force will only be detected by the residual if it generates a torque.
        The color represents $\|\tilde{\vc{m}}_d\|$ and is normalized to the highest residual being equal to 1.
        We used a linear aerodynamic torque model $\vc{m}_{d}(\vc{f}_{\!d})$,
        and a quadratic aerodynamic wrench model with a relative airspeed of $\vc{v}_r = [1.0,\ 5.0,\ 0.2]^T$ m/s.
        }
    \label{fig:residual_observability}
\end{figure*}
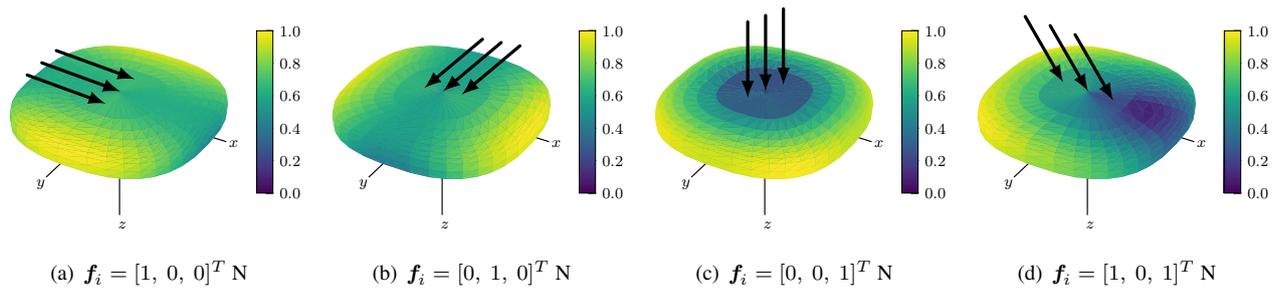
The proposed detection scheme relies on the torque produced by the interaction.
Therefore, as discussed above, not all combinations of force and torque will generate a residual.
The detection will be local, depending on the convex hull of the robot, and the aerodynamic model.
\fig{fig:residual_observability} illustrates the residual norm when the contact occurs on the convex hull of the robot.
Clearly, some forces generate a stronger signal than others.
In the cases \fig{fig:residual_observability_1_0_0} and \fig{fig:residual_observability_0_1_0}, contact positions
that generate a torque about the $z-$axis will generate the strongest residual.
Conversely, a purely vertical force as in \fig{fig:residual_observability_0_0_1} will produce no torque,
making it difficult to distinguish from wind when acting on the top or bottom of the hull.
Similarly, in the case \fig{fig:residual_observability_1_0_1}, when the force is acting at particular contact positions,
it will not be distinguishable from wind, as described in the failure cases.

\begin{figure*}
    \centering
    \includegraphics{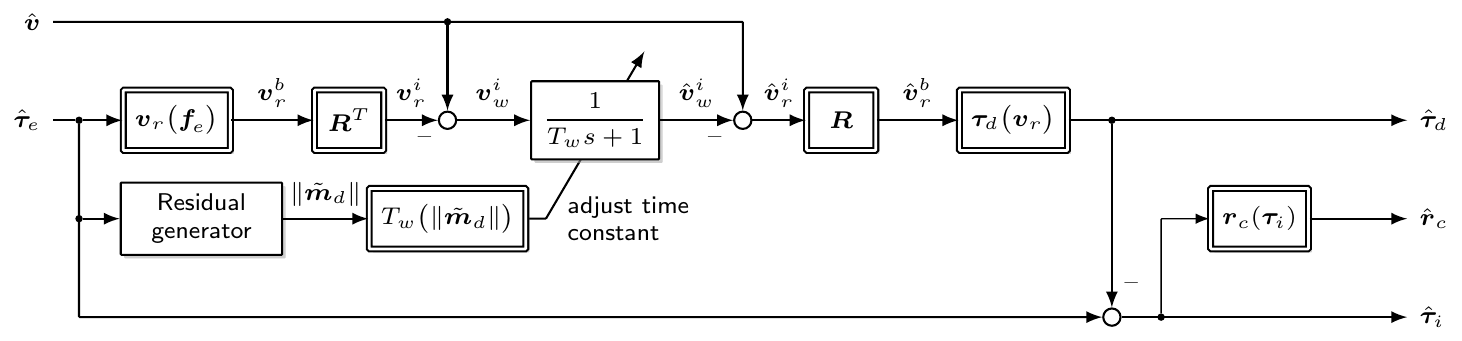}
    \caption{Modified model checking scheme for simultaneous wind and interaction estimation. We estimate the wind speed in the inertial
        frame by filtering the raw prediction made by an aerodynamics model learned in Section \ref{sec:aero_model_evaluation}.
        Once contact is detected, we decrease the time constant of the wind estimator and treat everything else as the interaction force.}
    \label{fig:residual_wind_estimation}
\end{figure*}

\subsection{Modified model-checking force discrimination scheme}
\label{sec:residual_discrimination}
Based on the contact detection signal, we propose the force discrimination scheme depicted in \fig{fig:residual_wind_estimation}.
This scheme makes the assumption that wind speed does not change significantly during physical interaction.
It estimates the wind speed in the inertial frame with a filter (e.g. Kalman filter).
For simplicity of argumentation, in \fig{fig:residual_wind_estimation} this is a simple first-order lowpass filter.
It is an modification of the model-checking scheme presented in \cite{Tomic2015} with a time-varying wind estimation time constant.
Once the contact detection signal $CD_1$ becomes true, the time constant $T_w$ of the wind estimator (or process noise in the case of a Kalman filter) is slowed down.
This essentially pauses wind speed estimation by slowing down wind estimation.
By reprojecting the estimated wind speed $\hat{\vc{v}}_w$ back into the body frame to get the relative airspeed $\hat{\vc{v}}_r$,
we obtain the estimated aerodynamic wrench ${\hat{\vc{\tau}}_d := \vc{\tau}_d(\hat{\vc{v}}_r)}$ by passing the airspeed through the aerodynamic model.
Note that this is similar to the model checking scheme in \cite{Tomic2015}.
We then obtain the interaction wrench from the forward aerodynamic model as ${\hat{\vc{\tau}}_i = \vc{\tau}_e - \hat{\vc{\tau}}_d}$.
However, it does not make any assumptions about the interaction force itself.

\begin{figure*}
    \centering
    \includegraphics{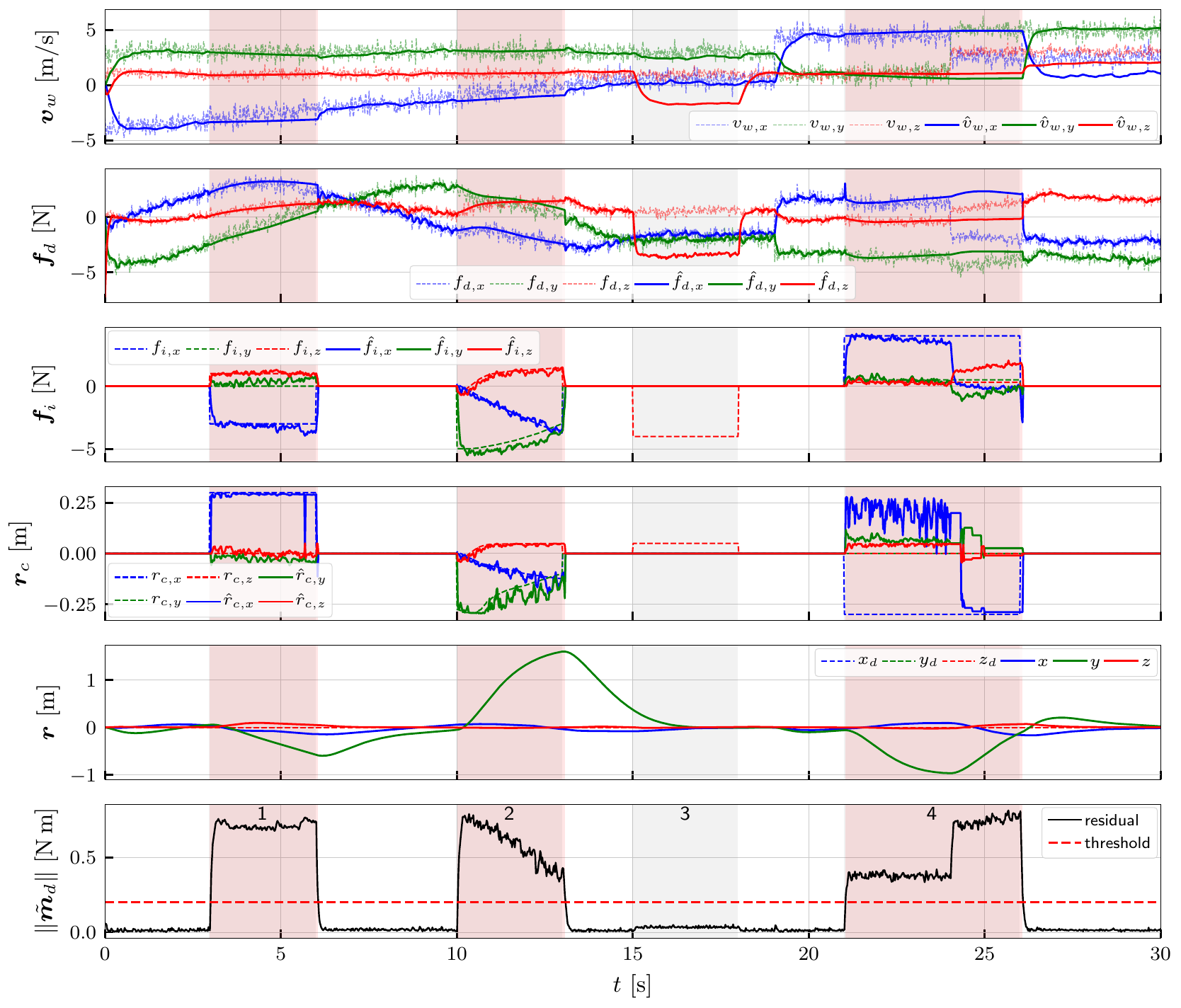}
    \caption{
        Simulation results for discrimination between aerodynamic and contact forces under wind influence,
        using exactly the residual based estimation scheme from \fig{fig:residual_wind_estimation}.
        The aerial robot is yawing in the time-varying wind field, while running the compensated impedance controller.
        From top to bottom: wind speed $\vc{v}_w$; robot position $\vc{r}$; aerodynamic force $\vc{f}_{\!d}$;
        contact force $\vc{f}_{\!i}$; contact position $\vc{r}_c$; torque residual $\tilde{\vc{m}}_e$.
        Periods where contact was detected are shaded in red.
        }
    \label{fig:force_discrimination_residual}
\end{figure*}
\textbf{Simulation results}.
In order to evaluate the proposed discrimnation scheme, we simulated the hexacopter identified in Section \ref{sec:parameter_identification},
with an aerodynamic wrench model identified from data presented in Section \ref{sec:aero_model_evaluation}.
The robot is controlled using the compensated impedance controller \eqref{eq:position_compensated_impedance_controller}, with a constant position setpoint (hover).
The controller is compensating the estimated aerodynamic wrench, while being compliant w.r.t. the estimated interaction wrench.
To illustrate determination of the contact position,
the convex hull is an oblate superellipsoid convex hull with horizontal major semiaxes of 0.3~m, and the vertical minor semiaxis of 0.05~m.
If \eqref{eq:contact_position} has no solution, i.e. does not intersect the convex hull, we keep the previous estimated values.
We then virtually inject a contact force at a position on the convex hull, while the robot motion is unconstrained.
To illustrate how the scheme behaves in time-varying conditions, the simulated conditions are as follows.
Until $t=15$~s, the vehicle is yawing at a 9\,${}^{\circ}$/s, and the wind speed is time-varying.
After $t=15$~s, the wind speed and yaw reference remain constant.
The simulation has 4 distinct contact phases to illustrate the behavior and failure cases of the discrimination scheme:
\begin{enumerate}
    \item Time-varying wind speed, constant contact force and position in the body frame for 3~s $<t<$ 6~s.
    \item Time-varying wind speed, constant contact force in the inertial frame, time-varying contact position for 10~s $<t<$ 13~s.
    \item Constant wind speed, pure contact force (no torque) for 15~s $<t<$ 18~s.
    \item Wind speed step change at $t=24$~s during contact during 21~s $<t<$ 26~s.
\end{enumerate}
Simulation results are shown in \fig{fig:force_discrimination_residual}.
Throughout the simulation, when no interaction is detected, the vehicle converges to the setpoint,
as the complete external wrench is compensated as a disturbance.
The wind velocity also converges to the correct value.
During phase (1), wind estimation is essentially paused, and the error of the time-varying aerodynamic force is interpreted as the interaction force.
The vehicle is compliant to the estimated interaction force.
The raw contact position is estimated correctly.
During phase (2), the time-varying interaction force is correctly estimated because
the wind speed is estimated in the inertial frame, and transformed into the body frame to obtain the aerodynamic wrench.
The compensated impedance controller is making the vehicle compliant to the estimated interaction wrench.
This makes it possible to physically move the vehicle in the inertial
frame while it is yawing under time-varying wind influence, see the robot position $\vc{r}$.
The time-varying contact position is also correctly estimated in this case. There is additional noise due to the small error in the interaction force.
In phase (3), no contact is detected because the purely vertical force does not exert a torque, as can also be seen in \fig{fig:residual_observability}. This is wrongly interpreted as a change in wind.
Lastly, in phase (4), the wind speed exhibits a step change.
The change from the inital aerodynamic force gets falsely interpreted as interaction, hence the scheme fails.
Here we can also see the ambiguity in the contact position determination -- the contact position is estimated on the wrong side of the convex hull (wrong sign of the $x-$axis) after the wind change.
Notice also that the estimated interaction wrench does not intersect with the convex hull most of the time, since $\hat{\vc{r}}_c$ is held constant.

In summary, the force discrimination scheme presented in \fig{fig:residual_wind_estimation} works well
for scenarios where
\begin{itemize}
    \item the wind speed is constant or slowly varying, and
    \item the interaction force exhibits a torque sufficiently different from the aerodynamic torque.
\end{itemize}
Conversely, it will fail if
\begin{itemize}
    \item the interaction force does not generate a torque residual, and/or
    \item the wind speed changes significantly during the interaction.
\end{itemize}
Note that this discrimination scheme does not assume a model of the interaction force.
The contact position determination in the simulation result is obtained under the assumption of a point contact using \eqref{eq:contact_position}.

\subsection{Interaction force at a known contact position}
\label{sec:interaction_force_at_known_location}
In the following, we show that given a known contact position, the interaction force can be computed
by using the aerodynamic torque model, under the assumption that the interaction can be reduced to a
point contact on the robot's convex hull.
This assumption applies to applications such as slung load transportation or contact inspection.
The location of the slung load may e.g. be known by mechanical design.
Alternatively, the contact position may be intialized on collision with the inspected surface, when the signal-to-noise
ratio allows for frequency-based discrimination.

When the contact position $\vc{r}_c$ is known, we may write the external torque
${\vc{m}_e = \vc{m}_d + \vc{m}_i}$ as
\begin{equation}
    \vc{m}_e = \vc{m}_{d} \bigl( \vc{f}_{\!e} - \vc{f}_{\!i} \bigr) + \Ss{\vc{r}_c} \vc{f}_i,
\end{equation}
where $\vc{m}_{d}(\bullet)$ is the nonlinear aerodynamic torque model.
In the nonlinear case, we must solve the nonlinear equation
\begin{equation}
    \vc{F}_{\!i} = \vc{m}_{d} \bigl( \vc{f}_{\!e} - \vc{f}_{\!i} \bigr) + \Ss{\vc{r}_c} \vc{f}_{\!i} - \vc{m}_e = \vc{0},
    \label{eq:interaction_force_known_position_error}
\end{equation}
which may be done using e.g. Levenberg-Marquardt. The Jacobian of \eqref{eq:interaction_force_known_position_error} is
\begin{equation}
    \mat{J}_{!i} = \frac{\partial \vc{F}_i}{\partial \vc{f}_{\!i}}
    = \Ss{\vc{r}_c} - \frac{\partial \vc{m}_{d} \bigl(\vc{f}_{\!d} \bigl)}{\partial \vc{f}_{\!d}}
    = \Ss{\vc{r}_c} - \mat{J}_d,
    \label{eq:interaction_force_known_position_jacobian}
\end{equation}
where ${\mat{J}_d = \tfrac{\partial \vc{m}_{d} (\vc{f}_{\!d} )}{\partial \vc{f}_{\!d}}}$ is the Jacobian of the aerodynamic torque model.
In the special case of a linear model ${\vc{m}_d\bigl(\vc{f}_{\!d}\bigr) = \mat{D} \vc{f}_{\!d}}$, the solution may then be directly found as
\begin{equation}
    \hat{\!\vc{f}}_{\!i} = \Bigl(\Ss{\vc{r}_c} - \mat{D}\Bigr)^{-1} \Bigl( \vc{m}_e - \mat{D} \vc{f}_e \Bigr)
    = \mat{J}_{\!i}^{-1} \tilde{\vc{m}}_d
    .
    \label{eq:interaction_force_known_position_linear}
\end{equation}
The observability of the interaction force will thus depend on the contact position and the aerodynamic model,
as $\mat{J}_{\!i}$ must be nonsingular.
Note also that \eqref{eq:interaction_force_known_position_error} allows the solution for nonlinear torque models.

\begin{figure}
    \centering
    \subfigure[Linear model]{\includegraphics[width=0.47\columnwidth]{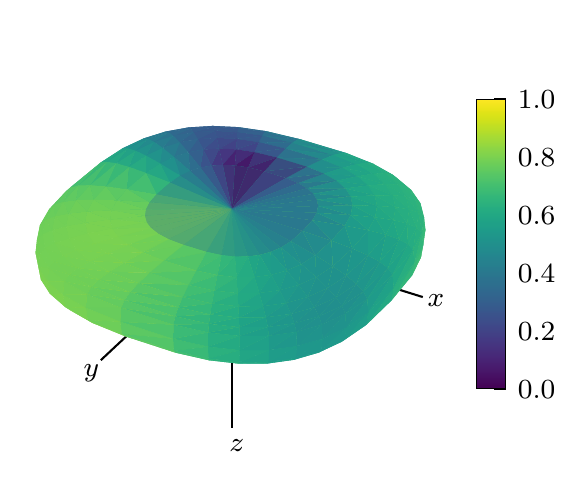} \label{fig:contact_observability_linear}}
    \subfigure[Perceptron model]{\includegraphics[width=0.47\columnwidth]{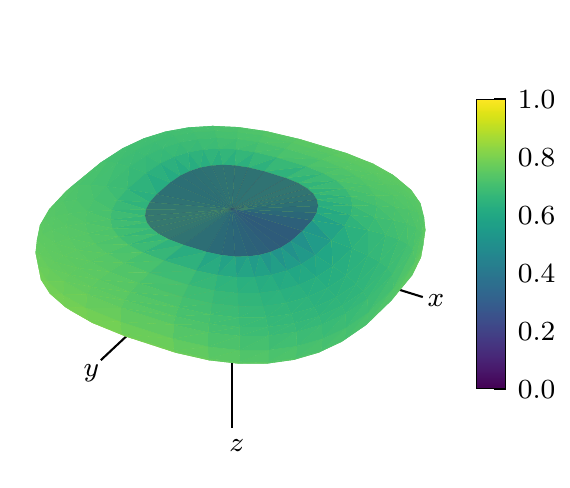} \label{fig:contact_observability_perceptron}}
    \caption{Observability of contact forces on a superellipsoidal convex hull, with different aerodynamic torque models.
        The scale is normalized to the smallest condition number being 1.0, and higher condition numbers going down to zero.
        Darker colors therefore indicate a higher condition number of the Jacobian at the contact position for this particular model.
        The solution will be more susceptible to noise in the darker shaded areas.
        For the perceptron model we show the initial Jacobian at $\hat{\vc{f}}_{\!d}=\vc{0}$.}
    \label{fig:contact_observability}
\end{figure}
\fig{fig:contact_observability} illustrates the interaction force observability on an example superellipsoidal
hull and a linear aerodynamic torque identified in Section \ref{sec:aerodynamic_torque_model}.
Here, we defined observability as the condition number of $\mat{J}_{\!i}$.
The color is normalized to the smallest condition number being brightest, and largest condition number (lowest observability) being darkest.
Areas with a lower observability will be more susceptible to modeling errors and noise in the measurements.
The solution \eqref{eq:interaction_force_known_position_linear} can also guide the contact point selection for best observability,
Note also that the contact point does not have to be on the convex hull, but can also be the end of a tool attached to the flying robot.
The model $\hat{\vc{m}}_d(\vc{f}_{\!d})$ contains implicitly a (possibly varying) center of pressure, in other words the lever $\vc{r}_{\!d}$ of aerodynamic force $\vc{f}_{\!d}$.
For our linear model, this is assumed to be constant, thus $\mat{D}$ is constant.
This explains the strongly asymmetric behavior in \fig{fig:contact_observability_linear} -- forces acting near to the constant center of pressure will
be less observable due to the shorter lever to $\vc{r}_d$.
Notice that the behavior tends towards more symmetry for the richer nonlinear model in \fig{fig:contact_observability_perceptron}.

\begin{figure*}
    \centering
    \subfigure[Linear aerodynamic torque model with $\alpha_1=10^{-5}$, using the direct solution \eqref{eq:interaction_force_known_position_linear}.]%
        {\includegraphics{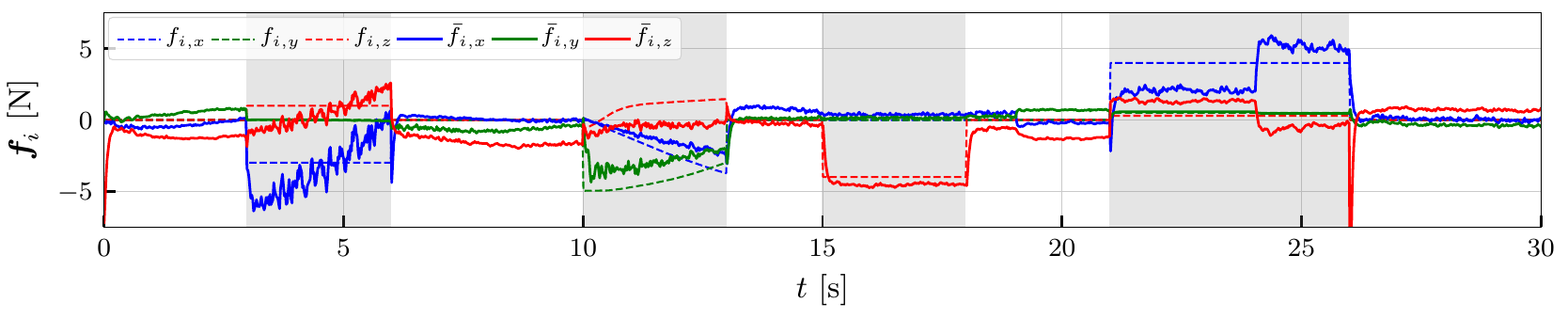}}
    \subfigure[Perceptron aerodynamic torque model (ground truth), using nonlinear optimization of \eqref{eq:interaction_force_known_position_error},
               initialized with $\hat{\vc{f}}_{\!i} = \vc{f}_{\!e}$.]%
        {\includegraphics{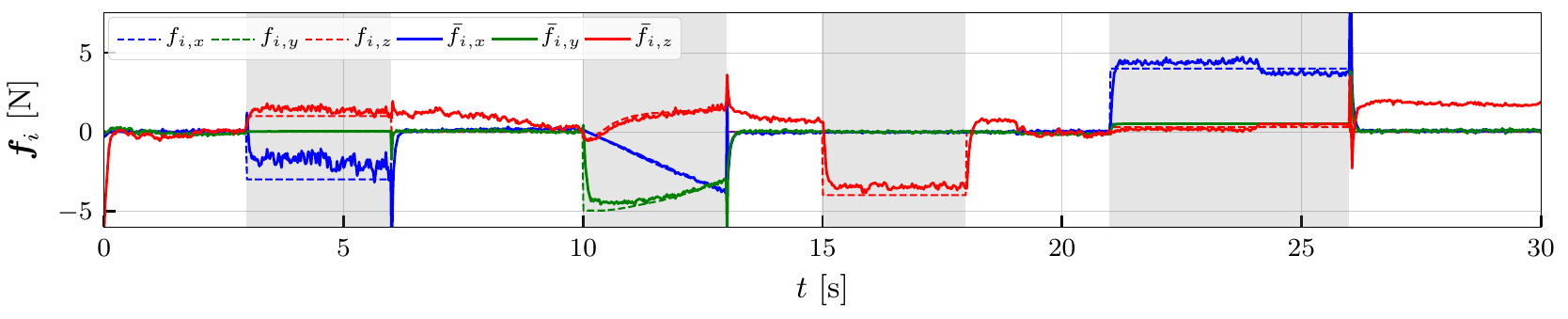}}
    \caption{Force estimation at the exactly known contact position for the simulation in \fig{fig:force_discrimination_residual},
            using different aerodynamic torque models.
            We used a perceptron aerodynamic torque model for simulation.
            The error in the estimated interaction force for the linear model is mainly caused by the aerodynamic torque modeling errormainly ,
            as the true aerodynamic wrench is modeled as a nonlinear perceptron model.
            Even with perfect model knowledge (b), the force is not reconstructed perfectly.
            The force spikes are caused by transients in the external wrench estimate.}
    \label{fig:discrimination_exact_force}
\end{figure*}
\fig{fig:discrimination_exact_force} shows the interaction force computed at the exact (simulated) contact position using
a linear model and direct computation \eqref{eq:interaction_force_known_position_linear}. The aerodynamic wrench in the simulation
was obtained using a nonlinear model, and therefore using a linear model results in an estimation error.
Furthermore, a delay in estimating the external wrench causes some transient errors, for example at $t=0$~s, and $t=26$~s.
Note that the pure force interaction in contact phase 3 is correctly identified because we are using the exact contact position.
However, modeling errors are reflected in a state-dependent error of the estimated interaction force.
The result shows that even with perfect knowledge of the contact position, under modeling errors, the method presented
here is not sufficient to accurately obtain the interaction force.

\subsection{Force discrimination particle filter}
\label{sec:discrimination_particle_filter}

As shown in the previous subsection, even exactly knowing the contact force leads to inaccurate estimation
of the interaction force when modeling errors are present.
In this section we propose a framework that fuses interaction force and wind speed
models into a unified force discrimination scheme.
It is based on directly estimating the contact position on the robot's convex hull by using a particle filter.
By estimating the contact position that best matches the observed external wrench, we essentially transfer
the modeling error from the interaction wrench to the estimated contact position. The latter is by definition bounded
through the convex hull, which results in a more robust overall scheme.
The scheme is based on a Sequential Importance Resampling (SIR) particle filter \citep{Doucet2009}.
Each particle ${\vc{x}_i = [\vc{r}_{c,i}^T\; \vc{v}_{w,i}]^T}$ contains a contact position $\vc{r}_{c,i}$ on the convex hull $C\!H$,
as well as a wind speed $\vc{v}_{w,i}$.
We then use \eqref{eq:interaction_force_known_position_jacobian} to obtain the interaction wrench $\vc{\tau}_{\!i,i}$ at the contact point,
and an aerodynamic wrench model $\vc{\tau}_{\!d}\bigl(\vc{v}_r\bigr)$ at the relative airspeed computed from the particle's wind speed
to obtain the particle aerodynamic wrench $\hat{\vc{\tau}}_{\!d,i}$.
In summary,
\begin{align}
	\hat{\vc{\tau}}_{\!i,i} &= \begin{bmatrix}
        \mat{J}_{\!i}^{-1} \tilde{\vc{m}}_d \\
        \vc{r}_{c,i} \times \bigl( \mat{J}_{\!i}^{-1} \tilde{\vc{m}}_d\bigr)
	\end{bmatrix},
	\\
	\hat{\vc{\tau}}_{\!d,i} &= \vc{\tau}_{\!d}\bigl(\mat{R}^T (\vc{v} - \vc{v}_{w,i} ) \bigr).
\end{align}
Based on ${\hat{\vc{\tau}}_{\!e,i} = \hat{\vc{\tau}}_{\!d,i} + \hat{\vc{\tau}}_{\!i,i}}$, we define
the external wrench error $\tilde{\vc{\tau}}_{\!e,i}$ of particle $i$ as
\begin{equation}
    \tilde{\vc{\tau}}_{\!e,i} = \vc{\tau}_{\!e} - \hat{\vc{\tau}}_{\!i,i} - \hat{\vc{\tau}}_{\!d},
\end{equation}
which is the error between the estimated external wrench and the wrench predicted by particle $i$.
We adopt a Gaussian distribution of the particles around zero of the external wrench error.
The probability of particle $i$ is then
\begin{equation}
	p\left(\hat{\vc{\tau}}_{\!e,i} | \vc{r}_{c,i} \right) =
	\exp \left( -\frac{1}{\sigma^2} \tilde{\vc{\tau}}_{\!e,i}^T \tilde{\vc{\tau}}_{\!e,i} \right),
\end{equation}
where $\sigma$ is a shape parameter.
The algorithm is described in detail in Algorithms \ref{alg:discrimination} and \ref{alg:auxiliary}.
We run the filter when contact is detected using the torque residual.
The contact positions $\vc{r}_{c,i}$ are intialized uniformly on the convex hull, and assigned to the currently estimated wind speed.
The process model for both, the contact position and wind speed, is constant with per-axis Gaussian noise.
At each iteration, we project the contact position onto the convex hull.

\newcommand{\MyKwFont}{\sffamily\small}
\newcommand{\MyCommentFont}{\sffamily\small\it}
\SetFuncSty{MyKwFont}
\SetCommentSty{MyCommentFont}
\begin{algorithm}[t]
    \caption{Contact particle filter for simultaneous estimation of contact and aerodynamic wrenches.}
    \label{alg:discrimination}
    \small
    \newlength\mylen
    \newcommand\MyInput[1]{%
        \settowidth\mylen{\Parameters{}}%
        \setlength\hangindent{\mylen}%
        \hspace*{\mylen}#1\\}
    \SetKwInOut{Input}{Input}
    \SetKwInOut{Output}{Output}
    \SetKwInOut{KwData}{Variables}
    \SetKwInOut{Models}{Models}
    \SetKwInOut{Parameters}{Parameters}
    \SetKwFunction{Initialize}{Initialize}
    \SetKwFunction{rand}{UniformRandomNumber}
    \SetKwFunction{randn}{GaussianRandom}
    \SetKwFunction{UniformSample}{UniformSampleConvexHull}
    \SetKwFunction{RandomSample}{RandomSampleConvexHull}
    \SetKwFunction{Project}{ProjectToConvexHull}
    \SetKwFunction{ProcessModel}{ProcessModel}
    \SetKwFunction{InteractionAtPoint}{InteractionWrench}
    \SetKwFunction{Resample}{Resample}
    \SetKwFunction{cumsum}{cumsum}
    %
    \Input{$\vc{\tau}_{\!e}$: External wrench}
    \MyInput{$\vc{v}$: Translational velocity}
    \MyInput{$\mat{R}$: Orientation}
    %
    \BlankLine
    \Output{$\hat{\vc{\tau}}_{\!i}$: Estimated interaction wrench}
    \MyInput{$\hat{\vc{\tau}}_{\!d}$: Estimated a erodynamic wrench}
    \MyInput{$\hat{\vc{r}}_c$: Estimated contact position}
    \MyInput{$\hat{\vc{v}}_w$: Estimated wind velocity}
    %
    \BlankLine
    \Models{$C\!H$: Convex hull}
    \MyInput{$\vc{\tau}_{\!d}(\vc{v}_r)$: Aerodynamic wrench model}
    \MyInput{$\vc{m}_{d}(\vc{f}_{\!d})$: Aerodynamic torque model}
    %
    \BlankLine
    \KwData{$\vc{x} \in \Realm{6}{N_p}$: Particles}
    \MyInput{$\vc{w} \in \Realv{N_p}$: Particle weights}
    %
    \BlankLine
    \Parameters{$N_p$: Number of particles}
    \MyInput{$N_r$: Resampling threshold}
    \MyInput{$\mat{Q}_r$: Contact position noise}
    \MyInput{$\mat{Q}_w$: Wind speed noise}
    \MyInput{$\rho_{\mathrm{rand}}$: Particle randomization ratio}
    \BlankLine
    ForceDiscrimination:\\
    \Initialize()\\
    \Repeat{$\| \vc{m}_d\bigl({\vc{f}}_{\!e}\bigr) - \vc{m}_e \| \leq \delta$}{
        $k \leftarrow k+1$\\
        \ProcessModel() \\
        \For{$i \leftarrow 1$ \KwTo $N_p$}{
            $\vc{v}_{r,i|k} \leftarrow \mat{R}^T\bigl(\vc{v} - \vc{v}_{w,i|k}\bigr)$ \\
            $\vc{\tau}_{\!d,i|k} \leftarrow \vc{\tau}_{\!d}(\vc{v}_{r,i|k})$\\
            $\vc{\tau}_{\!i,i | k} \leftarrow$ \InteractionAtPoint$\bigl(\vc{r}_{c,i|k}$, $\vc{\tau}_{\!e}\bigr)$\\
            $\tilde{\vc{\tau}}_{i|k} \leftarrow \vc{\tau}_{\!e} - \vc{\tau}_{\!i,i|k} - \vc{\tau}_{\!d,i|k}$\\
            $p\bigl(\vc{x}_{i|k} | \vc{x}_{i|k-1}, \vc{\tau}_e\bigr) = -\exp \bigl( -\tfrac{1}{2 \sigma^2}\tilde{\vc{\tau}}_{i|k}^T \tilde{\vc{\tau}}_{i|k} \bigr)$\\
            $w_{i|k} \leftarrow w_{i|k-1} \ p\bigl(\vc{x}_{i|k} | \vc{x}_{i|k-1}, \vc{\tau}_e\bigr)$
        }
        $\bar{\vc{\tau}}_{\!i|k} = \sum_{i=1}^{N_p} w_{i|k} {\vc{\tau}}_{\!i,i|k}$ \\
        $\bar{\vc{\tau}}_{\!d|k} = \sum_{i=1}^{N_p} w_{i|k} {\vc{\tau}}_{\!d,i|k}$ \\
        $\bar{\vc{r}}_{\!c|k} = \sum_{i=1}^{N_p} w_{i|k} {\vc{r}}_{\!c,i|k}$ \\
        $\hat{\vc{v}}_{\!w|k} = \sum_{i=1}^{N_p} w_{i|k} {\vc{v}}_{\!w,i|k}$ \\
        \Resample()
    }
\end{algorithm}
\begin{algorithm}[t]
    \caption{Auxiliary functions of the particle filter.}
    \label{alg:auxiliary}
    \small
    Initialize:\\
    $k \leftarrow 0$\\
    \For{$i \leftarrow 1$ \KwTo $N_p$}{
        $\vc{r}_{c,i | 0} \leftarrow$ \UniformSample($C\!H$)\\
        $\vc{v}_{w,i | 0} \leftarrow \vc{v}_{w|0}$\\
        $\vc{x}_{i|0} \leftarrow \bigl[ \vc{r}_{c,i | 0}^T \; \vc{v}_{w,i | 0}^T \bigr]^T$\\
        $w_i = \tfrac{1}{N_p}$
    }
    \BlankLine
    ProcessModel:\\
    \For{$i \leftarrow 1$ \KwTo $N_p$}{
        \uIf{\rand\textup{()} $< 1 - \rho_{\mathrm{rand}}$}{
            \emph{Contact position noise model}\\
            $\bar{\vc{r}}_{c,i|k-1} \leftarrow \vc{r}_{c,i|k-1} + \mat{Q}_r \cdot$ \randn()
        }\Else{
            \emph{Randomize contact position}\\
            $\bar{\vc{r}}_{c,i | k-1} \leftarrow$ \RandomSample($C\!H$)\\
        }
        $\vc{r}_{c,i | k} \leftarrow$ \Project($\bar{\vc{r}}_{c,i|k-1}$, $C\!H$)\\
        $\vc{v}_{w,i | k} \leftarrow \vc{v}_{w|k-1} + \mat{Q}_w \cdot $ \randn()
    }
    \BlankLine
    Resample:\\
    $N_{\mathrm{eff}} = 1/ \bigl(\sum_i w_i^2\bigr)$\ \emph{Number of effective samples}\\
    \If{$N_{\mathrm{eff}} < N_r$}{
        $\vc{\gamma} = $ \cumsum($\vc{w}$)\\
        \For{$i \leftarrow 1$ \KwTo $N_p$}{
            \tcc{Uniformly draw new sample index $m$ from the cumulative distribution in $\vc{\gamma}$}
            $\rho =$ \rand()\\
            $m = $ first $j$ for $\gamma_j < \rho$, $j \in 1 \ldots N_p$\\
            $\vc{x}_{i|k} = \vc{x}_{m|k-1}$\\
            $w_i = \tfrac{1}{N_p}$
        }
    }
\end{algorithm}

\begin{figure}
    \centering
    \subfigure[Iteration 1]{\includegraphics[width=0.4\columnwidth]{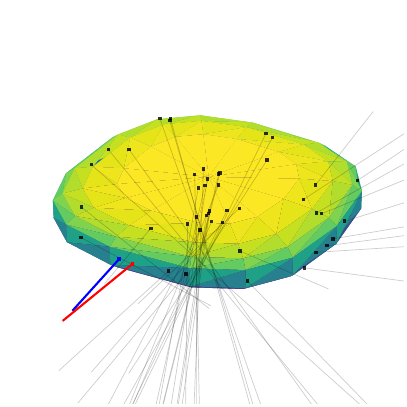}}
    \subfigure[Iteration 2]{\includegraphics[width=0.4\columnwidth]{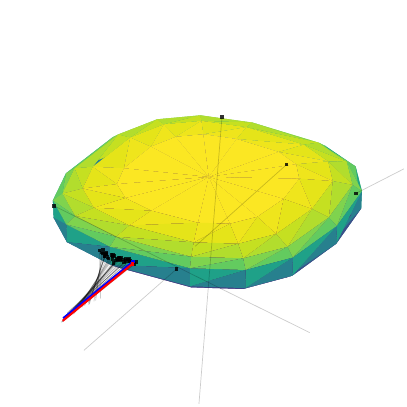}}
    \subfigure[Iteration 3]{\includegraphics[width=0.4\columnwidth]{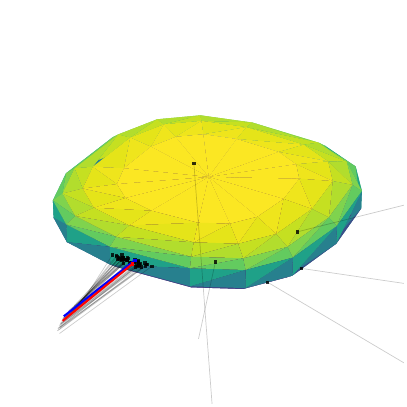}}
    \subfigure[Iteration 4]{\includegraphics[width=0.4\columnwidth]{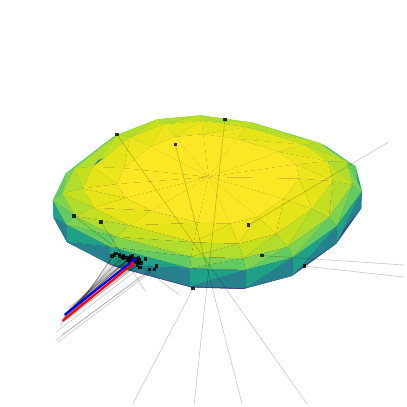}}
    \caption[Particle filter convergence with a linear model, while the aerodynamic wrench is obtained with a nonlinear model.]
    {Particle filter convergence with a linear model, while the aerodynamic wrench is obtained with a nonlinear model.
    Black points represent candidate contact positions on the convex hull (particles).
    The estimated interaction force at that position is represented as a line.
    The {\color{red}red} line depicts the ground truth interaction force, while the {\color{blue}blue} line
    represents the output of the particle filter (contact position and force).}
    \label{fig:particle_filter_convergence_linear}
\end{figure}

\begin{figure}
    \centering
    \subfigure[Iteration 1]{\includegraphics[width=0.4\columnwidth]{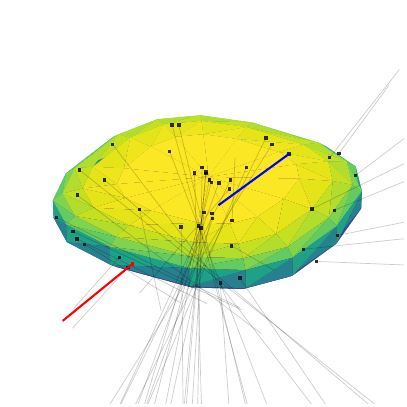}}
    \subfigure[Iteration 2]{\includegraphics[width=0.4\columnwidth]{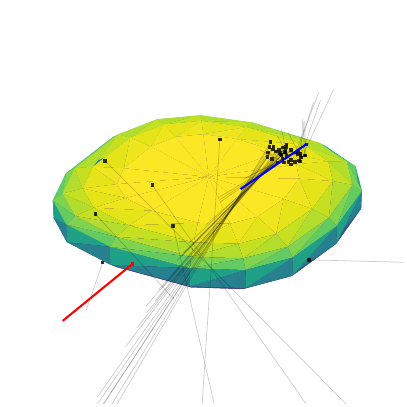}}
    \subfigure[Iteration 3]{\includegraphics[width=0.4\columnwidth]{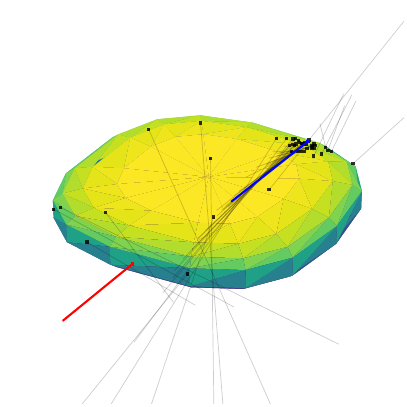}}
    \subfigure[Iteration 4]{\includegraphics[width=0.4\columnwidth]{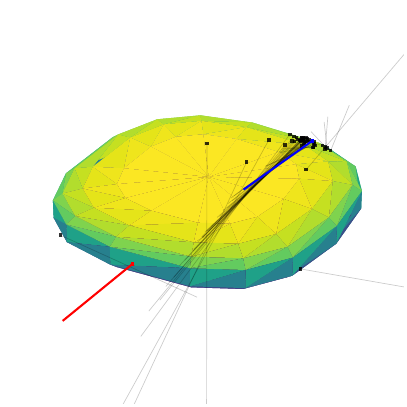}}
    \caption[Example when particles converge to the wrong side of the convex hull, estimating the wrong contact position.]
        {Example when particles converge to the wrong side of the convex hull, estimating the wrong
        contact position.
        This is due to the ambiguity of the contact determination problem: the same torque can be achieved
        by the force on both sides of the convex hull.
        The force direction and magnitude are still estimated correctly.
        Note the  high variability of the interaction force around the sample mean.
        }
    \label{fig:particle_filter_minimum_linear}
\end{figure}

\textbf{Convergence}.
\fig{fig:particle_filter_convergence_linear} depicts four iterations of the particle filter.
The aerodynamic force is obtained from a nonlinear perceptron model, while the particle filter
uses a linear model, which results in a modeling error.
Already after the first iteration, the result is near the actual contact position and force.
After resampling, particles are distributed around the actual contact position, as expected.
Here, we randomize a portion of the particles after each iteration as an "exploration" step.
Iterating further, as expected, particles are still concentrated around the contact position,
and variance of the contact force direction and magnitude is low.
This shows that the particle filter can converge to the actual contact position and force even
with modeling errors.
Conversely, \fig{fig:particle_filter_minimum_linear} shows a degenerate case where the
filter can fall into a local minimum. As discussed in Section \ref{sec:contact_position},
the same wrench can be generated by a force on opposite sides of the convex hull,
see \fig{fig:particle_filter_minimum_linear}.
Notice the large variance of the contact force direction and magnitude.


\begin{figure*}
    \centering
    \includegraphics{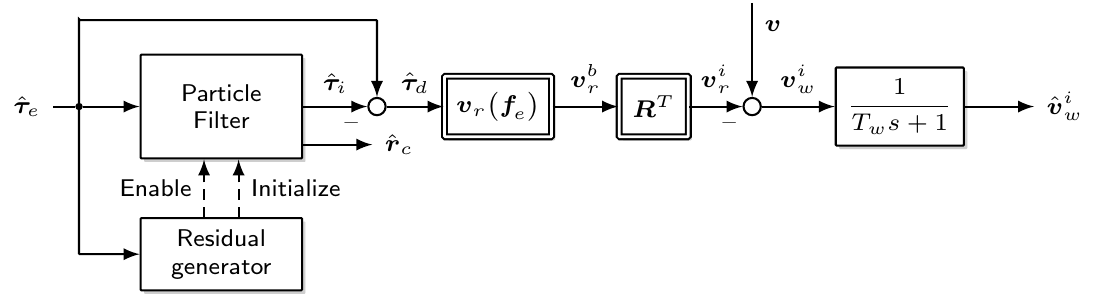}
    \caption[Input discrimination and wind estimation scheme based on the particle filter.]
       {Input discrimination and wind estimation scheme based on the particle filter. We do not adapt the time constant of the wind speed filter.
        Instead, the particle filter outputs the estimated interaction wrench $\hat{\vc{\tau}}_{\!i}$ only when contact is detected by the residual generator.
        Then, the drag wrench $\hat{\vc{\tau}}_{\!d}$ is subsequently used to obtain the relative airspeed.}
    \label{fig:particle_filter_wind_estimation}
\end{figure*}

\textbf{Discrimination scheme}.
The particle filter is one component in the discrimination scheme depicted in \fig{fig:particle_filter_wind_estimation}
In this scheme, we do not adapt the time constant of the wind estimator. Instead, we estimate the interaction wrench $\hat{\vc{\tau}}_{\!i}$ directly.
The estimated aerodynamic wrench ${\hat{\vc{\tau}}_{\!d} = \vc{\tau}_e - \hat{\vc{\tau}}_{\!i}}$ is then used directly to obtain an
estimated relative airspeed. The raw wind speed is filtered in the same manner as in the previous discrimination scheme.
Note, however, that we do not need to slow down estimation of the wind speed.

\begin{figure*}
    \centering
    \includegraphics{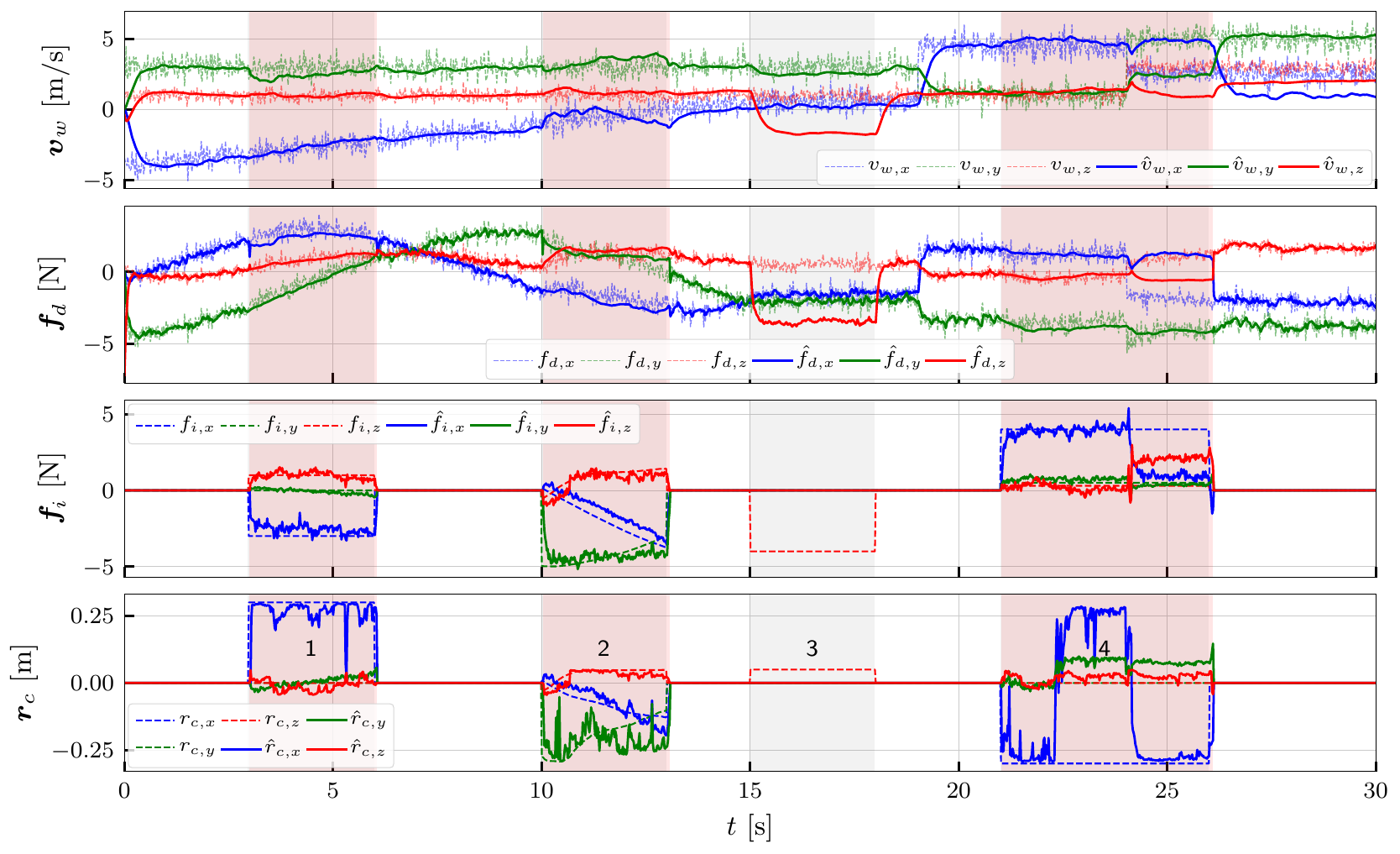}
    \caption{
        Discrimination between aerodynamic and contact forces under wind influence,
        using the particle filter based approach.
        The aerial robot is yawing in the time-varying wind field, while running the compensated impedance controller.
        From top to bottom: wind speed $\vc{v}_w$; robot position $\vc{r}$; aerodynamic force $\vc{f}_{\!d}$;
        contact force $\vc{f}_{\!i}$; contact position $\vc{r}_c$; torque residual $\tilde{\vc{m}}_e$.
        Periods where contact was detected are shaded in red.
        }
    \label{fig:particle_filter_discrimination}
\end{figure*}

\textbf{Results}. To test particle filter based discrimination, the same simulation scenario as for the residual based discrimination was used,
with $N_p=45$ particles, contact position noise ${\mat{Q}_r=\mathrm{diag}\{0.025, 0.025, 0.005 \}}$ m, and wind speed noise ${\mat{Q}_w = 0.001 \eye{3}}$ m/s.
The results are shown in \fig{fig:particle_filter_discrimination}.
The position, aerodynamic force, and residual plots are omitted as they are comparable.
The notable difference to \fig{fig:force_discrimination_residual} is that the contact position fluctuates more. This behavior can be tuned through $\mat{Q}_r$.
In phase (4), the contact position estimation is stable even after the wind speed changes abruptly.
However, the filter converges to the wrong side of the convex hull during the contact (see $\hat{r}_{c,x}$), but switches back to the correct side after a short period.
This is due to the ambiguity of the contact determination problem discussed above.
In this simulation, wind speed estimation is not slowed down during contact.
Instead, the particle filter outputs the estimated interaction wrench $\hat{\vc{\tau}}_{\!i}$, and the resulting $\hat{\vc{\tau}}_{\!d}$ is used to estimate wind speed.
Overall, the results for this simulation scenario are not significantly better than the modified model-checking estimation scheme, because it is based on the same principle.
However, this framework may be used to more accurately determine the contact position, given an interaction wrench.


\textbf{Extensions}.
Note that this algorithm can be extended to also include the interaction force $\vc{f}_{\!i}$ in the particle state.
However, that further increases the dimensionality of the problem.
Lastly, wind speed may be omitted from the particle state to simplify the filter to only determine the contact position.

\subsection{Power based discrimination}
\label{sec:power_based_discrimination}
\begin{figure*}
    \centering
    \includegraphics{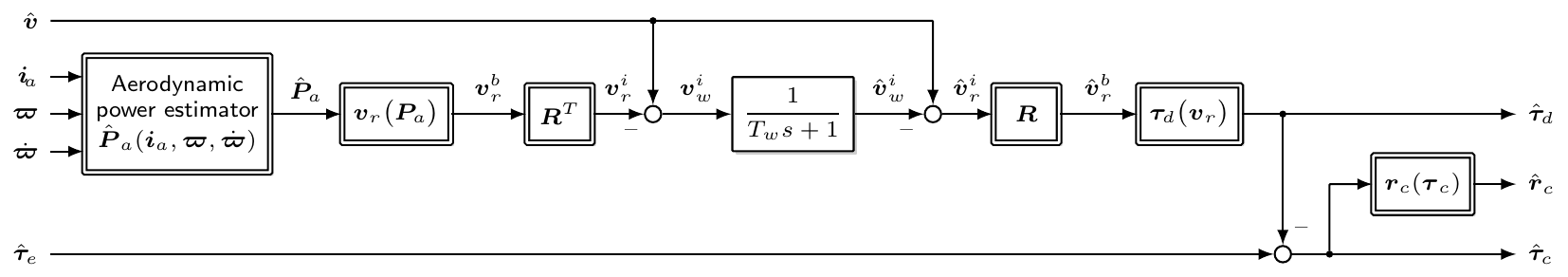}
    \caption{Aerodynamic power based force discrimination scheme. The motor current $i_a$, speed $\varpi$
    and angular acceleration $\dot{\varpi}$ are used to obtain the aerodynamic power $\hat{\vc{P}}_{\!a}$.
    The airspeed is then obtained by a model $\vc{v}_r(\vc{P}_a)$, as learned in Section \ref{sec:aero_model_evaluation}.}
    \label{fig:power_force_discrimination}
\end{figure*}

The previously presented force discrimination schemes rely on the aerodynamic torque residual for contact detection.
In Section \ref{sec:aero_model_evaluation} and Section \ref{sec:wind_from_power} we have shown that the airspeed
may be obtained also by using the aerodynamic power instead of the external wrench. In this section, we propose a
force discrimination scheme, that exploits this fact, see \fig{fig:power_force_discrimination}.

First, an estimate of the aerodynamic power $\hat{\vc{P}}_a$ is obtained from the motor current $\vc{i}_a$, motor speed $\vc{\varpi}$,
and motor acceleration $\dot{\vc{\varpi}}$ as described in Section~\ref{sec:motor_model}.
Second, a raw body-frame airspeed measurements is obtained by employing a model $\vc{v}_r(\vc{P}_a)$, identified in Section~\ref{sec:aero_model_evaluation}.
This is converted into an inertial-frame wind velocity and filtered using an appropriate time constant $T_w$.
The aerodynamic wrench $\hat{\vc{\tau}}_{\!d}$ is then obtained from
the aerodynamic model $\vc{\tau}_{\!d}(\vc{v}_r)$, identified in Section~\ref{sec:aero_model_evaluation}.
The estimated contact wrench and position are computed as in the modified model-checking scheme.
Note that if the airspeed was estimating directly in the body frame, the accuracy would depend on the trajectory, i.e. the
filter time constant would have to change with the movement of the robot, for example during the yawing motion in phases 1 and 2 of the simulation.
Estimating the wind speed in the inertial frame is therefore crucial because of the that is the relevant quantity which is slowly time-varying in the inertial frame.

\begin{figure*}
    \centering
    \includegraphics{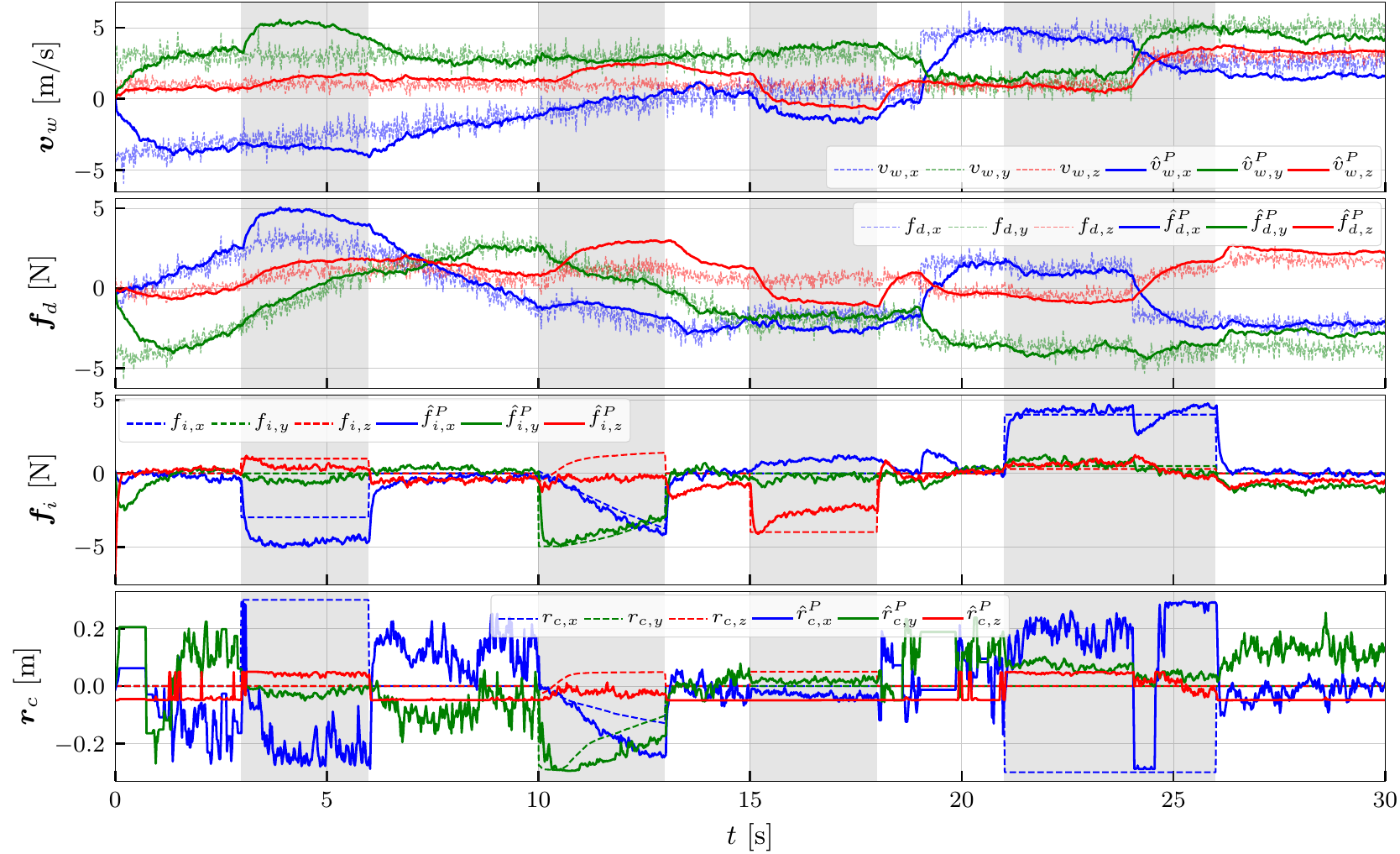}
    \caption{
        Simulation results for power based force discrimination, for the same case as in \fig{fig:force_discrimination_residual}.
        Periods of contact are shaded in gray. Explicit contact detection is not needed in this scheme.
        }
    \label{fig:force_discrimination_power}
\end{figure*}
The scheme is tested for the same simulation scenario as previous schemes, and results are shown in \fig{fig:force_discrimination_power}.
In the simulation, we used the aerodynamic power of coaxial rotor pairs, and added Gaussian noise with a standard deviation of 8~W to the
power measurements.
The wind estimation time constant was $T_w=0.2$~s.
The resulting wind speed and therefore aerodynamic wrench exhibit bias errors in some cases.
Note that, contary to the above discrimination schemes, there is no explicit slowing down of the wind estimation.
There is also no explicit contact detection, as the scheme runs continuously, causing some systematic errors.
For example, during the first contact phase the force is falsely interpreted as $x-$axis airspeed.
This is probably because the underlying model fitted from wind tunnel data has partly also fitted the external force.
The same would happen in the above schemes if wind speed estimation were not slowed down.
The power based algorithm performs better in the other contact phases, despite the same systematic errors.
Notably, scheme successfully detects the interaction force coming from above in the phase 3, which is a failure case of residual based methods.

In summary, we can conclude that the motor power may be used to provide an independent measurement of the aerodynamic wrench,
which can then successfully be applied to discriminate between interaction and aerodynamic wrenches.
The accuracy of the method will depend on measurement noise of the motor current and speed,
the fit from the motor power to propeller aerodynamic power,
the aerodynamic model mapping aerodynamic power to airspeed,
and the aerodynamic model mapping airspeed to the aerodynamic wrench.
Therfore, the difficulty in applying this scheme lies in the effort to obtain these
models accurately, as modeling errors propagate through the estimation chain.

\subsection{Kalman filter for force discrimination}
\label{sec:kalman_discussion}

\begin{figure*}
    \centering
    \includegraphics{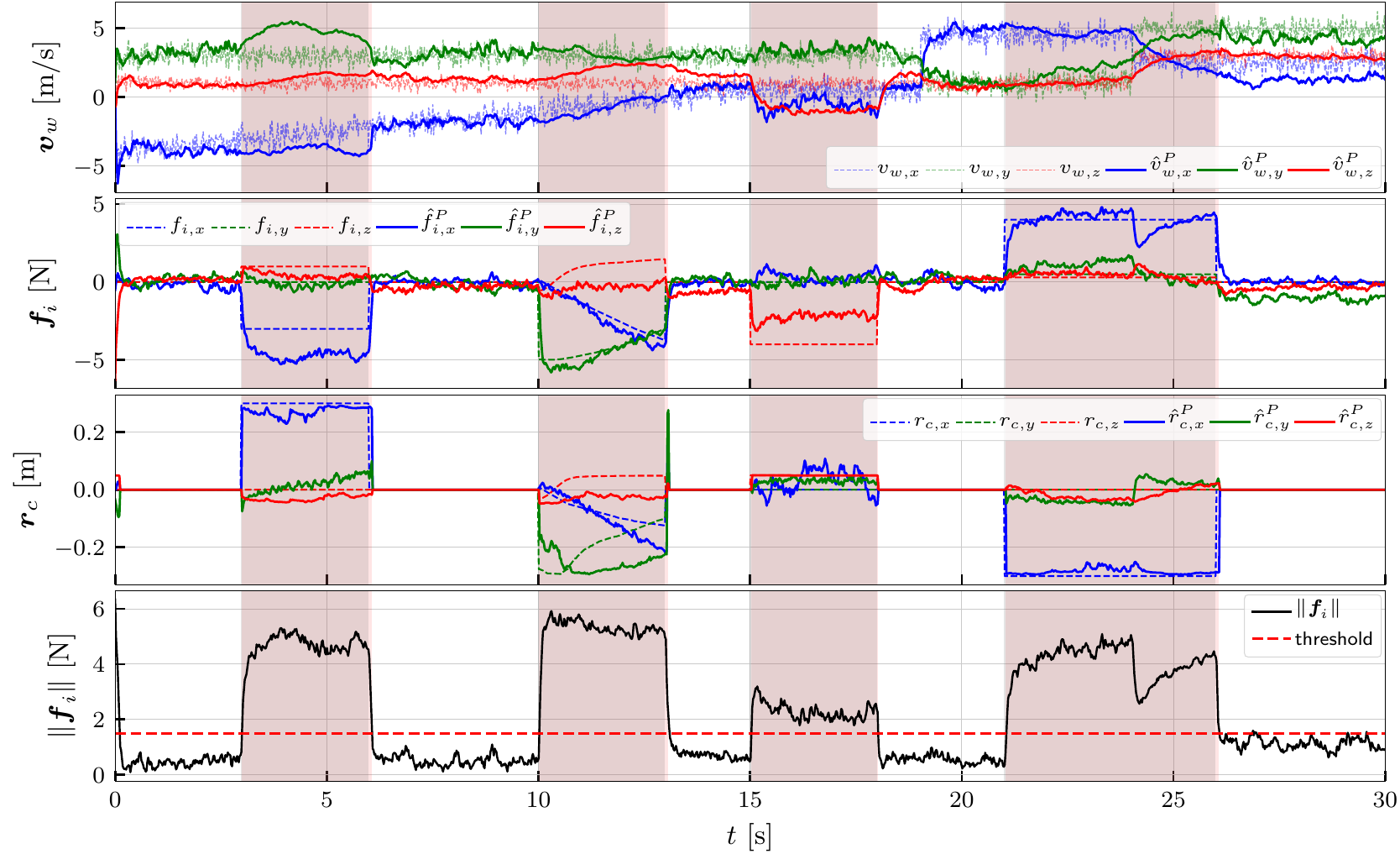}
    \caption{
        Simulation results for Kalman filter based force discrimination, which combines airspeed obtained
        from external forces, and airspeed obtained from power. Red shaded areas indicate periods during which
        the filter has detected contact.
        }
    \label{fig:force_discrimination_kalman}
\end{figure*}

The modified model checking method fails when a pure force is applied to the vehicle and when wind
speed changes during contact.
On the other hand, the power based discrimination can run continuously and can detect pure forces, but is prone to modeling errors (offset).
Neither method performs robustly for estimating the contact position directly from the estimated interaction wrench,
especially if the wind speed changes during contact.
The particle filter performs favorably for this task, as it is designed to estimate the contact position.
Our aim is to combine the strengths of the previously proposed discrimination schemes into a unifying framework.
A Kalman filter is an established tool for such sensor fusion.
In the filter we estimate the wind speed directly, i.e. the state is ${\vc{x} = \vc{v}_{\!w}}$.
The process model is a constant with Gaussian noise ${\mat{Q} = Q_w \eye{3}}$.
The measurements $\vc{z}$ are the instantaneous model-based estimates of the wind speed
\begin{equation}
    \vc{z} = \begin{bmatrix}
        \vc{v} - \mat{R} \vc{v}_{\!r}(\bar{\vc{f}}_{\!d}) \\
        \vc{v} - \mat{R} \vc{v}_{\!r}(\bar{\vc{P}}_{a})
    \end{bmatrix},
    \label{eq:kalman_measurement_vector}
\end{equation}
where $\bar{\vc{f}}_{\!d}$ is to be defined below.
The measurement matrix $\mat{H}$ and measurement covariance $\mat{R}_z$ are simply
\begin{equation}
    \mat{H} = \begin{bmatrix}
        \eye{3} \\ \eye{3}
    \end{bmatrix},
    \quad
    \mat{R}_z = \begin{bmatrix}
        R_f \eye{3} & \nil{3} \\
        \nil{3} & R_p \eye{3}
    \end{bmatrix}.
    \label{eq:kalman_measurement_matrices}
\end{equation}
We omit the implementation details of the Kalman filter, as they can be found in any textbook on the subject.
Because the estimated interaction force from the power based method can estimate a pure interaction force,
we add the additional contact detection signal
\begin{equation}
    C\!D_2 = \begin{cases}
        1 \quad \mathrm{if}\ \| \hat{\vc{f}}_{\!i} \| > \delta_f, \\
        0 \quad \mathrm{otherwise},
    \end{cases}
\end{equation}
where $\delta_f$ will depend on the aerodynamic modeling error and state (e.g. velocity).
The new contact detection signal for the Kalman filter is ${C\!D_K = C\!D_1 \lor C\!D_2}$, i.e.
if either of the contact detection signals are true.
When contact is detected, we initialize the particle filter. During contact, we use \emph{only} the contact
position estimate of the particle filter, obtained by applying the Kalman-filter estimated aerodynamic wrench to evaluate the sample probability density function.
This leads to a more robust contact position estimate than based purely on the torque residual, as it fuses information from multiple sources.

We simulated a discrete-time Kalman filter with a time step of ${T=0.02}$ s, and
used perceptron models for both the aerodynamic power and external force based airspeed estimation.
Similar to the modified model-checking method, we change the filter parameters when the contact detection signal $C\!D_K$ is true.
When not in contact, we use the parameters
\begin{equation*}
    \begin{aligned}
        &Q_w = 3.6 \cdot 10^{-7}, \quad
        &R_f = 1.6 \cdot 10^{-7}, \\
        &R_p = 3.6 \cdot 10^{-5}, \quad
        &\bar{\vc{f}}_{\!d} = \vc{f}_{\!e},
    \end{aligned}
\end{equation*}
and when ${C\!D_1=1}$, we use the parameters
\begin{equation*}
    \begin{aligned}
        &Q_w = 9.0 \cdot 10^{-10},\quad
        &R_f = 1.6 \cdot 10^{-7},\\
        &R_p = 1.0 \cdot 10^{-6},\quad
        &\bar{\vc{f}}_{\!d} = \vc{f}_{\!d}\bigl(\mat{R}^T (\vc{v} - \hat{\vc{x}} )\bigr).
    \end{aligned}
\end{equation*}
In essence, this mimics the behavior of the modified model checking methods where wind estimation is slowed
down when contact is detected. In contact, we decrease the measurement covariance of the power based estimate, to take it into account more strongly.
The particle filter is run with ${N_p=20}$ particles, and contact position noise ${\mat{Q}_r=\mathrm{diag}\{1.25,\,1.25, 0.25\}\cdot10^{-2}}$~m.
The particles are initialized uniformly across the convex hull once contact is detected.
Note that these parameters can be tweaked to fine-tune the overall behavior.
We omitted the optimization-based estimation of the vertical airspeed component, as it did not provide substantially
better performance than the data-driven model. However, this information can also be easily included in the filter if needed.
\fig{fig:force_discrimination_kalman} shows the resulting estimated wind speed, interaction force, and contact position.
Note that the result is largely similar to the power based estimation. However, in the periods without interaction, the offset of the interaction force is lower.
A negative side effect is that the pure force interaction in phase 3 is underestimated in this scheme, due to the filter also trusting the force based measurement.
In phase 4, the filter can maintain the same discrimination performance even if the wind speed rapidly changes.
Finally, because the contact position is estimated by the particle filter, it exhibits more stable behavior than computing it from the estimated interaction wrench directly.
Note that it is still prone to switching the side of convex hull during contact. This effect can be prevented by stronger filtering, however that is out of scope of this paper.
In summary, the Kalman filter is a framework that can easily combine the two developed methods to provide quite robust discrimination between aerodynamic and contact wrenches.

\subsection{Summary}
\label{sec:discrimination_discussion}
\begin{table*}
    \small \sf \centering
    \caption{Overview of pipeline stages for simultaneous collision handling and wind estimation,
        in the context of a Fault Detection, Identification and Isolation pipeline \citep{Haddadin2017}.}
    \label{tbl:discrimination_methods}
    \begin{tabular}{
            >{\raggedright\arraybackslash}p{1.5em}
            >{\raggedright\arraybackslash}p{8em}
            >{\raggedright\arraybackslash}p{18em}
            >{\raggedright\arraybackslash}p{3em}
            >{\raggedright\arraybackslash}p{17em}
            }
        \toprule
        \textbf{Sec.} &
        \textbf{Pipeline stage} &
        \textbf{Algorithm} &
        \textbf{Input} &
        \textbf{Required models} \\
        \midrule
        \ref{sec:collision_detection} &
        {Detection} &
        {$C\!D = H(|\hat{f}_{e,i}|, \omega_f) > f_{c,i}$} &
        {$\fext$} &
        \parbox[t]{17em}{Fault isolation frequency $\omega_f$,\\
                         threshold $f_{c,i}$} \\
        \midrule
        \ref{sec:contact_position} &
        {Isolation} &
        {$\vc{r}_c = \vc{o} + k \vc{d}$} &
        {$\tauext$} &
        \parbox[t]{17em}{Convex hull} \\
        \midrule
        \ref{sec:contact_detection} &
        {Detection} &
        {$C\!D_1 = \| \tilde{\vc{m}}_d \| > \delta $} &
        {$\tauext$} &
        \parbox[t]{17em}{Aerodynamic torque model $\vc{m}_d\bigl(\vc{f}_{\!d} \bigr)$} \\
        \midrule
        \ref{sec:residual_discrimination} &
        \parbox[t]{5em}{Isolation, \\ Identification} &
        {$\hat{\vc{\tau}}_{\!i}$, $\hat{\vc{\tau}}_{\!d}, \hat{\vc{v}}_w, \hat{\vc{r}}_c \leftarrow$~\fig{fig:residual_wind_estimation}} &
        {$\tauext$} &
        \parbox[t]{17em}{
            Aerodynamic torque model $\vc{m}_d(\vc{f}_{\!d})$,
            wind speed propagation model} \\
        \midrule
        \ref{sec:interaction_force_at_known_location} &
        {Identification} &
        {$\vc{f}_{\!i} = \mat{J}_{\!i}^{-1} \tilde{\vc{m}}_d$} &
        {$\tauext, \vc{r}_c$} &
        \parbox[t]{17em}{
            Aerodynamic torque model $\vc{m}_d\bigl(\vc{f}_{\!d} \bigr)$,\\
            interaction position $\vc{r}_c$} \\
        \midrule
        \ref{sec:discrimination_particle_filter} &
        \parbox[t]{5em}{Isolation, \\ Identification} &
        {$\hat{\vc{\tau}}_{\!i}$, $\hat{\vc{\tau}}_{\!d}, \hat{\vc{v}}_w, \hat{\vc{r}}_c \leftarrow$~
            Algorithms \ref{alg:discrimination} and \ref{alg:auxiliary},
            \fig{fig:particle_filter_wind_estimation}} &
        {$\tauext$} &
        \parbox[t]{17em}{
            Aerodynamic torque model $\vc{m}_d\bigl(\vc{f}_{\!d} \bigr)$,\\
            convex hull, contact position model} \\
        \midrule
        \ref{sec:discrimination_particle_filter} &
        \parbox[t]{5em}{Isolation, \\ Identification} &
        {$\hat{\vc{\tau}}_{\!i}$, $\hat{\vc{\tau}}_{\!d}, \hat{\vc{v}}_w, \hat{\vc{r}}_c \leftarrow$~
            Algorithms \ref{alg:discrimination} and \ref{alg:auxiliary}
            with $\hat{\vc{v}}_w$ in particle state
        } &
        {$\tauext$} &
        \parbox[t]{17em}{
            Aerodynamic torque model $\vc{m}_d\bigl(\vc{f}_{\!d} \bigr)$,\\
            aerodynamic wrench model $\vc{\tau}_d\bigl(\vc{v}_{r} \bigr)$,\\
            convex hull, contact position model, wind speed propagation model}  \\
        \midrule
        \ref{sec:power_based_discrimination} &
        {Isolation, Identification} &
        \parbox[t]{18em}{\raggedright
            $\hat{\vc{\tau}}_{\!d} := \vc{\tau}_{\!d}\Bigl(\vc{v}_r\bigl( \hat{\vc{P}}_{\!a} \bigr)\Bigr)$,\\
            $\hat{\vc{\tau}}_{\!i}, \hat{\vc{v}}_w, \hat{\vc{r}}_c \leftarrow$~\fig{fig:power_force_discrimination}
        } &
        {$\tauext, \bar{\vc{P}}_{\!a}$} &
        \parbox[t]{17em}{\raggedright
            Airspeed model $\vc{v}_r\bigl(\vc{P}_{\!a} \bigr)$, convex hull,\\
            aerodynamic power model $\hat{\vc{P}}_{\!a} ( \vc{P}_m )$,
            aerodynamic wrench model $\vc{\tau}_d\bigl(\vc{v}_{r} \bigr)$} \\
        \midrule
        \ref{sec:kalman_discussion} &
        {Identification} &
        {Kalman filter with measurement vector \eqref{eq:kalman_measurement_vector} and matrices \eqref{eq:kalman_measurement_matrices}} &
        {$\tauext, \bar{\vc{P}}_{\!a}$} &
        \parbox[t]{17em}{Aerodynamic torque model $\vc{m}_d\bigl(\vc{f}_{\!d} \bigr)$,\\
                Airspeed model $\vc{v}_r\bigl(\hat{\vc{P}}_{\!a} \bigr)$,\\
                aerodynamic power model $\hat{\vc{P}}_{\!a} ( \vc{P}_m )$,
                aerodynamic wrench model $\vc{\tau}_{\!d}\bigl(\vc{v}_{r} \bigr)$} \\
        \midrule
        \ref{sec:kalman_discussion} &
        {Detection} &
        {$C\!D_2 = \| \hat{\vc{f}}_i \| > \delta_f $} &
        {$\hat{\vc{f}}_{\!i}$} &
        \parbox[t]{17em}{Threshold $\delta_f$} \\
        \bottomrule
    \end{tabular}
\end{table*}

Table~\ref{tbl:discrimination_methods} provides an overview of the novel force discrimination methods developed in this paper,
in the context of a Fault Detection, Identification and Isolation pipeline \citep{Haddadin2017}.
Three different \emph{detection} signals may be used.
The collision detection signal $C\!D$ detects collisions based on frequency.
The contact detection signal $C\!D_1$ detects contacts based on the external torque.
Finally, the contact detection signal $C\!D_2$ detects contacts based on the estimated interaction force.
\emph{Isolation} in this context means obtaining the contact position on the robot's convex hull.
This may be achieved by raycasting as discussed in Section~\ref{sec:contact_position},
or by a particle filter as discussed in Section~\ref{sec:discrimination_particle_filter}.
Finally, the \emph{identification} stage reconstructs the constituent wrenches $\vc{\tau}_{\!i}$ and $\vc{\tau}_{\!d}$.
Two distinct approaches are used for this purpose.
In the modified model-checking scheme, Section~\ref{sec:residual_discrimination}, wind estimation is slowed down when contact is detected.
In the power-based scheme, Section~\ref{sec:power_based_discrimination}, the wind speed is obtained from the motor power, which is a measurement independent of the external wrench.
Finally, we showed that these two schemes may be successfully combined in a Kalman filter framework.

\section{Conclusion}
\label{sec:conclusion}
In this paper, we made significant steps towards simultaneous contact and aerodynamic force estimation for aerial robots without the need for additional onboard sensors.
We achieve this by modeling the relevant aerodynamics models and reasoning about wind velocity.
This paper has several contributions.
First, we provide a systematic procedure for parameter identification of aerial robots.
Second, we analyze the prediction and generalization performance of data-driven aerodynamic models from flight data obtained in a wind tunnel.
Third, we show how the airspeed can be obtained by using motor power, independent of the vehicle's IMU.
This is achieved using machine learning methods.
We also develop an optimization based method for this purpose that provides insights into the underlying physics of the problem.

The second part of the paper deals with discrimination between interaction and aerodynamic wrenches, using models identified in the first part of the paper.
The results are applied in a compensated impedance controller that compensates the wind disturbance, while being compliant to interaction forces.
This allows interaction control under wind influence without the need for additional sensors.
We developed four novel methods to discriminate between the inputs and evaluated them in simulation, using previously identified models.
All of them rely on estimating the wind speed in an inertial frame and applying an aerodynamics model to obtain the aerodynamic wrench.
This is subtracted from the estimated external wrench to obtain the interaction wrench.
The first, modified model-checking scheme, is based on slowing down wind estimation when contact is detected.
Contact detection is based on the residual between the external torque and the expected aerodynamic torque as a function of the external force.
We show the limits of this contact detection methods.
The second method is a particle filter that directly estimates the contact position, under the assumption of a point contact on the robot's convex hull.
The third method uses the estimated aerodynamic power to obtain the airspeed. Therefore, airspeed estimation in this scheme is independent of the vehicle's IMU.
Lastly, we combine the modified model-checking and power-based method in a Kalman filter and use the particle filter to obtain the contact position.
This results in a robust overall scheme for obtaining the interaction wrench and contact position when a flying robot is operating under wind influence,
without the need for additional sensors.

In summary, this paper is a significant step forward toward physical interaction of aerial robots under wind influence.
Next steps include experimental validation of the force discrimination methods in interaction control scenarios.

\begin{acks}
This work was partially funded by the project EuRoC (grant agreement no. CP-IP 608849).
The authors would like to thank H. Wagner, R. Rittgarn, B. Pleitinger, F. Schmidt, T. W\"usthoff, I. Bargen and S. Moser for their support during development of the flying robot platform,
and I. Kossyk and S. Dasgupta for helpful discussions about machine learning.
\end{acks}

\appendix
\section{Appendix}
\subsection{Choosing optimal measurements for wind estimation}
Due to the sensitivity of the optimization problem, a natural question is how to choose measurements to maximize observability of the wind velocity,
i.e. reduce sensitivity to noise.
This problem is closely related to generating exciting trajectories for optimal parameter estimation in robotics, see \cite{Armstrong1989, Swevers1997, Park2006}.
The results of the following analysis may then be applied to path planning, trajectory generation, or as a control signal to achieve active sensing.

According to matrix function literature \citep{Trefethen1997, Higham2008, Golub2012}, the sensitivity to noise of an optimization problem such as \eqref{eq:minim_goal_function} depends on the condition number of its Jacobian.
By minimizing the condition number of the Jacobian ${\mat{J} = \partial \vc{f}(\vc{x}) / \partial\vc{x}}$, we maximize the observability of the problem.
The absolute condition number $\hat{\kappa}(\cdot)$ is defined as
\begin{equation}
    \hat{\kappa} = \| \mat{J}(\vc{x}) \|,
    \label{eq:absolute_condition_number}
\end{equation}
whereas the \emph{relative condition number} $\kappa(\cdot)$ is defined as
\begin{equation}
    \kappa = \frac{\norm{\mat{J}(\vc{x})} }{\norm{\vc{f}(\vc{x})} \norm{\vc{x}}}
    ,
    \label{eq:relative_condition_number}
\end{equation}
where the Frobenius norm of a matrix
\begin{equation}
    \norm{\mat{A}}_F = \sqrt{\mathrm{Tr}\{ \mat{A}^T \mat{A} \}} = \sqrt{\sum_{i=1}^m \sum_{j=1}^n | a_{ij} | ^2 }
\end{equation}
is commonly used.
We are interested in the Jacobian of the previously defined least squares problem.
Recall the Levenberg-Marquardt update step
\begin{equation}
    \delta \vc{x} = \left( \mat{J}^T \mat{J} + \mu \mat{I} \right)^{-1} \mat{J}^T \, \vc{f}(\vc{x}),
\end{equation}
where $\mu$ is the regularization parameter,
$\delta \vc{x}$ is the update step of the optimized variable,
and $\vc{f}(\vc{x})$ is the cost function value evaluated at $\vc{x}$.
Hence, the condition number of the inverted matrix $\mat{J}^T \mat{J}$ is a measure of the problem's sensitivity.
Note also that ${\kappa(\mat{J}^T\mat{J}) = \kappa^2(\mat{J})}$.

Next, we define the optimization problem for choosing propeller orientations that result in the best conditioning of the least squares problem.
First, we define a \emph{sequential optimization problem} where a set of ${K-1}$ measurements is available, and we want to choose the next measurement $K$ that will maximize the observability of the least squares problem.
This approach is useful for online motion planning or control, where only local information is available.
Let ${\vc{\varphi}}$ be a suitable parameterization of the measurement orientation.
The orientation $\vc{\varphi}_K^*$ associated with the \emph{next best measurement} is then obtained by solving the optimization problem
\begin{equation}
    \vc{\varphi}_K^* = \underset{\vc{\varphi} }{\mathrm{arg\,min}} \ \kappa(\mat{J})
    .
    \label{eq:sequential_optimal_measurement}
\end{equation}
The problem \eqref{eq:minim_goal_function} is normalized for easily interpretable physical values.
We may therefore simplify the optimization to minimizing the absolute condition number $\hat{\kappa}$, obtaining
\begin{equation}
    \vc{\varphi}_K^* = \underset{\vc{\varphi}}{\mathrm{arg\,min}} \ \norm{\mat{J}}_F^2
    ,
\end{equation}
where we squared the norm to simplify computation of the derivatives.
Note that this approach is similar to observability analysis, with the Jacobian $\mat{J}$ corresponding to the observation matrix.

This local problem may also be solved online by means of a gradient descent algorithm.
Define ${\vc{\varphi}}$ to be a suitable parameterization of the measurement orientation.
In continuous time, solving
\begin{equation}
    \dot{\vc{\varphi}} = - \gamma \, \frac{\partial \norm{\mat{J}}_F^2 }{\partial \vc{\varphi}}
    ,
    \label{eq:sequential_optimal_potential}
\end{equation}
where $\gamma$ is the descent factor, will locally converge to the optimal measurement angles.
Note that barrier functions should be included as well in order to ensure physically feasible solutions only.
Using the squared Frobenius norm leads to
\begin{equation}
    \frac{\partial \norm{\mat{J}}_F^2}{\partial \varphi_i} =
    2 \sum_{i=1}^m \sum_{j=1}^n J_{ij} \frac{\partial J_{ij}}{\partial \varphi_i}
\end{equation}

In other cases we want to plan a path or trajectory that will contain optimal measurement poses.
Such a \emph{simultaneous optimization} problem can be defined by having a set of roll and pitch angles ${(\vc{\phi}, \vc{\theta})}$.
The optimal angles are then a solution to the optimization problem over all $\vc{\varphi}$ such that
\begin{equation}
    \vc{\varphi}_{1 \ldots K}^* = \underset{\vc{\varphi}_{1 \ldots K}}{\mathrm{arg\,min}} \ \kappa(\mat{J})
    .
    \label{eq:simultaneous_optimal_measurement}
\end{equation}
Though we have shown how trajectories for optimal wind estimation can be obtained in principle,
this active sensing approach to wind estimation is clearly out of scope of this paper and left for future work.

\subsection{Jacobian of the transformed formulation}
\label{sec:jacobian}
Take ${j=(1,2,3)}$ and
the substitutions
${\vc{z} = \vc{v} + \mat{R}_k^T \vc{v}_{0,k}}$,
and ${\vc{y} = \vc{v} - \mat{R}_k^T \left(\vc{v}_{0,k} - v_{i,k} \vc{e}_3\right)}$.
Elements of the Jacobian \eqref{eq:f_jacobian} can now be written as
\begin{equation}
\begin{aligned}
    J_{1j,k} &= 2 v_{i,k}^{2} \, z_{1,j} - 2 R_{3j,k} \, v_{i,k}^{3} , \\
    J_{14,k} &= 4 v_{i,k}^{3} - 6 v_{z,k} v_{i,k}^{2} + 2 v_{i,k} \| \vc{v}_{k} \|^2 , \\
    J_{2j,k} &= -\tfrac{v_{i,k}}{U_k} \left(R_{3j,k} U_k^2 + \left(v_{z,k} - v_{i,k}\right) y_j \right) , \\
    J_{24,k} &= \tfrac{1}{U_k} \left(U_k^2 \left(2 v_{i,k} - v_{z,k} \right) + v_{i,k} \left(v_{z,k} - v_{i,k}\right)^{2}\right) , \\
    J_{3j,k} &= -R_{3j,k} , \\
    J_{34,k} &= 1
    .
\end{aligned}
\label{eq:jacobian_elements}
\end{equation}

\subsection{Derivative of Jacobian}
\label{sec:jacobian_derivative}
In order to obtain the gradient \eqref{eq:sequential_optimal_potential}, the partial derivatives of the Jacobian \eqref{eq:f_jacobian} are required.
Define the partial derivative of $A$ w.r.t $\varphi$ as ${A^{\varphi} := \frac{\partial A}{\partial \varphi}}$.
For ${j=(1,2,3)}$, the partial derivatives of the Jacobian w.r.t a rotation parameter $\varphi$ are
\begin{equation}
\begin{aligned}
    J_{1j}^{\varphi} &=
        2 v_i^2 z_j - v_i^3 R_{3j}^{\varphi}
    , \\
    J_{14}^{\varphi} &=
        - 6 v_i^2 v_z^{\varphi}
    , \\
    J_{2j}^{\varphi} &=
        -v_i \Bigl\{
            U R_{3j}^{\varphi}
            + U^{\varphi} R_{3j}
            \\
            &\quad + \tfrac{1}{U} \bigl[
                v_z^{\varphi} y_j
                + \tfrac{U - 1}{U} y_j^{\varphi} (v_z - v_i)
            \bigr]
        \Bigr\}
    , \\
    J_{24}^{\varphi} &=
        U^{\varphi} ( 2 v_i - v_z )
        - U v_z^{\varphi}
        \\
        &\quad + \tfrac{v_z - v_i}{U}
        \Bigl[
            2 v_i v_z^{\varphi}
            - \tfrac{U^{\varphi}}{U} (v_z - v_i)
        \Bigr]
    , \\
    J_{3j}^{\varphi} &= -R_{3j}^{\varphi}
    , \\
    J_{34}^{\varphi} &= 0
    .
\end{aligned}
\label{eq:jacobian_derivatives}
\end{equation}
Note that we dropped the term associated with ${\norm{\vc{v}_k}^{\varphi}=0}$, because the rotation does not affect the wind velocity norm.

\bibliographystyle{SageH}
\bibliography{bibliography}

\end{document}